\documentclass{article} %
\usepackage{colm2024_conference}

\usepackage{microtype}
\usepackage{hyperref}
\usepackage{url}
\usepackage{booktabs}

\usepackage{pifont}        %
\usepackage{multirow}
\usepackage[]{mdframed}
\usepackage{array}
\usepackage{listings}
\usepackage{xcolor}
\usepackage{xspace}
\usepackage{enumitem}
\usepackage{tabularx}
\usepackage{graphicx}
\usepackage{caption}

\usepackage{wrapfig}

\usepackage{amsmath,amsfonts,bm}
\usepackage{amssymb}
\usepackage{mathtools}
\usepackage{amsthm}
\usepackage{bbm}

\def\eqref#1{equation~\ref{#1}}

\def\1{\bm{1}}

\DeclareMathAlphabet{\mathsfit}{\encodingdefault}{\sfdefault}{m}{sl}
\SetMathAlphabet{\mathsfit}{bold}{\encodingdefault}{\sfdefault}{bx}{n}

\usepackage[capitalize,noabbrev]{cleveref}

\newif\ifisarxiv

\isarxivtrue  %

\newcommand{\myskip}[1]{}
\newcommand{\ngram}{$n$-gram}
\newcommand{\pythia}{\textsc{Pythia}}
\newcommand{\pythiadedup}{\textsc{Pythia-dedup}}
\newcommand{\neo}{\textsc{GPT-Neo}}
\newcommand{\olmo}{\textsc{OLMo}}
\newcommand{\dolma}{\textsc{Dolma}}
\newcommand{\mimir}{\textsc{Mimir}}
\newcommand{\datablations}{\textsc{Datablations}}
\newcommand{\silo}{\textsc{SILO}}
\newcommand{\minkprob}{Min-$k$\% Prob}
\newcommand{\mink}{min-$k$}

\newcommand{\eg}{e.g.,\xspace}
\newcommand{\ie}{i.e.,\xspace}

\makeatletter
\newcommand{\crefnames}[3]{\@for\next:=#1\do{\expandafter\crefname\expandafter{\next}{#2}{#3}}}\makeatother

\crefnames{part,chapter,section}{\S\kern-0.25em}{\S\S\kern-0.25em}

\ifisarxiv
    \newcommand{\repourl}{\url{http://github.com/iamgroot42/mimir}}
\else
    \newcommand{\repourl}{\url{https://anonymous.4open.science/r/mimir-D6E3/}}
\fi

\newcommand\bfsection[1]{\vspace{6pt}{\noindent\bf #1}} %
\newcommand\shortsection[1]{\vspace{6pt}{\noindent\bf #1.}}
\newcommand\shortersection[1]{\vspace{6pt}{\noindent\em #1.}}

\newcommand{\myabstract}{
Membership inference attacks (MIAs) attempt to predict whether a particular datapoint is a member of a target model's training data. Despite extensive research on traditional machine learning models, there has been limited work studying MIA on the pre-training data of large language models (LLMs).
We perform a large-scale evaluation of MIAs over a suite of language models (LMs) trained on the Pile, ranging from 160M to 12B parameters. We find that MIAs barely outperform random guessing for most settings across varying LLM sizes and domains.
Further analyses reveal that this poor performance can be attributed to (1) the combination of a large dataset and few training iterations, and (2) an inherently fuzzy boundary between members and non-members.
We also find that, when LLMs have been shown to be vulnerable to MIAs, this apparent success can be attributed to a distribution shift, e.g., members and non-members are seemingly drawn from identical domain but with different temporal ranges.
Finally, we observe that existing MIAs are highly sensitive to even small changes in a sample. Such changes may cause samples that are lexically or semantically similar to members to be classified as non-members, which may be at odds with leakage that privacy auditors care about.
We release our code and data as a unified benchmark package that includes all existing MIAs, supporting future work.
}

\newcolumntype{C}[1]{>{\centering\arraybackslash}m{#1}}
\newcolumntype{R}[1]{>{\raggedleft\arraybackslash}m{#1}}

\title{Do Membership Inference Attacks Work on Large Language Models?}

\author{Michael Duan$^*$ \And Anshuman Suri$^*$ \And
Niloofar Mireshghallah \AND Sewon Min \And Weijia Shi
\And Luke Zettlemoyer \And Yulia Tsvetkov \AND Yejin Choi
\And David Evans \And Hannaneh Hajishirzi}

\newcommand{\affilsup}[1]{\rlap{\textsuperscript{\normalfont#1}}}

\author{Michael Duan\affilsup{*1} \qquad
    Anshuman Suri\affilsup{*2} \\
    \textbf{Niloofar Mireshghallah}\affilsup{1} \qquad \textbf{Sewon Min}\affilsup{1} \qquad
    \textbf{Weijia Shi}\affilsup{1} \qquad
    \textbf{Luke Zettlemoyer}\affilsup{1} \\
    \textbf{Yulia Tsvetkov}\affilsup{1} \qquad
    \textbf{Yejin Choi}\affilsup{1} \qquad
    \textbf{David Evans}\affilsup{2} \qquad
    \textbf{Hannaneh Hajishirzi}\affilsup{1,3}
    \\
    $^1$University of Washington \quad
    $^2$University of Virginia \quad $^3$Allen Institute for AI
    \\
    \texttt{$<$micdun@cs.washington.edu$>$, $<$as9rw@virginia.edu>}
}

\ifisarxiv
    \colmfinalcopy
\fi
\begin{document}

\maketitle

\def\thefootnote{*}\footnotetext{Equal Contribution. %
}
\def\thefootnote{\arabic{footnote}}

\begin{abstract}
    \myabstract
\end{abstract}

 \section{Introduction}\label{sec:intro}
Membership inference attacks (MIAs) aim to predict whether a particular record belongs to the training dataset of a given model. Thus, MIAs have great utility for privacy auditing of models \citep{steinke2023privacy}, as well as investigating memorization of training data, copyright violations and test-set contamination \citep{shi2023detecting,oren2023proving}.
While MIAs have been found to achieve high attack performance, alluding to high levels of training-data memorization \citep{zarifzadeh2023low, bertran2023scalable, lukas2023analyzing}, most analyses are limited to classifiers or LM fine-tuning \citep{mireshghallah-etal-2022-empirical,fu2023practical}.
The performance of existing MIAs on LLMs and their pre-training data is largely unexplored.
In this work, we set out to explore the challenges in evaluating membership inference attacks on LLMs, across an array of five commonly-used membership inference attacks. %
\ifisarxiv
We introduce \mimir\footnote{\repourl}, a unified repository for evaluating MIAs for LMs, with implementations of several attacks from literature.
\fi
We report on experiments extensively evaluating these MIAs against target models from the Pythia suite \citep{biderman2023pythia} over the Pile \citep{gao2020pile} (\Cref{sec:main-exp}). For the most part, we find that the performance across most MIAs and target domains is \textit{near-random}. %

Our further analysis suggests that the inherent characteristics of LLMs at scale---specifically, the use of massive training data and near-one epoch training (\Cref{subsubsec:ablation-llm-characteristics})---considerably decrease current MIA performance.
This suggests that the success of current MIAs in previous settings does not transfer well to attacking pre-trained LLMs seemingly due to a lack of memorization of member data.
We also find that the frequent overlap between members and non-members from natural language domains considerably decreases MIA performance
and raises the question of how membership should be interpreted (\Cref{subsec:ablation-ambiguity}).
Notably, in several domains, non-members have high \ngram\ overlap with members, \eg non-members from the Pile Wikipedia and ArXiv test samples have average 7-gram overlaps of over 30\%. 
Notably, non-members with lower \ngram\ overlap are more distinguishable by existing MIAs.
We also suggest that high MIA performance reported by prior work~\citep{shi2023detecting} is likely because non-members are chosen from the same domain as members but are temporally shifted, and these seemingly in-domain non-members likely belong to a different distribution as a result of \ngram\ overlap shift (\Cref{sec:distribution-shift}).

Finally, building off membership ambiguity due to $n$-gram overlap, we discuss how the precise definition of members in standard MI may not capture important information leakage under generative text-modeling.
We generate modified members preserving lexical and/or semantic similarity by altering a tiny fraction of tokens and show that existing MIAs classify them as non-members with a high degree of confidence, often more definitively than actual non-members (\Cref{app:edit_distance_members}).
We encourage future work to study MI using membership definitions accounting for such fuzzy members to better understand privacy leakage.

\section{Background}\label{subsec:main-exp-mia-methods}

The goal of an MIA is to infer whether a given data point $x$ was part of the training dataset $\mathcal{D}$ for model $\mathcal{M}$, by computing a membership score $f(\mathbf{x}; \mathcal{M})$. This score is then thresholded to determine a target sample's membership.

MIAs are often used as a proxy to determine whether a machine-learning model leaks information related to its training data \citep{shokri2017membership, shokri2022auditing, cummings2024hdsr}.
It is the de-facto threat model when discussing machine-learning privacy \citep{shokri2017membership}, with a large array of attacks~\citep{yeom2018privacy, carlini2022membership,mireshghallah2022quantifying} and defenses~\citep{Abadi_2016,tang2021mitigating,chen2022relaxloss}. 
More involved approaches include training shadow models  \citep{shokri2017membership,ye2022enhanced} on non-overlapping data from the target model's underlying data distribution.
While attacks like LiRA \citep{carlini2022membership} show promise, they require training multiple copies of shadow models, which is often intractable for LLMs. Other stronger assumptions for MIAs include
white-box access to the model (\ie access to model parameters) or access to a ground-truth subset of member/in-distribution non-member samples for training meta-classifiers.

In our setting, $\mathcal{M}$ is an auto-regressive language model that outputs a probability distribution of the next token given a prefix, denoted as $P(x_t|x_1...x_{t-1}; \mathcal{M})$. We consider five MIAs (See \Cref{subsec:appendix-attack-details} for detailed descriptions): 

(1)~\textbf{LOSS}~\citep{yeom2018privacy} - the target sample's loss under the model:
$
f(\mathbf{x}; \mathcal{M}) = \mathcal{L}(\mathbf{x};\mathcal{M}).
$

(2)~\textbf{Reference-based}~\citep{carlini2021extracting} calibrates $\mathcal{L}(\mathbf{x};\mathcal{M})$ with respect to another \textit{reference model} ($\mathcal{M}_{ref}$)
to account for the intrinsic complexity of the target sample \textbf{x}:
$
f(\mathbf{x}; \mathcal{M}) = \mathcal{L}(\mathbf{x}; \mathcal{M})  - \mathcal{L}(\mathbf{x}; \mathcal{M}_\text{ref}).
$

(3)~\textbf{Zlib Entropy}~\citep{carlini2021extracting} calibrates $\mathcal{L}(\mathbf{x};\mathcal{M})$ with target sample \textbf{x}'s zlib compression size:
 $
f(\mathbf{x}; \mathcal{M}) = \mathcal{L}(\mathbf{x}; \mathcal{M}) / \text{zlib}(\mathbf{x}).
$

(4)~\textbf{Neighborhood attack} \citep{mattern2023membership} - the curvature of the loss function at \textbf{x}, estimated by perturbing the target sequence to create $n$ 'neighboring' samples, and comparing the loss of the target \textbf{x} with its neighbors $\tilde{\mathbf{x}}$: $
f(\mathbf{x}; \mathcal{M}) = \mathcal{L}(\mathbf{x}; \mathcal{M}) - \frac{1}{n}\sum_{i=1}^n \mathcal{L}(\mathbf{\tilde{x}}_i; \mathcal{M}).
$

(5) \textbf{\minkprob}~\citep{shi2023detecting} uses the $k\%$ of tokens with the lowest likelihoods to compute a score instead of averaging over all token probabilities as with LOSS:
$
f(\mathbf{x}; \mathcal{M}) = \frac{1}{\lvert \text{min-$k(\mathbf{x})$}\rvert} \sum_{x_i \in \text{min-$k(\mathbf{x})$}}-\log(p(x_i \mid x_1,...,x_{i-1})).
$

\shortsection{Membership Inference vs.\ Data Extraction}\label{subsec:future-work-mi-vs-extraction} MIA advantage is frequently used as a measure of information leakage \citep{shokri2022auditing,shokri2017membership,mireshghallah2022quantifying} and a proxy for measuring memorization \citep{carlini2021extracting,mireshghallah-etal-2022-empirical}, with recent attempts studying user-level leakage for the fine-tuning setting \citep{kandpal2023user}. However, the `extractability' of training samples has recently become synonymous with memorization and is increasingly used to compare memorization across models \citep{biderman2023emergent,carlini2023quantifying,tirumala2022memorization}. \citet{kandpal2022deduplicating} investigated the impact of factors such as training data deduplication on extractability in a similar vein to our work on MIA. With extraction, a prefix is used as a prompt to measure the memorization of a sequence by comparing the resulting generation against the suffix.
Both MIA and extraction are useful techniques for studying leakage in models, but rely on different assumptions and reveal different types of leakage risks. While MIAs require knowledge of candidates and only reveal directly which of those candidates are included in the training data, extraction requires knowledge of sufficient-length prefixes to perform extraction and additional measures to determine if extracted texts are valid.

\section{Membership Inference on LLMs is Difficult}\label{sec:main-exp}

We perform a large-scale evaluation of five state-of-the-art MIAs (\Cref{subsec:main-exp-mia-methods}) on a range of LLMs with up to 12B parameters and diverse benchmarks. For the reference-based attack in \Cref{tab:gen_exp_results} and all following experiments, we use \textsc{Stablelm-Base-Alpha-3B-v2} as the reference model (determined empirically in \Cref{subsec:ref_model_choice}).
Code is available at \repourl.

\begin{table*}[t]
\scriptsize
\begin{center} 
\setlength{\tabcolsep}{0.7pt}
\begin{tabularx}{\textwidth}{l *{20}{>{\centering\arraybackslash}X}@{}}
    \toprule
    \multirow{2}{*}{}  & \multicolumn{5}{c}{Wikipedia} & \multicolumn{5}{c}{Github} & \multicolumn{5}{c}{Pile CC} & \multicolumn{5}{c}{PubMed Central} \\
    \cmidrule(lr){2-6}  \cmidrule(lr){7-11} \cmidrule(lr){12-16} \cmidrule(lr){17-21}
    \textbf{\# Params} & \tiny{LOSS} & \tiny{Ref} & \tiny{\mink} & \tiny{zlib} & \tiny{Ne}
    & \tiny{LOSS} & \tiny{Ref} & \tiny{\mink} & \tiny{zlib} & \tiny{Ne}
    & \tiny{LOSS} & \tiny{Ref} & \tiny{\mink} & \tiny{zlib} & \tiny{Ne}
    & \tiny{LOSS} & \tiny{Ref} & \tiny{\mink} & \tiny{zlib} & \tiny{Ne}
    \\
    \midrule
        70M & .503 & .504 & .494 & .508 & \textbf{.510} & .629 & .584 & .627 &\textbf{.648} & .635 & .494 & .489 & \textbf{.503} & .495 & .489 & .502 & \textbf{.516} &.510 &.502 & .485\\
        160M & .504 & \textbf{.515}& .488& .514& .513 & .638& .591& .634& \textbf{.656}& .638 & .497& .497& \textbf{.503}& .498& .496 & .500 & \textbf{.516}& .504& .500& .486\\
        1.4B & .510 & \textbf{.544}& .506& .518& .518& .656& .587& .654& \textbf{.670}& .650& .500& \textbf{.525}& .509& .502& .499 & .496& \textbf{.530}& .505& .500 & .490 \\
        2.8B & .516 & \textbf{.565}& .511& .522& .517& .707& .657& .708& \textbf{.717}& .698 & .501& \textbf{.537}& .509& .503& .502 & .498& \textbf{.536}& .502& .500& .497\\
        6.9B & .514 & \textbf{.571}& .512& .521& .514& .672& .573& .675& \textbf{.684}& .654 & .511& \textbf{.564}& .516& .512& .505 & .504& \textbf{.552}& .508& .504& .497 \\
        12B & .516 & \textbf{.579}& .517& .524& .520 & .678& .559& .683& \textbf{.690}& .660 & .516& \textbf{.582}& .521& .517& .514 & .506& \textbf{.559}& .512& .506& .497\\
        \vspace{-.6em} \\
    \toprule
    \multirow{2}{*}{}  & \multicolumn{5}{c}{ArXiv} & \multicolumn{5}{c}{DM Math} & \multicolumn{5}{c}{HackerNews} & \multicolumn{5}{c}{The Pile}\\
    \cmidrule(lr){2-6}  \cmidrule(lr){7-11} \cmidrule(lr){12-16} \cmidrule(lr){17-21}
    \tiny{\textbf{\# Params}} & \tiny{LOSS} & \tiny{Ref} & \tiny{\mink} & \tiny{zlib} & \tiny{Ne}
    & \tiny{LOSS} & \tiny{Ref} & \tiny{\mink} & \tiny{zlib} & \tiny{Ne}
    & \tiny{LOSS} & \tiny{Ref} & \tiny{\mink} & \tiny{zlib} & \tiny{Ne}
    & \tiny{LOSS} & \tiny{Ref} & \tiny{\mink} & \tiny{zlib} & \tiny{Ne}
    \\
    \midrule
        70M & \textbf{.506} & .481 & .499 & .495 & .496 & 
        .492& \textbf{.520} & .495& .485& .481 &
        .494 & .495 & \textbf{.507} & .497 & .506 &
        .503& \textbf{.511}& .508& .506& .499\\
        160M & \textbf{.507} & .486& .501& .500& \textbf{.507} & .490& \textbf{.523}& .493& .482&  .489 & .492& .490 & .497& .497&\textbf{ .505} & .502 & \textbf{.511} & .506 & .505 & .499\\
        1.4B &\textbf{.513}& .510& .511& .508& .511& .486& \textbf{.512} & .497& .481& .465 & .503& \textbf{.514} & .509& .502& .504 & .504 & \textbf{.521} & .508 & .507 &  .504\\
        2.8B & .517& \textbf{.531}& .522& .512& .519 & .485& \textbf{.504}& .497& .482& .467 & .510& \textbf{.549} & .518& .507& .513 & .507 & \textbf{.530} & .512 & .510 & .506\\
        6.9B & .521& \textbf{.538}& .524& .516& .519  & .485& \textbf{.508}& .496& .481& .469 & .513& \textbf{.546} & .528& .508& .512 & .510 &\textbf{.549} & .516 & .512 & .510 \\
        12B & .527& \textbf{.555}& .530& .521& .519 & .485& \textbf{.512}& .495& .481& .475 & .518& \textbf{.565}& .533& .512& .515 & .513 & \textbf{.558} & .521 & .515 & .511 \\
    \bottomrule
\end{tabularx}
\end{center}
\vskip -1em
\caption{AUC ROC of MIAs against \pythiadedup\ (TPR@low\%FPR results in \Cref{tab:appendix-gen_exp_tprlowfpr_results}). Highest performance across different MIAs is bolded per domain. \textbf{MIA methods perform near random ($<.6$) in most domains.} See Appendix~\ref{subsec:appendix-github-outlier} for GitHub outlier discussion.}\label{tab:gen_exp_results}
\vskip -.2em
\end{table*}

\shortsection{Target models} \label{subsec:main-exp-target-model}
We primarily target the \pythia\ model suite, including (1) five models of \textbf{\pythia} \citep{biderman2023pythia} with 160M, 1.4B, 2.8B, 6.7B, and 12B parameters \footnote{We include results for the 70M variant in \Cref{tab:gen_exp_results} for completeness over the model suite, but choose to focus mainly on the larger models for the later experiments.}, trained on the original Pile data~\citep{gao2020pile}, and (2) five models of \textbf{\pythiadedup}~\citep{biderman2023pythia} with the same parameter counts as \pythia\, but trained on the deduplicated Pile data.
We also experiment with the \neo\ family to validate our findings with a different model family, observing similar trends in most domains (see \Cref{app:neo_results}).

\shortsection{Datasets}\label{subsec:main-exp-datasets}
We use seven diverse data sources included in the Pile: general web (Pile-CC), knowledge sources (Wikipedia), academic papers (PubMed Central, ArXiv), dialogues (HackerNews), and specialized-domains (DM Math, Github). We also perform experiments over the entire Pile. Members and non-members for each data source are sampled from the train and test sets of the Pile, respectively. \citet{gao2020pile} decontaminated the Pile test set against the training set at a document level; nonetheless, to be more rigorous, we perform additional deduplication %
following \citet{bff2023}. We additionally sample from documents greater than 100 words, and truncate samples up to 200 words.
Refer to \Cref{subsec:appendix-benchmark-details} for further details.

\shortsection{Evaluation metrics}
We primarily report \textbf{AUC ROC} for our evaluations, and 
additionally record \textbf{TPR}@\textbf{low\%FPR}~\citep{carlini2022membership} to assess performance in high-confidence settings. We visualize the 95\% confidence interval for AUC ROC scores via shaded regions.

\subsection{Main Results}\label{subsec:main-exp}

\begin{figure*}[t]
\begin{center}
\includegraphics[width=\textwidth]{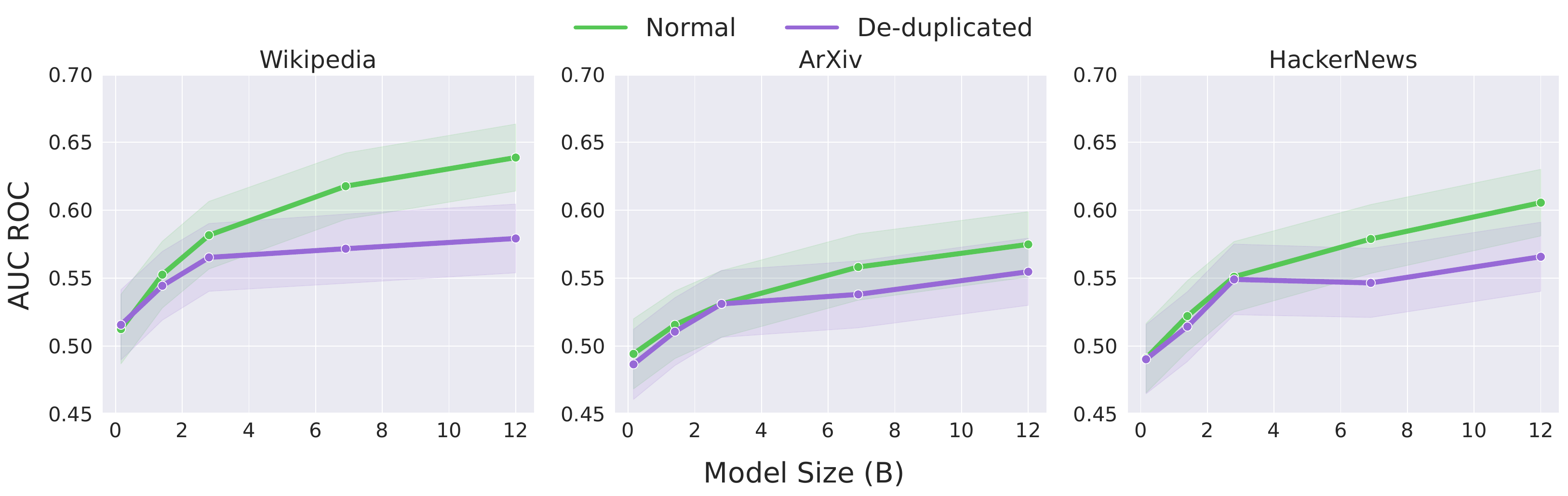}
\vspace{-.5em}
\caption{
MIA performance as model size increases for the reference-based attack over select domains.
We also plot the AUC ROC trajectory against the non-deduped Pythia suite for comparison. \textbf{Increasing model size slightly boosts MIA performance while deduplication decreases performance}. Other attacks follow similar trends (Appendix \Cref{fig:gen_exp_model_size+dedup_others}). 
}
\label{fig:gen_exp_model_size+dedup}
\end{center}
\vspace{-1em}
\end{figure*}

\shortsection{MIAs perform near random}
\Cref{tab:gen_exp_results} shows that all existing MIAs perform near random for most domains\footnote{Similar trends for TPR@1\%FPR. See \Cref{tab:appendix-gen_exp_tprlowfpr_results} in Appendix.}.
No single MIA or target model demonstrates attack AUC above 0.6, with the exception of Github domain (see Appendix~\ref{subsec:appendix-github-outlier} for discussion).
Overall, the reference-based attack performs best, although there are a few settings where other attacks perform better, e.g., \minkprob~on Pile CC for the 160M \pythiadedup\ model.
Marginal differences in performance across MIAs make it hard to single out an overall {\em best} attack. 

MIA performance tends to increase with the target model size (\Cref{tab:gen_exp_results}, \Cref{fig:gen_exp_model_size+dedup}), in agreement with prior work~\citep{shi2023detecting, li2023mope, watson2022on}. This is likely because larger models are more prone to overfitting the training data~\citep{nakkiran2019deep}.  %
We also find deduplication of the training data reduces MIA performance (\Cref{fig:gen_exp_model_size+dedup}), confirming the findings from \citet{kandpal2022deduplicating}.

\shortsection{Difficulty in Choosing a Reference Model} \label{subsec:ref_model_choice}
We ablate the choice of reference models in the reference-based attack in \Cref{subsec:appendix-ref-model}.
In summary, (1) most reference models yield poor performance, with \textsc{Stablelm-Base-Alpha-3B-v2} being the best overall, and (2) even aggregating all reference models performs poorly.
In general, we find choosing the right reference model for a target LLM challenging and largely empirical.
A reference model should be trained on the data that is same-distribution but largely disjoint from the training data of the target model.
However, this assumption is hard to impose at the scale of pre-training corpora; common practice is to collect all the data available on the web, leading independently collected datasets to naturally overlap with each other.

\myskip{
\begin{table}[h]
\begin{center} \footnotesize
\caption{The effect of different reference model choices when targeting \textsc{\pythiadedup-12B}. The reference model yielding the highest performance is bolded. AUC ROC reported.
}\label{tab:ref-results} \vspace{.3em}\scriptsize
\begin{tabular}{lcc}
    \toprule
    \multirow{2}{*}{Reference Model}  & \multirow{2}{*}{\textsc{Wikipedia}} & \textsc{PubMed} \\
    & & \textsc{Central} \\
    \midrule
    \textsc{Gpt-2-small} & .514 & .501\\
    \textsc{distilGpt2} & .510 & .498\\
    \textsc{Opt-1.3B} & .522 & .523\\
    \textsc{Gpt-Neo-1.3B} & .528 & .531\\
    \textsc{Pythia-dedup-1.4B} & .534 & .537\\
    \textsc{Silo-pdswby} & .529 & .521\\
    \textsc{Stablelm-Base-Alpha-3B-v2} & \textbf{.579} & \textbf{.559}\\
    \textsc{Llama-7B} & .546 & .531\\
    \midrule
    Aggregation of Eight Models & .534 & .554\\
    \bottomrule
\end{tabular}
\end{center}
\vskip -0.1in
\end{table}
}

\subsection{Why is MI Challenging against LLMs}\label{sec:abl}
We identify several key factors that may contribute to the decreased performance of MIAs on LLMs.
Some factors are due to unique characteristics of practices in LLM pre-training (\Cref{subsubsec:ablation-llm-characteristics}) while others are due to inherent ambiguity in MIA (\Cref{subsec:ablation-ambiguity}).

\subsubsection{Characteristics of LLMs}\label{subsubsec:ablation-llm-characteristics}%
\shortsection{Training Data Size} Current state-of-the-art pretrained LLMs are trained with billions and trillions of tokens~\citep{touvron2023llama1, touvron2023llama, MosaicML2023Introducing}.
We hypothesize the \textit{\textbf{large pretraining corpora characteristic to LMs decreases MIA performance}}, as larger pretraining datasets lead to better generalization~\citep{hoffmann2022training, muennighoff2023scaling}.
\begin{figure}[!t]
\begin{center}
\includegraphics[width=.49\columnwidth]{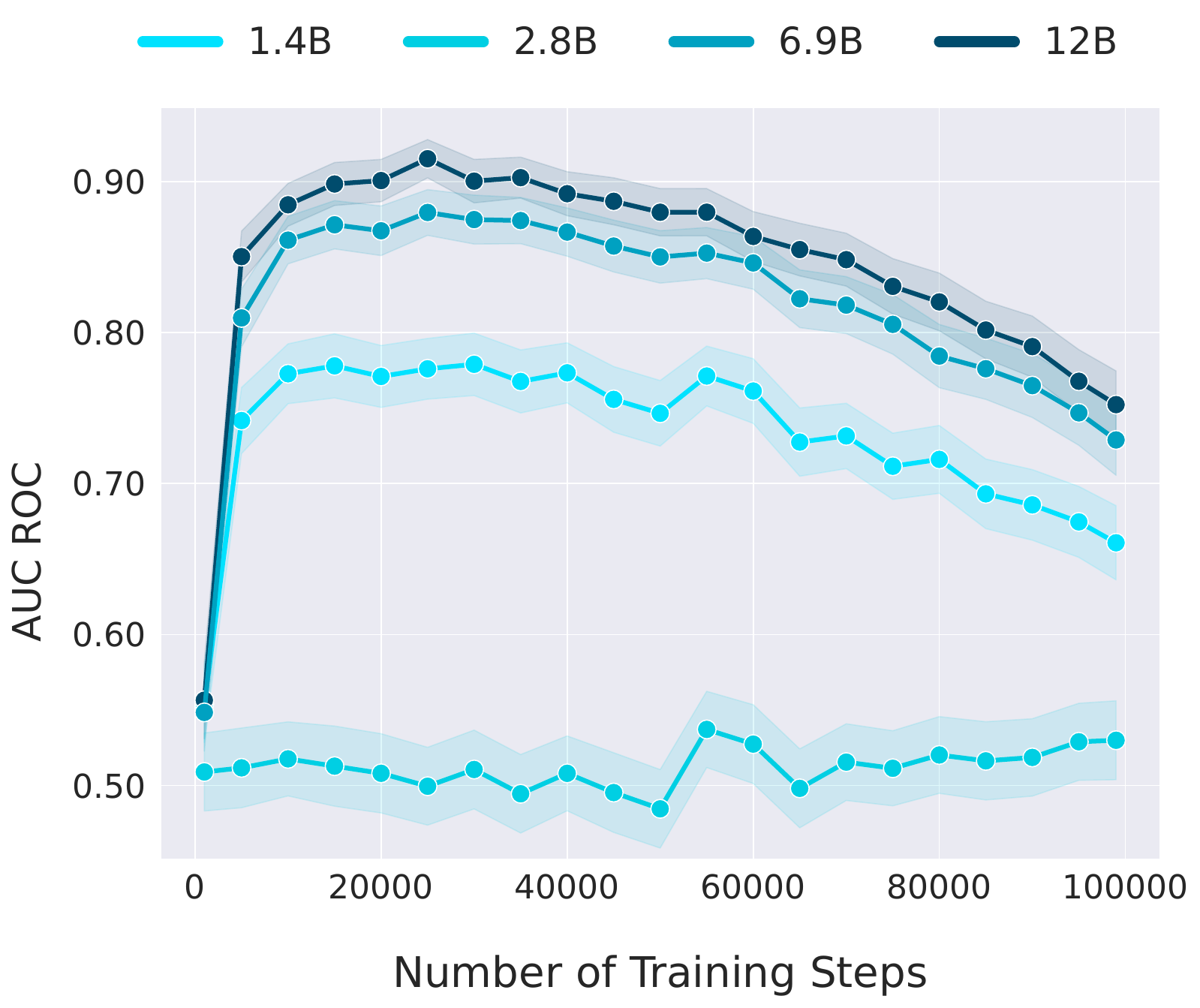}
\includegraphics[width=.49\columnwidth]{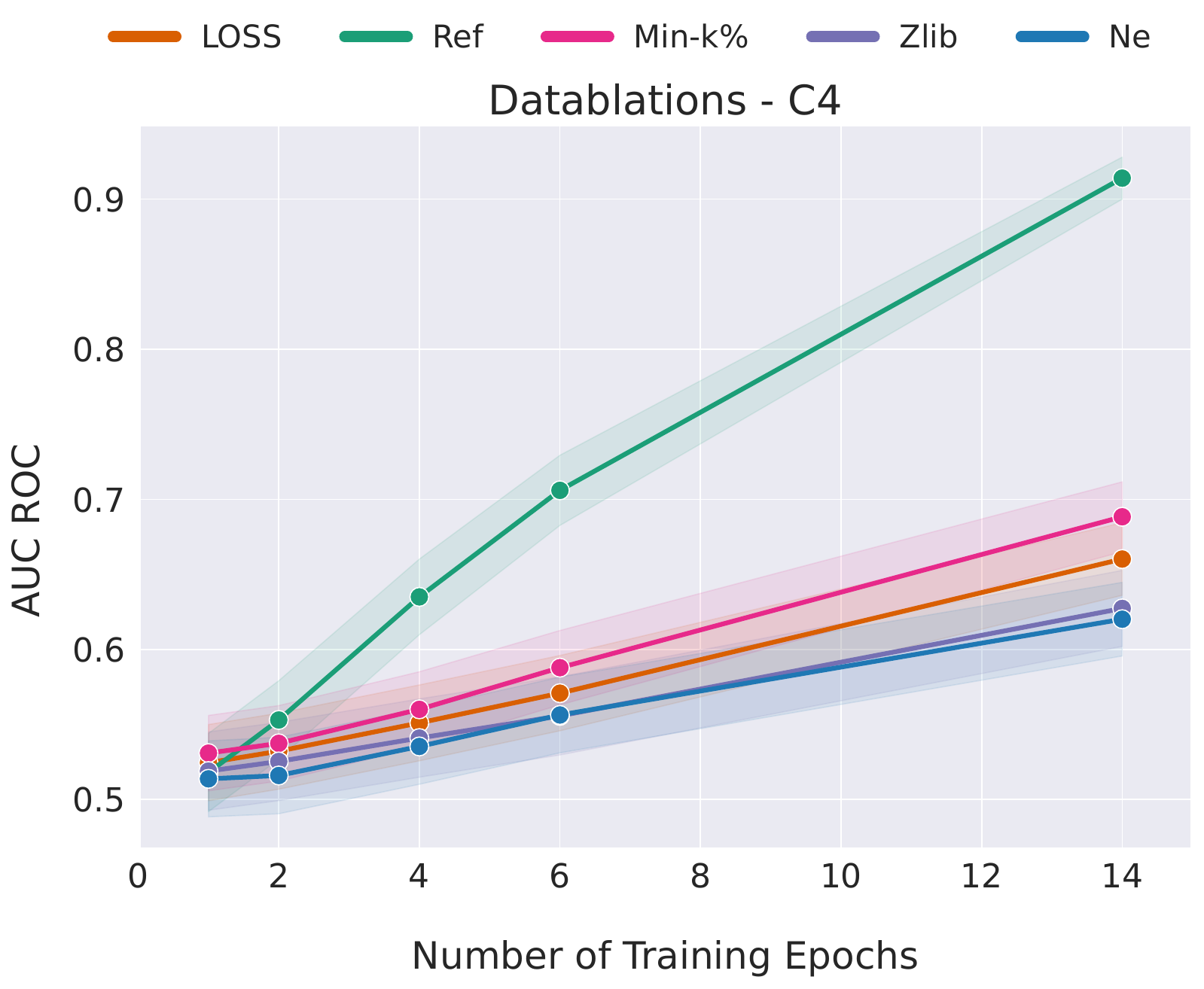}
\caption{
    (Left) Reference-based attack performance as the amount of training data seen, measured in the number of training steps, increases across 1 epoch of the deduplicated Pile.
    In general, \textbf{performance spikes greatly before gradually decreasing as the amount of training data seen increases}. Other attacks (\Cref{fig:training-data-size-others}, Appendix) follow similar trends. 
    (Right) MIA performance on target model \datablations\ as the number of effective epochs increases via increasing epoch count. \textbf{Performance increases linearly with the number of effective epochs}. See \Cref{fig:num-epochs-silo} for results on \silo. %
}
\label{fig:training-data-size}
\end{center}
\vspace{-.75em}
\end{figure}

We employ the \pythiadedup\ model suite's intermediate checkpoints to assess the impact of different amounts of training data. %
While keeping non-members fixed, we sample members for each checkpoint from its most recent 100 steps to remove the impact of the recency bias of the members.\footnote{Using recently seen members elevates MIA performance noticeably, but doesn't disrupt the impact of increasing training data size.
See Appendix~\ref{sec:appendix-data-recency} for the impact of the recency of data seen in training.} See Appendix~\ref{subsubsec:appendix-tds-benchmark-details} for details. 

MIA performance generally starts as near-random, then rapidly increases within the next few thousand steps, before decreasing across successive checkpoints\footnote{\pythiadedup-2.8B stands apart with a performance trajectory that is consistently near-random. Previous work also observes unexplainable behavior for this model~\citep{biderman2023emergent}.} (\Cref{fig:training-data-size}, left). We speculate the initial low performance is due to the model warming up in training, with high losses across both member and non-member samples. We believe the following rapid rise and then gradual decline in performance are because the data-to-parameter-count ratio is smaller early in training and the model may tend to overfit \citep{yeom2018privacy}, but generalizes better as training progresses, in line with observations in existing work \citep{nakkiran2019deep}. 

\shortsection{Number of Training Epochs} %
It is standard practice to pre-train LLMs for around one epoch, given the scale of data and their tendency to overfit quickly~\citep{muennighoff2023scaling, komatsuzaki2019epoch}. Previous MIA works that demonstrate attack effectiveness consider supervised fine-tuning or masked LM pre-training~\citep{nakamura2020kart,lehman2021does,mireshghallah2022quantifying,mireshghallah-etal-2022-empirical}, where models are trained for more than 10 epochs. %
We explore how the \textit{\textbf{near-one epoch training of LLMs leads to decreased MIA performance}}.

To verify this hypothesis, we perform MIA against
the \textbf{Datablations} suite \citep{muennighoff2023scaling}, consisting of models trained on subsets of C4~\citep{2019t5} train data for varying numbers of \textit{epochs} (see \Cref{subsec:appendix-target-model-details} for model details).
We also experiment with \textbf{SILO}~\citep{silo}; see \Cref{sec:appendix-training-epoch} for \silo\ results.

Increasing the number of effective epochs corresponds to an increase in attack performance (\Cref{fig:training-data-size}, right). While \citet{muennighoff2023scaling} shows training for multiple epochs helps improve performance, our results suggest that such multi-epoch training (and/or large upsampling factors) can increase training data leakage. %

\subsubsection{Inherent Ambiguity in MIA}\label{subsec:ablation-ambiguity}
Natural language documents commonly have repeating text---even with the best efforts in decontamination and deduplication. These include common phrasings and quotes, natural use of similar texts, and syntactical similarities inherent to specific domains. 
This leads to substantial text overlap between members and non-members, which motivates the following hypothesis: \textit{\textbf{higher overlap between members and non-members increases MIA difficulty}}.

We quantify overlap using the percentage of \ngram\ overlap, defined as\footnote{Non-members can still have 100\% \ngram\ overlap without being members as {\ngram}s can appear in different member samples. However, this is increasingly unlikely for larger $n$.}: For a non-member sample $\mathbf{x}$ consisting of $m$ words such that $\mathbf{x} = x_1x_2...x_m$ and an \ngram\ in $\mathbf{x}$ defined as a continuous substring $x_i...x_{i+n-1}$ %
, the \ngram\ overlap of $\mathbf{x}$ on training dataset $D$ is
$$\frac{1}{m - n + 1}\sum\limits_{i=1}^{m-n+1} \mathbbm{1}\{\exists y \in D : x_i...x_{i+n-1} \in y\}$$

\begin{figure*}[!t]
\begin{center}
\includegraphics[width=\columnwidth]%
{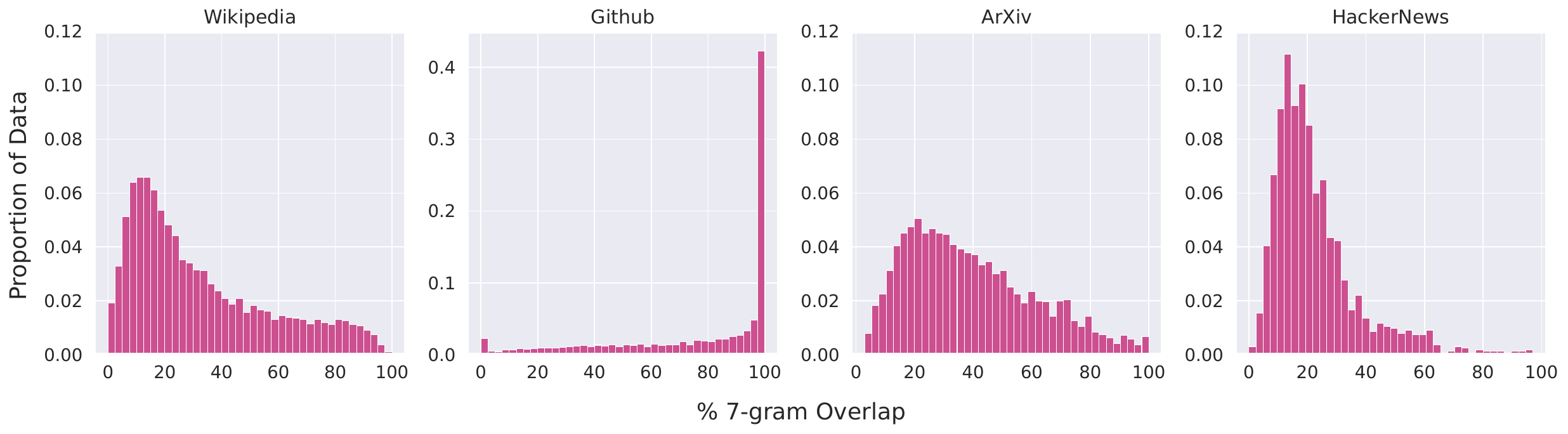}
\vspace{-.75em}
\caption{Natural distributions of 7-gram overlap of non-member data over select domains. Github has a considerably higher overlap than other domains.
}
\label{fig:ngram-overlap-distribution}
\vspace{-1.25em}
\end{center}
\end{figure*}

For a given non-member sample, this gives us the percentage of {\ngram}s in the non-member that can be found in at least one member sample. We first compute the percentage of $7$-gram overlap for non-members against the entire Pile training (member) set (\Cref{fig:ngram-overlap-distribution})\footnote{\Cref{fig:appendix_other_ngram_overlap} shows \ngram\ overlap distributions for other $n$.}; see \Cref{subsec:appendix-ngram-overlap-implementation} for implementation details.
We observe high \ngram\ overlap with training data for a substantial portion of non-members; \eg the Wikipedia, ArXiv, and PubMed Central domains have average 7-gram overlaps of $32.5\%$, $39.3\%,$ and $41.0\%$, respectively.
Domains such as GitHub, DM Mathematics, and FreeLaw see even higher overlap, with mean 7-gram overlap of $76.9\%, 72.8\%$, and $62.3\%$, respectively.

High \ngram\ overlap suggests that substrings of non-members may be seen exactly during training, which makes the distinction between members and non-members even less clear.
To verify our hypothesis, we resample non-members ensuring $\leq20\%$ \ngram\ overlap with members, and report MIA performance (\Cref{tab:ngram_threshold_results}). While this step is designed to more strictly eliminate instances of non-member records that may overlap with training records,
it also introduces an explicit drift between member and non-member distributions by selecting non-members that are most ``unlike" training records. We clarify that this step is not a suggestion for researchers to alter their benchmarks; such a processing step drifts away from the standard membership inference game \citep{yeom2018privacy}.

\begin{table}[!t]
\begin{center} \scriptsize
\begin{tabular}{l cccccccccc}
    \toprule
    \multirow{2}{*}{Domain} & \multicolumn{2}{c}{Wikipedia}
    & \multicolumn{2}{c}{Github}
    & \multicolumn{2}{c}{PubMed Central}
    & \multicolumn{2}{c}{Pile CC}
    & \multicolumn{2}{c}{ArXiv}
    \\
    \cmidrule(lr){2-3} \cmidrule(lr){4-5} \cmidrule(lr){6-7} \cmidrule(lr){8-9} \cmidrule(lr){10-11} 
    & \textsc{Orig} & \textsc{7-gram} & \textsc{Orig} & \textsc{7-gram}& \textsc{Orig} & \textsc{7-gram}& \textsc{Orig} & \textsc{7-gram}& \textsc{Orig} & \textsc{7-gram} \\
    \midrule
    LOSS    & .516 & .666 & .678 & .878 & .506 & .780 & .516 & .574 & .527 & .787 \\
    Ref     & .579 & .677 & .559 & .615 & .559 & .595 & .582 & .644 & .555 & .715 \\
    min-$k$ & .517 & .644 & .683 & .890 & .512 & .792 & .521 & .578 & .530 & .734\\
    zlib    & .524 & .631 & .690 & .908 & .506 & .772 & .517 & .560 & .521 & .780 \\
    Ne      & .520 & .612 & .660 & .877 & .497 & .737 & .514 & .566 & .519 & .773\\
    \bottomrule
\end{tabular}
\caption{Comparison of MIA performance over select domains with varying non-member sets at $\leq20$\% \ngram\ overlap threshold for $n = 7$, as well as the natural non-member set. Target model is \pythiadedup-12B and AUC ROC reported. \textbf{Strict \ngram\ overlap thresholding results in higher performance}.}\label{tab:ngram_threshold_results}
\vskip -.5em
\end{center}
\vskip -0.5em
\end{table}

\shortsection{Results}
MIAs perform significantly better as the non-member distribution concentrates towards lower \ngram\ overlap and further diverges from the natural \ngram\ overlap distribution of non-members from the training distribution
\eg\ .516$\rightarrow$.666 in Wikipedia, .690$\rightarrow$.908 in Github, and .512$\rightarrow$.792 in PubMed Central for various attacks.
This is intuitive as the target model is likely to assign a lower likelihood to non-members further from its training data, making members and non-members more distinguishable.
Note that decreasing the \ngram\ overlap threshold, especially for smaller $n$, pushes the setting closer to distribution inference \citep{suri2022formalizing}, since the distributions of `member' and `non-member' records are no longer the same. We further discuss outlier behavior in \Cref{sec:appendix-other-ngram-takeaways}.

We note that \ngram\ overlap is an intrinsic property of natural language rather than a problem of the Pile train-test split. These splits are already deduplicated at a document level, following standard practice in decontamination~\citep{gao2020pile, gpt3}. Nonetheless, repeating texts across distinct documents are fundamental and natural properties of domain data.
We also note that \ngram\ overlap distribution analysis can help assess how representative of a target domain a set of candidate non-members is when constructing MIA benchmarks.
Ultimately, we highlight the need to consider qualities of the data domain, \eg \ngram\ overlap, and understand their potential impact on MIA performance.

\section{Importance of Candidate Set Selection}\label{sec:distribution-shift}

In contrast to our findings in \Cref{sec:main-exp}, recent works report state-of-the-art MIAs achieving $>.7$ AUC ROC on pretrained LLMs~\citep{shi2023detecting, meeus2023did}. We investigate how the non-member candidate selection methods in these works can result in an inherent but likely unintended distribution shift between members and non-members as one such reason for the observed performance differences.

\shortsection{Experimental Setup}
Prior work distinguishes members and non-members of a target domain based on the knowledge cutoff date of the target model, with members coming before and non-members coming after the cutoff. We construct similar experimental settings under two domains also used in earlier works: Wikipedia and ArXiv. %

We follow the same setup
as \Cref{sec:main-exp} on the Wikipedia domain, but replace the non-member set with samples from the entire RealTimeData WikiText dataset~\citep{li2023avoiding}.
The Pile members are sampled from articles in a Wikipedia dump from before  March 2020~\citep{gao2020pile}, whereas non-members consist of Wikipedia articles created from August 2023 and onwards. For the ArXiv domain or further details on Wikipedia, see~\Cref{subsubsec:appendix-temporal-benchmark-details}.

 \begin{figure}[ht]
    \centering
    \begin{minipage}[b]{0.48\textwidth}
    \centering
            \scriptsize
            \begin{tabularx}{\columnwidth}{l *{5}{>{\centering\arraybackslash}X}@{}}%
                \toprule
                \multirow{2}{*}{\textbf{\# Params}}  & \multicolumn{4}{c}{Temporal Wiki}\\
                \cmidrule(lr){2-6}
                 & LOSS & Ref & min-k & zlib & Ne
                \\
                \midrule
                    160M & .643 & .602 & \textbf{.648} & .541 & .600\\
                    1.4B & .653 & \textbf{.705} & .682 & .572 & .603\\
                    2.8B & .667 & \textbf{.754} & .701 & .593 & .615\\
                    6.9B & .675 & \textbf{.788} & .714 & .601 & .620\\
                    12B & .680 & \textbf{.796} & .719 & .607 & .626\\
                \bottomrule
            \end{tabularx}
            \captionof{table}{AUC-ROC on the temporally shifted Wikipedia benchmark across various MIAs. Target models are the \pythiadedup\ suite models. For each model, the highest score across MIAs is bolded.
            }\label{tab:wikiMIA_results}
    \end{minipage}
    \hfill
    \begin{minipage}[b]{0.49\textwidth}
        \centering
        \includegraphics[width=\columnwidth]{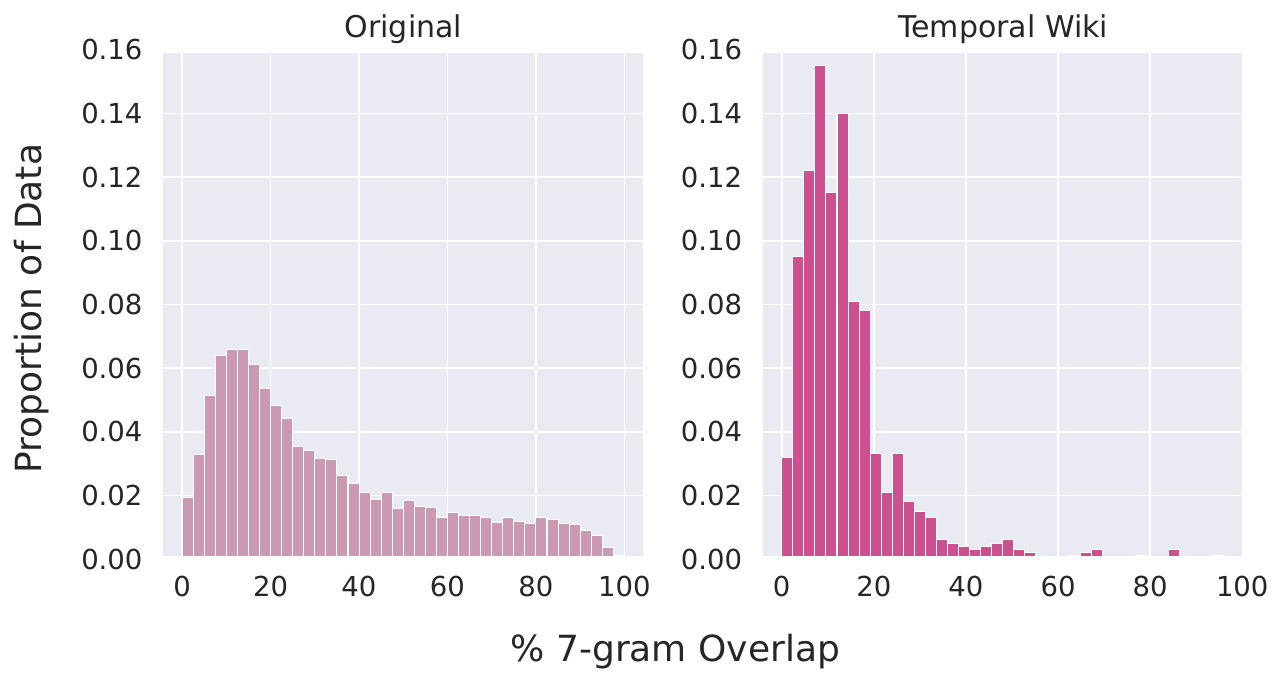}
        \vspace{-1.5em}
        \caption{Distribution of 7-gram overlap for the original and temporally-shifted non-members.}
        \label{fig:temporal_wiki}
    \end{minipage}
\vspace{-.5em}
\end{figure}

\shortsection{Results}
\Cref{tab:wikiMIA_results} demonstrates that the temporally shifted settings yield MIA performances significantly higher than when members and non-members are from the same temporal range. \Cref{fig:temporal-arxiv} (Appendix) also demonstrates that MIA performance generally increases as non-members are further temporally shifted from member data. We speculate this follows from changes in language such as the introduction of new terminology and ideas over time. 

\shortersection{Temporal Shift as Change in \ngram\ Overlap}
We interpret temporal shift as a change in \ngram\ overlap distribution between the original and temporally shifted non-members. \Cref{fig:temporal_wiki} demonstrates that the distribution of 7-gram overlap of such shifted data concentrates at lower overlap percentages compared to their natural counterparts. The natural Wikipedia non-members have an average 7-gram overlap of 39.3\%, whereas for the temporally shifted Wikipedia non-members it is 13.9\%. See \Cref{subsec:appendix-temporal-arxiv} for temporal ArXiv discussion.

In general, when aiming to assess MIA performance, we advise estimating how representative a sample non-member set is of the member domain by comparing its \ngram\ overlap distribution with that of a left-out sample set from the pretraining corpora. Particularly, if the distribution of the candidate set is noticeably shifted towards lower \ngram\ overlap compared to the left-out member sample set, the candidate non-member set may not be representative of the member distribution from the target domain and potentially high MIA performances should be carefully examined.
Closer inspection (\Cref{{tab:temporal_crossdomain}}, Appendix) reveals the extent of such over-estimation;  decision thresholds derived using temporally-shifted non-members end up testing for temporal shift rather than membership.
We note that distinguishing between members and temporally shifted non-members is a realistic inference game with practical implications but differs from the classical MI game as temporally-shifted data may belong to a different distribution.  Furthermore, as we aren't able to reproduce the experimental settings of  prior works~\citep{shi2023detecting, meeus2023did} for distributional shift analysis, it is inconclusive if their differing results are solely due to the temporal shift between member/non-member samples (see \Cref{subsubsec:appendix-temporal-benchmark-details} for details).

\section{Revisiting Membership} \label{app:edit_distance_members}

\begin{figure*}[!t]
\begin{center}
\includegraphics[width=\textwidth]{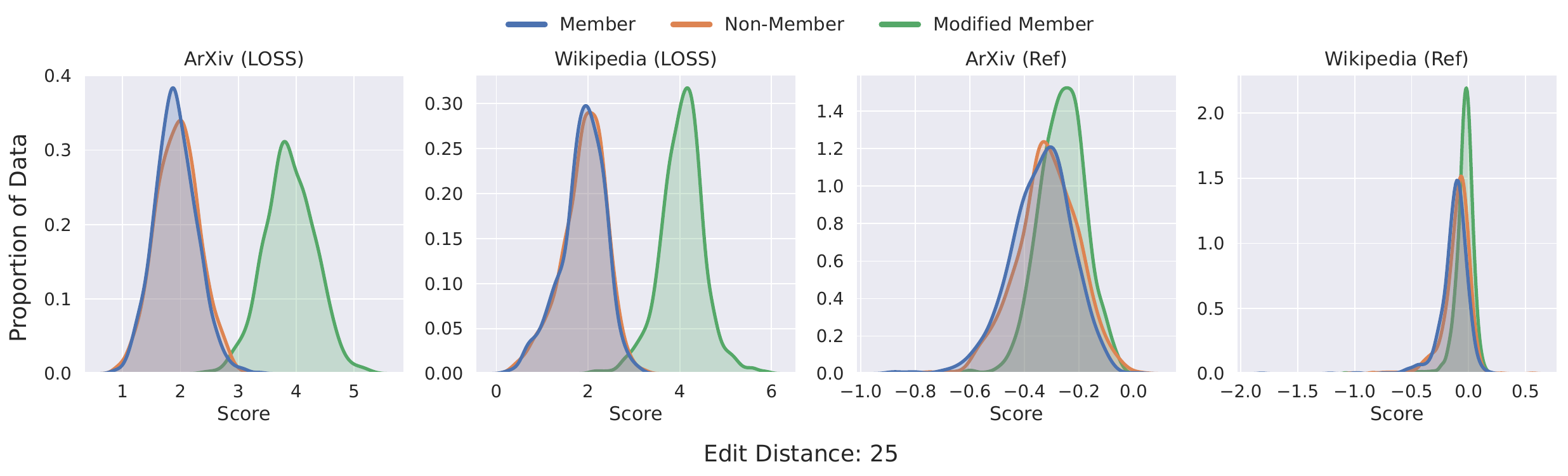}

\includegraphics[width=\textwidth]
{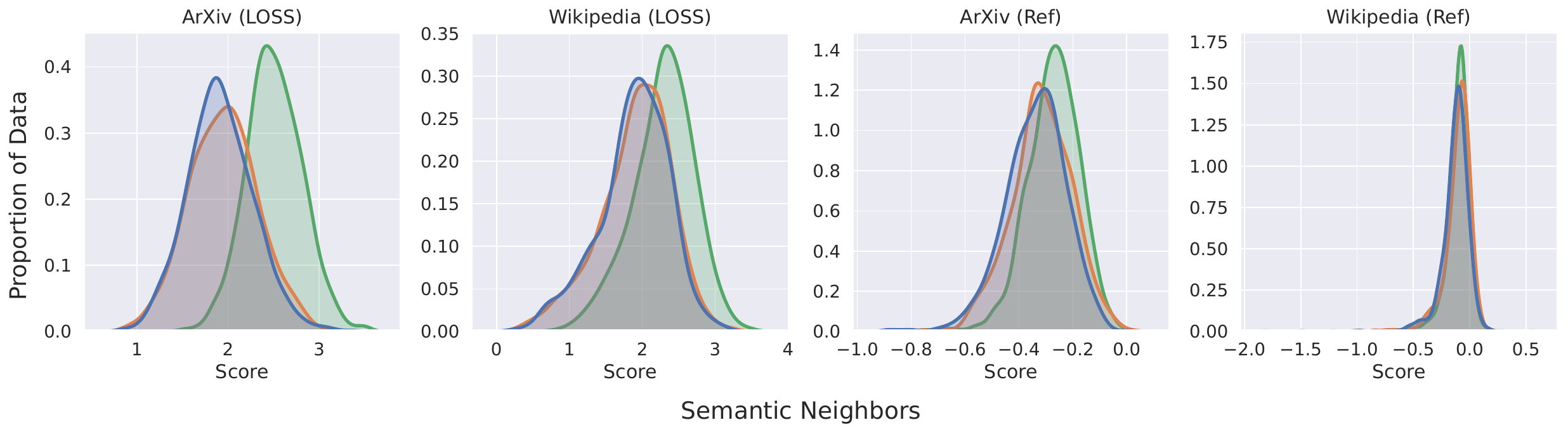}
\vspace{-.5em}
\caption{Distribution of scores for LOSS and Reference-based attacks for members, non-members, and modified members across ArXiv and Wikipedia domains. 
(Top) Modified members generated by random token replacement for edit distance 25. 
(Bottom) Modified members generated by replacing 5\% of tokens with semantically similar tokens.
}
\label{fig:score_distributions_edit_members}
\end{center}\vspace{-.5em}
\end{figure*}

The discussion of high \ngram\ overlap of non-members with members raises the question of whether exact membership is always useful concerning information leakage. The definition of membership in the standard MI game treats only records seen exactly during training as members, \eg for language models, substrings appearing exactly in the training corpus. However, this may be at odds with what adversaries and privacy auditors care about when concerning information leakage. For generative models especially, guessing the membership of some sample via other sufficiently close samples can be useful. For example, any paraphrase of ``Product launch in Q2, 2025" may be relevant as long as it preserves information regarding the product launch timeline, even if the paraphrase has a significant lexical difference, \eg ``Launching product in Q2, 2025" or ``Q2 2025 release".
Note that standard notions of Differential Privacy~\citep{dwork2006dp} do not immediately protect against such cases, since records being tested for membership are not, in the literal sense, members.
We explore two methods of constructing such ``sufficiently close" samples as an initial step in studying approximate membership definitions. %

\shortsection{Lexical Distance} We first experiment with creating modified member samples by replacing $n$ random tokens in a given sample with tokens randomly sampled from the model's vocabulary. We do so for $n = \{1,10,25\}$ (20 trials per $n$) and visualize the distribution for MIA scores using LOSS and Reference-based attacks (\Cref{fig:score_distributions_edit_members}, top). The LOSS attack yields distinct loss distributions between the modified members and original member/non-members, suggesting that the model is sensitive to out-of-place random tokens even for lightly perturbed member samples. The Reference-based attack, on the other hand, has a distribution of modified members much closer to both members and non-members, likely due to the reference model calibration accounting for the complexity introduced by the random tokens. This further reinforces the ambiguity of such samples---should they be considered members or non-members?

\begin{table}[!t]
\begin{center} 
\scriptsize %
\begin{tabularx}{.48\columnwidth}{l *{9}{>{\centering\arraybackslash}X}}
    \toprule
    \multirow{2}{*}{Domain} & \multicolumn{3}{c}{1\%} & \multicolumn{3}{c}{5\%}& \multicolumn{3}{c}{10\%}\\
    \cmidrule(lr){2-4}  \cmidrule(lr){5-7}  \cmidrule(lr){8-10} 
    & 1 & 10 & 25 &  1 & 10 & 25 &  1 & 10 & 25 \\
    \midrule
    ArXiv     & 0.1 & 0.0 & 0.0 & 0.3 & 0.1 & 0.1 & 0.7 & 0.3 & 0.2 \\ 
    Wikipedia & 0.0 & 0.0 & 0.0 & 0.2 & 0.1 & 0.1 & 0.6 & 0.4 & 0.1 \\
    \bottomrule
\end{tabularx}
\hspace{1em}
\begin{tabularx}{.45\columnwidth}{l *{6}{>{\centering\arraybackslash}X}@{}}
    \toprule
    \multirow{2}{*}{Domain} & \multicolumn{3}{c}{LOSS} & \multicolumn{3}{c}{Ref} \\
    \cmidrule(lr){2-4}  \cmidrule(lr){5-7}
    & 1\% & 5\% & 10\% & 1\% & 5\% & 10\% \\
    \midrule
    ArXiv     & 0.0 & 0.8 & 2.5 & 0.7 & 1.9 & 4.0\\
    Wikipedia & 0.0 & 0.5 & 2.3 & 0.4 & 3.0 & 8.2\\
    \bottomrule
\end{tabularx}
\end{center}
\vspace{-.5em}
\caption{FPR (\%) on modified members (treated as non-members) when using a score threshold that achieves a 1, 5, or 10\% FPR on the original member and non-member data for ArXiv and Wikipedia domains. (Left) Results for lexically similar modified members at edit-distances $n=\{1, 10, 25\}$. Reference-based attack is shown. For LOSS attack, all FPR values are 0 across all tested FPR thresholds and values of $n$. (Right) Results for semantically close modified members. LOSS and Reference-based attack reported.}\label{tab:edit_members_tprs}
\vspace{-.5em}
\end{table}

We also compute the thresholds corresponding to certain FPRs for actual member and non-member data and use these thresholds to compute the FPR on the modified members. We consider these modified members as ``non-members", which they are with regards to exact match\footnote{We perform token replacements at random with random tokens, so it is unlikely that these edited members are also actually members.}.
\Cref{tab:edit_members_tprs} (left) shows that these modified members have extremely low FPRs for edit distance as low as $n=1$, suggesting that these samples would be classified as non-members by the MIA, even though from the perspective of information leakage such a sample is effectively a member. We highlight the need to rethink membership for samples with extremely low lexical distance from actual training members, though even membership at higher lexical distances is important with respect to what information is still leaked in the unperturbed portions.

\shortsection{Semantic Distance} 
While a small edit distance suggests closeness in meaning, a higher edit distance does not necessarily imply loss of semantics. We compute MIA scores for neighbors generated for member samples as part of the Neighborhood attack (see \Cref{subsec:appendix-attack-details}) for the Wikipedia and ArXiv benchmarks and repeat the above pipeline. Visualizing the scores shows how the modified members are not too far from original member score distributions, especially for the Reference attack (\Cref{fig:score_distributions_edit_members}, bottom). We repeat the same FPR experiment as edit-distance-based modified members. While the FPR for these semantically close records is noticeably higher than records close by edit distance, the false positive rates are still low (\Cref{tab:edit_members_tprs}, right). Again, these results suggest that semantically close members would be classified as non-members even though they may be as useful as actual members depending on the inference goal and the semantic information preserved.

While it is not surprising that semantically close neighbors have MIA scores more similar to actual member samples than randomly-replaced tokens, it is clear that an ideal distance function should combine the benefits of lexical distance and semantics in defining a membership neighborhood. 
Such observations also motivate a fully semantic MI game, where a neighbor member may be defined by its proximity to an exact member in a semantic embedding space. This may provide a clearer interpretation of knowledge leakage than lexical matching, especially when samples naturally have high lexical (i.e., \ngram) overlap. %

\section{Conclusion}\label{sec:concl}

We shed light on the difficulty of membership inference against LLMs from the lens of an adversary. Our results suggest two possibilities: (1) data does not leave much of an imprint, owing to characteristics of the pre-training process at scale, such as large datasets and single-epoch training, and (2) the similarity between in and out members (which we demonstrate via \ngram\ overlap), coupled with huge datasets, makes this distinction fuzzy.
While MIA performance could improve via stronger attacks \citep{casper2024blackbox}, the second factor requires rethinking the membership game itself.
Our empirical results suggest that both of these might be confounding factors while measuring leakage. The membership inference game needs to be extended for such generative models to better align with information leakage that adversaries and auditors may care about, such as user-level leakage \citep{kandpal2023user} and PII \citep{lukas2023analyzing}. %
In the meanwhile, special care should thus be taken to avoid unintentional distributional shifts while constructing non-members for MIA benchmark construction.

\section*{Acknowledgements}

We are grateful to Nicholas Carlini for comments on early drafts of this paper. This research is supported in part by DARPA SemaFor Program No. HR001120C0123, and grants from the National Science Foundation through the AI Institute for Agent-based Cyber Threat Intelligence and Operation (\#2229876), Center for Trustworthy Machine Learning (\#1804603) and the DARPA MCS program through NIWC Pacific (N66001-19-2-4031).

This work is also supported in part by the Office of the Director of National Intelligence (ODNI), Intelligence Advanced Research Projects Activity (IARPA), via the HIATUS Program contract \#2022-22072200004. The views and conclusions contained herein are those of the authors and should not be interpreted as necessarily representing the official policies, either expressed or implied, of ODNI, IARPA, or the U.S. Government. The U.S. Government is authorized to reproduce and distribute reprints for governmental purposes notwithstanding any copyright annotation therein.

\bibliography{colm2024_conference}

\begin{thebibliography}{65}
\providecommand{\natexlab}[1]{#1}
\providecommand{\url}[1]{\texttt{#1}}
\expandafter\ifx\csname urlstyle\endcsname\relax
  \providecommand{\doi}[1]{doi: #1}\else
  \providecommand{\doi}{doi: \begingroup \urlstyle{rm}\Url}\fi

\bibitem[Abadi et~al.(2016)Abadi, Chu, Goodfellow, McMahan, Mironov, Talwar, and Zhang]{Abadi_2016}
Martin Abadi, Andy Chu, Ian Goodfellow, H.~Brendan McMahan, Ilya Mironov, Kunal Talwar, and Li~Zhang.
\newblock Deep learning with differential privacy.
\newblock In \emph{Proceedings of the 2016 ACM SIGSAC Conference on Computer and Communications Security}, CCS’16, 2016.

\bibitem[Andonian et~al.(2023)Andonian, Anthony, Biderman, Black, Gali, Gao, Hallahan, Levy-Kramer, Leahy, Nestler, Parker, Pieler, Phang, Purohit, Schoelkopf, Stander, Songz, Tigges, Thérien, Wang, and Weinbach]{gpt-neox-library}
Alex Andonian, Quentin Anthony, Stella Biderman, Sid Black, Preetham Gali, Leo Gao, Eric Hallahan, Josh Levy-Kramer, Connor Leahy, Lucas Nestler, Kip Parker, Michael Pieler, Jason Phang, Shivanshu Purohit, Hailey Schoelkopf, Dashiell Stander, Tri Songz, Curt Tigges, Benjamin Thérien, Phil Wang, and Samuel Weinbach.
\newblock {GPT-NeoX: Large Scale Autoregressive Language Modeling in PyTorch}, 2023.
\newblock URL \url{https://www.github.com/eleutherai/gpt-neox}.

\bibitem[Bertran et~al.(2023)Bertran, Tang, Roth, Kearns, Morgenstern, and Wu]{bertran2023scalable}
Martin~Andres Bertran, Shuai Tang, Aaron Roth, Michael Kearns, Jamie~Heather Morgenstern, and Steven Wu.
\newblock Scalable membership inference attacks via quantile regression.
\newblock In \emph{Advances in Neural Information Processing Systems}, 2023.

\bibitem[Biderman et~al.(2023{\natexlab{a}})Biderman, Prashanth, Sutawika, Schoelkopf, Anthony, Purohit, and Raf]{biderman2023emergent}
Stella Biderman, USVSN~Sai Prashanth, Lintang Sutawika, Hailey Schoelkopf, Quentin Anthony, Shivanshu Purohit, and Edward Raf.
\newblock Emergent and predictable memorization in large language models.
\newblock In \emph{Advances in Neural Information Processing Systems}, 2023{\natexlab{a}}.

\bibitem[Biderman et~al.(2023{\natexlab{b}})Biderman, Schoelkopf, Anthony, Bradley, O'Brien, Hallahan, Khan, Purohit, Prashanth, Raff, et~al.]{biderman2023pythia}
Stella Biderman, Hailey Schoelkopf, Quentin Anthony, Herbie Bradley, Kyle O'Brien, Eric Hallahan, Mohammad~Aflah Khan, Shivanshu Purohit, USVSN~Sai Prashanth, Edward Raff, et~al.
\newblock {Pythia: A Suite for Analyzing Large Language Models Across Training and Scaling}.
\newblock In \emph{International Conference on Learning Representations}, 2023{\natexlab{b}}.

\bibitem[Black et~al.(2021)Black, Leo, Wang, Leahy, and Biderman]{gpt-neo}
Sid Black, Gao Leo, Phil Wang, Connor Leahy, and Stella Biderman.
\newblock {GPT-Neo: Large Scale Autoregressive Language Modeling with Mesh-Tensorflow}, 2021.
\newblock URL \url{http://github.com/eleutherai/gpt-neo}.

\bibitem[Brown et~al.(2020)Brown, Mann, Ryder, Subbiah, Kaplan, Dhariwal, Neelakantan, Shyam, et~al.]{gpt3}
Tom~B. Brown, Benjamin Mann, Nick Ryder, Melanie Subbiah, Jared Kaplan, Prafulla Dhariwal, Arvind Neelakantan, Pranav Shyam, et~al.
\newblock {Language Models are Few-Shot Learners}.
\newblock In \emph{Advances in Neural Information Processing Systems}, 2020.

\bibitem[Carlini et~al.(2019)Carlini, Liu, Erlingsson, Kos, and Song]{carlini2019secret}
Nicholas Carlini, Chang Liu, {\'U}lfar Erlingsson, Jernej Kos, and Dawn Song.
\newblock {The Secret Sharer: Evaluating and Testing Unintended Memorization in Neural Networks}.
\newblock In \emph{USENIX Security Symposium}, 2019.

\bibitem[Carlini et~al.(2021)Carlini, Tramer, Wallace, Jagielski, Herbert-Voss, Lee, Roberts, Brown, Song, Erlingsson, et~al.]{carlini2021extracting}
Nicholas Carlini, Florian Tramer, Eric Wallace, Matthew Jagielski, Ariel Herbert-Voss, Katherine Lee, Adam Roberts, Tom~B Brown, Dawn Song, Ulfar Erlingsson, et~al.
\newblock {Extracting Training Data from Large Language Models}.
\newblock In \emph{USENIX Security Symposium}, 2021.

\bibitem[Carlini et~al.(2022)Carlini, Chien, Nasr, Song, Terzis, and Tramer]{carlini2022membership}
Nicholas Carlini, Steve Chien, Milad Nasr, Shuang Song, Andreas Terzis, and Florian Tramer.
\newblock {Membership Inference Attacks From First Principles}.
\newblock In \emph{IEEE Symposium on Security and Privacy}, 2022.

\bibitem[Carlini et~al.(2023)Carlini, Ippolito, Jagielski, Lee, Tramer, and Zhang]{carlini2023quantifying}
Nicholas Carlini, Daphne Ippolito, Matthew Jagielski, Katherine Lee, Florian Tramer, and Chiyuan Zhang.
\newblock Quantifying memorization across neural language models.
\newblock In \emph{International Conference on Learning Representations}, 2023.

\bibitem[Casper et~al.(2024)Casper, Ezell, Siegmann, Kolt, Curtis, Bucknall, Haupt, Wei, Scheurer, Hobbhahn, Sharkey, Krishna, Hagen, Alberti, Chan, Sun, Gerovitch, Bau, Tegmark, Krueger, and Hadfield-Menell]{casper2024blackbox}
Stephen Casper, Carson Ezell, Charlotte Siegmann, Noam Kolt, Taylor~Lynn Curtis, Benjamin Bucknall, Andreas Haupt, Kevin Wei, Jérémy Scheurer, Marius Hobbhahn, Lee Sharkey, Satyapriya Krishna, Marvin~Von Hagen, Silas Alberti, Alan Chan, Qinyi Sun, Michael Gerovitch, David Bau, Max Tegmark, David Krueger, and Dylan Hadfield-Menell.
\newblock Black-box access is insufficient for rigorous ai audits.
\newblock \emph{arXiv preprint arXiv:2401.14446}, 2024.

\bibitem[Chen et~al.(2022)Chen, Yu, and Fritz]{chen2022relaxloss}
Dingfan Chen, Ning Yu, and Mario Fritz.
\newblock Relaxloss: Defending membership inference attacks without losing utility.
\newblock In \emph{International Conference on Learning Representations}, 2022.

\bibitem[Cummings et~al.(2024)Cummings, Desfontaines, Evans, Geambasu, Huang, Rogers, Jagielski, Kairouz, Kamath, Oh, Ohrimenko, Papernot, Shen, Song, Su, Terzis, Thakurta, Vassilvitskii, Wang, Xiong, Yekhanin, Yu, Zhang, and Zhang]{cummings2024hdsr}
Rachel Cummings, Damien Desfontaines, David Evans, Roxana Geambasu, Yangsibo Huang, Ryan Rogers, Matthew Jagielski, Peter Kairouz, Gautam Kamath, Sewoong Oh, Olga Ohrimenko, Nicolas Papernot, Milan Shen, Shuang Song, Weijie Su, Andreas Terzis, Abhradeep Thakurta, Sergei Vassilvitskii, Yu-Xiang Wang, Li~Xiong, Sergey Yekhanin, Da~Yu, Huanyu Zhang, and Wanrong Zhang.
\newblock Advancing differential privacy: Where we are now and future directions for real-world deployment.
\newblock \emph{Harvard Data Science Review}, Jan 2024.
\newblock URL \url{https://doi.org/10.1162/99608f92.d3197524}.

\bibitem[Devlin et~al.(2019)Devlin, Chang, Lee, and Toutanova]{devlin2019bert}
Jacob Devlin, Ming-Wei Chang, Kenton Lee, and Kristina Toutanova.
\newblock Bert: Pre-training of deep bidirectional transformers for language understanding, 2019.

\bibitem[Dwork et~al.(2006)Dwork, McSherry, Nissim, and Smith]{dwork2006dp}
Cynthia Dwork, Frank McSherry, Kobbi Nissim, and Adam Smith.
\newblock Calibrating noise to sensitivity in private data analysis.
\newblock In \emph{Proceedings of the Third Conference on Theory of Cryptography}, 2006.

\bibitem[Fu et~al.(2023)Fu, Wang, Gao, Liu, Li, and Jiang]{fu2023practical}
Wenjie Fu, Huandong Wang, Chen Gao, Guanghua Liu, Yong Li, and Tao Jiang.
\newblock Practical membership inference attacks against fine-tuned large language models via self-prompt calibration, 2023.

\bibitem[Gao et~al.(2020)Gao, Biderman, Black, Golding, Hoppe, Foster, Phang, He, Thite, Nabeshima, et~al.]{gao2020pile}
Leo Gao, Stella Biderman, Sid Black, Laurence Golding, Travis Hoppe, Charles Foster, Jason Phang, Horace He, Anish Thite, Noa Nabeshima, et~al.
\newblock {The Pile: An 800GB Dataset of Diverse Text for Language Modeling}.
\newblock \emph{arXiv preprint arXiv:2101.00027}, 2020.

\bibitem[Groeneveld et~al.(2023)Groeneveld, Ha, and Magnusson]{bff2023}
Dirk Groeneveld, Chris Ha, and Ian Magnusson.
\newblock {BFF}: The big friendly filter, 2023.
\newblock URL \url{https://github.com/allenai/bff}.

\bibitem[Groeneveld et~al.(2024)Groeneveld, Beltagy, Walsh, Bhagia, Kinney, Tafjord, Jha, Ivison, Magnusson, Wang, Arora, Atkinson, Authur, Chandu, Cohan, Dumas, Elazar, Gu, Hessel, Khot, Merrill, Morrison, Muennighoff, Naik, Nam, Peters, Pyatkin, Ravichander, Schwenk, Shah, Smith, Subramani, Wortsman, Dasigi, Lambert, Richardson, Dodge, Lo, Soldaini, Smith, and Hajishirzi]{Groeneveld2023OLMo}
Dirk Groeneveld, Iz~Beltagy, Pete Walsh, Akshita Bhagia, Rodney Kinney, Oyvind Tafjord, Ananya~Harsh Jha, Hamish Ivison, Ian Magnusson, Yizhong Wang, Shane Arora, David Atkinson, Russell Authur, Khyathi Chandu, Arman Cohan, Jennifer Dumas, Yanai Elazar, Yuling Gu, Jack Hessel, Tushar Khot, William Merrill, Jacob Morrison, Niklas Muennighoff, Aakanksha Naik, Crystal Nam, Matthew~E. Peters, Valentina Pyatkin, Abhilasha Ravichander, Dustin Schwenk, Saurabh Shah, Will Smith, Nishant Subramani, Mitchell Wortsman, Pradeep Dasigi, Nathan Lambert, Kyle Richardson, Jesse Dodge, Kyle Lo, Luca Soldaini, Noah~A. Smith, and Hannaneh Hajishirzi.
\newblock Olmo: Accelerating the science of language models.
\newblock \emph{Preprint}, 2024.

\bibitem[Hoffmann et~al.(2022)Hoffmann, Borgeaud, Mensch, Buchatskaya, Cai, Rutherford, de~Las~Casas, Hendricks, Welbl, Clark, et~al.]{hoffmann2022training}
Jordan Hoffmann, Sebastian Borgeaud, Arthur Mensch, Elena Buchatskaya, Trevor Cai, Eliza Rutherford, Diego de~Las~Casas, Lisa~Anne Hendricks, Johannes Welbl, Aidan Clark, et~al.
\newblock An empirical analysis of compute-optimal large language model training.
\newblock In \emph{Advances in Neural Information Processing Systems}, 2022.

\bibitem[Jagielski et~al.(2023)Jagielski, Thakkar, Tramèr, Ippolito, Lee, Carlini, Wallace, Song, Thakurta, Papernot, and Zhang]{jagielski2023measuring}
Matthew Jagielski, Om~Thakkar, Florian Tramèr, Daphne Ippolito, Katherine Lee, Nicholas Carlini, Eric Wallace, Shuang Song, Abhradeep Thakurta, Nicolas Papernot, and Chiyuan Zhang.
\newblock Measuring forgetting of memorized training examples.
\newblock In \emph{International Conference on Learning Representations}, 2023.

\bibitem[Kandpal et~al.(2022)Kandpal, Wallace, and Raffel]{kandpal2022deduplicating}
Nikhil Kandpal, Eric Wallace, and Colin Raffel.
\newblock {Deduplicating Training Data Mitigates Privacy Risks in Language Models}.
\newblock In \emph{International Conference on Machine Learning}, 2022.

\bibitem[Kandpal et~al.(2023)Kandpal, Pillutla, Oprea, Kairouz, Choquette-Choo, and Xu]{kandpal2023user}
Nikhil Kandpal, Krishna Pillutla, Alina Oprea, Peter Kairouz, Christopher~A Choquette-Choo, and Zheng Xu.
\newblock User inference attacks on large language models.
\newblock \emph{arXiv preprint arXiv:2310.09266}, 2023.

\bibitem[Komatsuzaki(2019)]{komatsuzaki2019epoch}
Aran Komatsuzaki.
\newblock One epoch is all you need.
\newblock \emph{arXiv preprint arXiv:1906.06669}, 2019.

\bibitem[Lehman et~al.(2021)Lehman, Jain, Pichotta, Goldberg, and Wallace]{lehman2021does}
Eric Lehman, Sarthak Jain, Karl Pichotta, Yoav Goldberg, and Byron Wallace.
\newblock {Does {BERT} Pretrained on Clinical Notes Reveal Sensitive Data?}
\newblock In \emph{Proceedings of the 2021 Conference of the North American Chapter of the Association for Computational Linguistics: Human Language Technologies}, 2021.

\bibitem[Li et~al.(2023{\natexlab{a}})Li, Wang, Wang, and Neel]{li2023mope}
Marvin Li, Jason Wang, Jeffrey Wang, and Seth Neel.
\newblock Mope: Model perturbation-based privacy attacks on language models.
\newblock In \emph{Conference on Empirical Methods in Natural Language Processing}, 2023{\natexlab{a}}.

\bibitem[Li et~al.(2023{\natexlab{b}})Li, Allal, Zi, Muennighoff, Kocetkov, Mou, Marone, Akiki, Li, Chim, Liu, Zheltonozhskii, Zhuo, Wang, Dehaene, Davaadorj, Lamy-Poirier, Monteiro, Shliazhko, Gontier, Meade, Zebaze, Yee, Umapathi, Zhu, Lipkin, Oblokulov, Wang, Murthy, Stillerman, Patel, Abulkhanov, Zocca, Dey, Zhang, Fahmy, Bhattacharyya, Yu, Singh, Luccioni, Villegas, Kunakov, Zhdanov, Romero, Lee, Timor, Ding, Schlesinger, Schoelkopf, Ebert, Dao, Mishra, Gu, Robinson, Anderson, Dolan-Gavitt, Contractor, Reddy, Fried, Bahdanau, Jernite, Ferrandis, Hughes, Wolf, Guha, von Werra, and de~Vries]{li2023starcoder}
Raymond Li, Loubna~Ben Allal, Yangtian Zi, Niklas Muennighoff, Denis Kocetkov, Chenghao Mou, Marc Marone, Christopher Akiki, Jia Li, Jenny Chim, Qian Liu, Evgenii Zheltonozhskii, Terry~Yue Zhuo, Thomas Wang, Olivier Dehaene, Mishig Davaadorj, Joel Lamy-Poirier, João Monteiro, Oleh Shliazhko, Nicolas Gontier, Nicholas Meade, Armel Zebaze, Ming-Ho Yee, Logesh~Kumar Umapathi, Jian Zhu, Benjamin Lipkin, Muhtasham Oblokulov, Zhiruo Wang, Rudra Murthy, Jason Stillerman, Siva~Sankalp Patel, Dmitry Abulkhanov, Marco Zocca, Manan Dey, Zhihan Zhang, Nour Fahmy, Urvashi Bhattacharyya, Wenhao Yu, Swayam Singh, Sasha Luccioni, Paulo Villegas, Maxim Kunakov, Fedor Zhdanov, Manuel Romero, Tony Lee, Nadav Timor, Jennifer Ding, Claire Schlesinger, Hailey Schoelkopf, Jan Ebert, Tri Dao, Mayank Mishra, Alex Gu, Jennifer Robinson, Carolyn~Jane Anderson, Brendan Dolan-Gavitt, Danish Contractor, Siva Reddy, Daniel Fried, Dzmitry Bahdanau, Yacine Jernite, Carlos~Muñoz Ferrandis, Sean Hughes, Thomas Wolf, Arjun Guha, Leandro von
  Werra, and Harm de~Vries.
\newblock Starcoder: may the source be with you!
\newblock \emph{arXiv preprint arXiv:2305.06161}, 2023{\natexlab{b}}.

\bibitem[Li et~al.(2023{\natexlab{c}})Li, Guerin, and Lin]{li2023avoiding}
Yucheng Li, Frank Guerin, and Chenghua Lin.
\newblock Avoiding data contamination in language model evaluation: Dynamic test construction with latest materials, 2023{\natexlab{c}}.

\bibitem[Lukas et~al.(2023)Lukas, Salem, Sim, Tople, Wutschitz, and Zanella-B{\'e}guelin]{lukas2023analyzing}
Nils Lukas, Ahmed Salem, Robert Sim, Shruti Tople, Lukas Wutschitz, and Santiago Zanella-B{\'e}guelin.
\newblock {Analyzing Leakage of Personally Identifiable Information in Language Models}.
\newblock In \emph{IEEE Symposium on Security and Privacy}, 2023.

\bibitem[Magnusson et~al.(2023)Magnusson, Bhagia, Hofmann, Soldaini, Harsh~Jha, Tafjord, Schwenk, Walsh, Elazar, Lo, Groenveld, Beltagy, Hajishirz, Smith, Richardson, and Dodge]{paloma}
Ian Magnusson, Akshita Bhagia, Valentin Hofmann, Luca Soldaini, Ananya Harsh~Jha, Oyvind Tafjord, Dustin Schwenk, Evan~Pete Walsh, Yanai Elazar, Kyle Lo, Dirk Groenveld, Iz~Beltagy, Hanneneh Hajishirz, Noah~A. Smith, Kyle Richardson, and Jesse Dodge.
\newblock {Paloma}: A benchmark for evaluating language model fit.
\newblock \emph{technical report}, 2023.
\newblock URL \url{https://paloma.allen.ai/}.

\bibitem[Mattern et~al.(2023)Mattern, Mireshghallah, Jin, Schoelkopf, Sachan, and Berg-Kirkpatrick]{mattern2023membership}
Justus Mattern, Fatemehsadat Mireshghallah, Zhijing Jin, Bernhard Schoelkopf, Mrinmaya Sachan, and Taylor Berg-Kirkpatrick.
\newblock {Membership Inference Attacks against Language Models via Neighbourhood Comparison}.
\newblock In \emph{Findings of the Association for Computational Linguistics (ACL)}, 2023.

\bibitem[Meeus et~al.(2023)Meeus, Jain, Rei, and de~Montjoye]{meeus2023did}
Matthieu Meeus, Shubham Jain, Marek Rei, and Yves-Alexandre de~Montjoye.
\newblock Did the neurons read your book? document-level membership inference for large language models.
\newblock \emph{arXiv preprint arXiv:2310.15007}, 2023.

\bibitem[Min et~al.(2023)Min, Gururangan, Wallace, Hajishirzi, Smith, and Zettlemoyer]{silo}
Sewon Min, Suchin Gururangan, Eric Wallace, Hannaneh Hajishirzi, Noah~A. Smith, and Luke Zettlemoyer.
\newblock {{SILO} Language Models: Isolating Legal Risk in a Nonparametric Datastore}.
\newblock \emph{Regulatable ML Workshop at NeurIPS}, 2023.

\bibitem[Mireshghallah et~al.(2022{\natexlab{a}})Mireshghallah, Goyal, Uniyal, Berg-Kirkpatrick, and Shokri]{mireshghallah2022quantifying}
Fatemehsadat Mireshghallah, Kartik Goyal, Archit Uniyal, Taylor Berg-Kirkpatrick, and Reza Shokri.
\newblock {Quantifying Privacy Risks of Masked Language Models Using Membership Inference Attacks}.
\newblock In \emph{Conference on Empirical Methods in Natural Language Processing}, 2022{\natexlab{a}}.

\bibitem[Mireshghallah et~al.(2022{\natexlab{b}})Mireshghallah, Uniyal, Wang, Evans, and Berg-Kirkpatrick]{mireshghallah-etal-2022-empirical}
Fatemehsadat Mireshghallah, Archit Uniyal, Tianhao Wang, David Evans, and Taylor Berg-Kirkpatrick.
\newblock {An Empirical Analysis of Memorization in Fine-tuned Autoregressive Language Models}.
\newblock In \emph{Conference on Empirical Methods in Natural Language Processing}, 2022{\natexlab{b}}.

\bibitem[Muennighoff et~al.(2023)Muennighoff, Rush, Barak, Scao, Piktus, Tazi, Pyysalo, Wolf, and Raffel]{muennighoff2023scaling}
Niklas Muennighoff, Alexander~M Rush, Boaz Barak, Teven~Le Scao, Aleksandra Piktus, Nouamane Tazi, Sampo Pyysalo, Thomas Wolf, and Colin Raffel.
\newblock Scaling data-constrained language models.
\newblock In \emph{Advances in Neural Information Processing Systems}, 2023.

\bibitem[Nakamura et~al.(2020)Nakamura, Hanaoka, Nomura, Hayashi, Abe, Yada, Wakamiya, and Aramaki]{nakamura2020kart}
Yuta Nakamura, Shouhei Hanaoka, Yukihiro Nomura, Naoto Hayashi, Osamu Abe, Shuntaro Yada, Shoko Wakamiya, and Eiji Aramaki.
\newblock {KART: Parameterization of Privacy Leakage Scenarios from Pre-trained Language Models}.
\newblock \emph{arXiv preprint arXiv:2101.00036}, 2020.

\bibitem[Nakkiran et~al.(2021)Nakkiran, Kaplun, Bansal, Yang, Barak, and Sutskever]{nakkiran2019deep}
Preetum Nakkiran, Gal Kaplun, Yamini Bansal, Tristan Yang, Boaz Barak, and Ilya Sutskever.
\newblock Deep double descent: Where bigger models and more data hurt.
\newblock \emph{Journal of Statistical Mechanics: Theory and Experiment}, 2021\penalty0 (12), 2021.

\bibitem[OpenAI et~al.(2024)OpenAI, Achiam, Adler, Agarwal, Ahmad, Akkaya, Aleman, Almeida, Altenschmidt, Altman, Anadkat, Avila, Babuschkin, Balaji, Balcom, Baltescu, Bao, Bavarian, Belgum, Bello, Berdine, Bernadett-Shapiro, Berner, Bogdonoff, Boiko, Boyd, Brakman, Brockman, Brooks, Brundage, Button, Cai, Campbell, Cann, Carey, Carlson, Carmichael, Chan, Chang, Chantzis, Chen, Chen, Chen, Chen, Chen, Chess, Cho, Chu, Chung, Cummings, Currier, Dai, Decareaux, Degry, Deutsch, Deville, Dhar, Dohan, Dowling, Dunning, Ecoffet, Eleti, Eloundou, Farhi, Fedus, Felix, Fishman, Forte, Fulford, Gao, Georges, Gibson, Goel, Gogineni, Goh, Gontijo-Lopes, Gordon, Grafstein, Gray, Greene, Gross, Gu, Guo, Hallacy, Han, Harris, He, Heaton, Heidecke, Hesse, Hickey, Hickey, Hoeschele, Houghton, Hsu, Hu, Hu, Huizinga, Jain, Jain, Jang, Jiang, Jiang, Jin, Jin, Jomoto, Jonn, Jun, Kaftan, Łukasz Kaiser, Kamali, Kanitscheider, Keskar, Khan, Kilpatrick, Kim, Kim, Kim, Kirchner, Kiros, Knight, Kokotajlo, Łukasz Kondraciuk, Kondrich,
  Konstantinidis, Kosic, Krueger, Kuo, Lampe, Lan, Lee, Leike, Leung, Levy, Li, Lim, Lin, Lin, Litwin, Lopez, Lowe, Lue, Makanju, Malfacini, Manning, Markov, Markovski, Martin, Mayer, Mayne, McGrew, McKinney, McLeavey, McMillan, McNeil, Medina, Mehta, Menick, Metz, Mishchenko, Mishkin, Monaco, Morikawa, Mossing, Mu, Murati, Murk, Mély, Nair, Nakano, Nayak, Neelakantan, Ngo, Noh, Ouyang, O'Keefe, Pachocki, Paino, Palermo, Pantuliano, Parascandolo, Parish, Parparita, Passos, Pavlov, Peng, Perelman, de~Avila Belbute~Peres, Petrov, de~Oliveira~Pinto, Michael, Pokorny, Pokrass, Pong, Powell, Power, Power, Proehl, Puri, Radford, Rae, Ramesh, Raymond, Real, Rimbach, Ross, Rotsted, Roussez, Ryder, Saltarelli, Sanders, Santurkar, Sastry, Schmidt, Schnurr, Schulman, Selsam, Sheppard, Sherbakov, Shieh, Shoker, Shyam, Sidor, Sigler, Simens, Sitkin, Slama, Sohl, Sokolowsky, Song, Staudacher, Such, Summers, Sutskever, Tang, Tezak, Thompson, Tillet, Tootoonchian, Tseng, Tuggle, Turley, Tworek, Uribe, Vallone, Vijayvergiya,
  Voss, Wainwright, Wang, Wang, Wang, Ward, Wei, Weinmann, Welihinda, Welinder, Weng, Weng, Wiethoff, Willner, Winter, Wolrich, Wong, Workman, Wu, Wu, Wu, Xiao, Xu, Yoo, Yu, Yuan, Zaremba, Zellers, Zhang, Zhang, Zhao, Zheng, Zhuang, Zhuk, and Zoph]{openai2024gpt4}
OpenAI, Josh Achiam, Steven Adler, Sandhini Agarwal, Lama Ahmad, Ilge Akkaya, Florencia~Leoni Aleman, Diogo Almeida, Janko Altenschmidt, Sam Altman, Shyamal Anadkat, Red Avila, Igor Babuschkin, Suchir Balaji, Valerie Balcom, Paul Baltescu, Haiming Bao, Mohammad Bavarian, Jeff Belgum, Irwan Bello, Jake Berdine, Gabriel Bernadett-Shapiro, Christopher Berner, Lenny Bogdonoff, Oleg Boiko, Madelaine Boyd, Anna-Luisa Brakman, Greg Brockman, Tim Brooks, Miles Brundage, Kevin Button, Trevor Cai, Rosie Campbell, Andrew Cann, Brittany Carey, Chelsea Carlson, Rory Carmichael, Brooke Chan, Che Chang, Fotis Chantzis, Derek Chen, Sully Chen, Ruby Chen, Jason Chen, Mark Chen, Ben Chess, Chester Cho, Casey Chu, Hyung~Won Chung, Dave Cummings, Jeremiah Currier, Yunxing Dai, Cory Decareaux, Thomas Degry, Noah Deutsch, Damien Deville, Arka Dhar, David Dohan, Steve Dowling, Sheila Dunning, Adrien Ecoffet, Atty Eleti, Tyna Eloundou, David Farhi, Liam Fedus, Niko Felix, Simón~Posada Fishman, Juston Forte, Isabella Fulford, Leo
  Gao, Elie Georges, Christian Gibson, Vik Goel, Tarun Gogineni, Gabriel Goh, Rapha Gontijo-Lopes, Jonathan Gordon, Morgan Grafstein, Scott Gray, Ryan Greene, Joshua Gross, Shixiang~Shane Gu, Yufei Guo, Chris Hallacy, Jesse Han, Jeff Harris, Yuchen He, Mike Heaton, Johannes Heidecke, Chris Hesse, Alan Hickey, Wade Hickey, Peter Hoeschele, Brandon Houghton, Kenny Hsu, Shengli Hu, Xin Hu, Joost Huizinga, Shantanu Jain, Shawn Jain, Joanne Jang, Angela Jiang, Roger Jiang, Haozhun Jin, Denny Jin, Shino Jomoto, Billie Jonn, Heewoo Jun, Tomer Kaftan, Łukasz Kaiser, Ali Kamali, Ingmar Kanitscheider, Nitish~Shirish Keskar, Tabarak Khan, Logan Kilpatrick, Jong~Wook Kim, Christina Kim, Yongjik Kim, Jan~Hendrik Kirchner, Jamie Kiros, Matt Knight, Daniel Kokotajlo, Łukasz Kondraciuk, Andrew Kondrich, Aris Konstantinidis, Kyle Kosic, Gretchen Krueger, Vishal Kuo, Michael Lampe, Ikai Lan, Teddy Lee, Jan Leike, Jade Leung, Daniel Levy, Chak~Ming Li, Rachel Lim, Molly Lin, Stephanie Lin, Mateusz Litwin, Theresa Lopez, Ryan
  Lowe, Patricia Lue, Anna Makanju, Kim Malfacini, Sam Manning, Todor Markov, Yaniv Markovski, Bianca Martin, Katie Mayer, Andrew Mayne, Bob McGrew, Scott~Mayer McKinney, Christine McLeavey, Paul McMillan, Jake McNeil, David Medina, Aalok Mehta, Jacob Menick, Luke Metz, Andrey Mishchenko, Pamela Mishkin, Vinnie Monaco, Evan Morikawa, Daniel Mossing, Tong Mu, Mira Murati, Oleg Murk, David Mély, Ashvin Nair, Reiichiro Nakano, Rajeev Nayak, Arvind Neelakantan, Richard Ngo, Hyeonwoo Noh, Long Ouyang, Cullen O'Keefe, Jakub Pachocki, Alex Paino, Joe Palermo, Ashley Pantuliano, Giambattista Parascandolo, Joel Parish, Emy Parparita, Alex Passos, Mikhail Pavlov, Andrew Peng, Adam Perelman, Filipe de~Avila Belbute~Peres, Michael Petrov, Henrique~Ponde de~Oliveira~Pinto, Michael, Pokorny, Michelle Pokrass, Vitchyr~H. Pong, Tolly Powell, Alethea Power, Boris Power, Elizabeth Proehl, Raul Puri, Alec Radford, Jack Rae, Aditya Ramesh, Cameron Raymond, Francis Real, Kendra Rimbach, Carl Ross, Bob Rotsted, Henri Roussez,
  Nick Ryder, Mario Saltarelli, Ted Sanders, Shibani Santurkar, Girish Sastry, Heather Schmidt, David Schnurr, John Schulman, Daniel Selsam, Kyla Sheppard, Toki Sherbakov, Jessica Shieh, Sarah Shoker, Pranav Shyam, Szymon Sidor, Eric Sigler, Maddie Simens, Jordan Sitkin, Katarina Slama, Ian Sohl, Benjamin Sokolowsky, Yang Song, Natalie Staudacher, Felipe~Petroski Such, Natalie Summers, Ilya Sutskever, Jie Tang, Nikolas Tezak, Madeleine~B. Thompson, Phil Tillet, Amin Tootoonchian, Elizabeth Tseng, Preston Tuggle, Nick Turley, Jerry Tworek, Juan Felipe~Cerón Uribe, Andrea Vallone, Arun Vijayvergiya, Chelsea Voss, Carroll Wainwright, Justin~Jay Wang, Alvin Wang, Ben Wang, Jonathan Ward, Jason Wei, CJ~Weinmann, Akila Welihinda, Peter Welinder, Jiayi Weng, Lilian Weng, Matt Wiethoff, Dave Willner, Clemens Winter, Samuel Wolrich, Hannah Wong, Lauren Workman, Sherwin Wu, Jeff Wu, Michael Wu, Kai Xiao, Tao Xu, Sarah Yoo, Kevin Yu, Qiming Yuan, Wojciech Zaremba, Rowan Zellers, Chong Zhang, Marvin Zhang, Shengjia
  Zhao, Tianhao Zheng, Juntang Zhuang, William Zhuk, and Barret Zoph.
\newblock Gpt-4 technical report, 2024.

\bibitem[Oren et~al.(2023)Oren, Meister, Chatterji, Ladhak, and Hashimoto]{oren2023proving}
Yonatan Oren, Nicole Meister, Niladri Chatterji, Faisal Ladhak, and Tatsunori~B Hashimoto.
\newblock Proving test set contamination in black box language models.
\newblock \emph{arXiv preprint arXiv:2310.17623}, 2023.

\bibitem[Radford et~al.(2019)Radford, Wu, Child, Luan, Amodei, Sutskever, et~al.]{radford2019language}
Alec Radford, Jeffrey Wu, Rewon Child, David Luan, Dario Amodei, Ilya Sutskever, et~al.
\newblock {Language Models are Unsupervised Multitask Learners}.
\newblock \emph{OpenAI blog}, 2019.

\bibitem[Raffel et~al.(2019)Raffel, Shazeer, Roberts, Lee, Narang, Matena, Zhou, Li, and Liu]{2019t5}
Colin Raffel, Noam Shazeer, Adam Roberts, Katherine Lee, Sharan Narang, Michael Matena, Yanqi Zhou, Wei Li, and Peter~J. Liu.
\newblock Exploring the limits of transfer learning with a unified text-to-text transformer.
\newblock \emph{arXiv e-prints}, 2019.

\bibitem[Sablayrolles et~al.(2019)Sablayrolles, Douze, Schmid, Ollivier, and J{\'e}gou]{sablayrolles2019white}
Alexandre Sablayrolles, Matthijs Douze, Cordelia Schmid, Yann Ollivier, and Herv{\'e} J{\'e}gou.
\newblock White-box vs black-box: Bayes optimal strategies for membership inference.
\newblock In \emph{International Conference on Machine Learning}, 2019.

\bibitem[Sanh et~al.(2019)Sanh, Debut, Chaumond, and Wolf]{sanh2019distilbert}
Victor Sanh, Lysandre Debut, Julien Chaumond, and Thomas Wolf.
\newblock Distilbert, a distilled version of bert: smaller, faster, cheaper and lighter.
\newblock In \emph{NeurIPS $\text{EMC}^2$ Workshop}, 2019.

\bibitem[Shi et~al.(2023)Shi, Ajith, Xia, Huang, Liu, Blevins, Chen, and Zettlemoyer]{shi2023detecting}
Weijia Shi, Anirudh Ajith, Mengzhou Xia, Yangsibo Huang, Daogao Liu, Terra Blevins, Danqi Chen, and Luke Zettlemoyer.
\newblock Detecting pretraining data from large language models.
\newblock \emph{Regulatable ML Workshop at NeurIPS}, 2023.

\bibitem[Shokri(2022)]{shokri2022auditing}
Reza Shokri.
\newblock {Auditing Data Privacy for Machine Learning}.
\newblock In \emph{USENIX Security Symposium}, 2022.

\bibitem[Shokri et~al.(2017)Shokri, Stronati, Song, and Shmatikov]{shokri2017membership}
Reza Shokri, Marco Stronati, Congzheng Song, and Vitaly Shmatikov.
\newblock Membership inference attacks against machine learning models.
\newblock In \emph{IEEE Symposium on Security and Privacy}, 2017.

\bibitem[Soldaini \& Lo(2023)Soldaini and Lo]{peS2o}
Luca Soldaini and Kyle Lo.
\newblock {peS2o (Pretraining Efficiently on S2ORC) Dataset}.
\newblock Technical report, {Allen Institute for AI}, 2023.
\newblock ODC-By, \url{https://github.com/allenai/pes2o}.

\bibitem[Soldaini et~al.(2023)Soldaini, Kinney, Bhagia, Schwenk, Atkinson, Authur, Bogin, Chandu, Dumas, Elazar, Hofmann, Jha, Kumar, Lucy, Lyu, Magnusson, Morrison, Muennighoff, Naik, Nam, Peters, Ravichander, Richardson, Shen, Strubell, Subramani, Tafjord, Walsh, Hajishirzi, Smith, Zettlemoyer, Beltagy, Groeneveld, Dodge, and Lo]{dolma}
Luca Soldaini, Rodney Kinney, Akshita Bhagia, Dustin Schwenk, David Atkinson, Russell Authur, Ben Bogin, Khyathi Chandu, Jennifer Dumas, Yanai Elazar, Valentin Hofmann, Ananya~Harsh Jha, Sachin Kumar, Li~Lucy, Xinxi Lyu, Ian Magnusson, Jacob Morrison, Niklas Muennighoff, Aakanksha Naik, Crystal Nam, Matthew~E. Peters, Abhilasha Ravichander, Kyle Richardson, Zejiang Shen, Emma Strubell, Nishant Subramani, Oyvind Tafjord, Evan~Pete Walsh, Hannaneh Hajishirzi, Noah~A. Smith, Luke Zettlemoyer, Iz~Beltagy, Dirk Groeneveld, Jesse Dodge, and Kyle Lo.
\newblock {Dolma: An Open Corpus of Three Trillion Tokens for Language Model Pretraining Research}.
\newblock \emph{arXiv preprint}, 2023.

\bibitem[Steinke et~al.(2023)Steinke, Nasr, and Jagielski]{steinke2023privacy}
Thomas Steinke, Milad Nasr, and Matthew Jagielski.
\newblock Privacy auditing with one (1) training run.
\newblock In \emph{Advances in Neural Information Processing Systems}, 2023.

\bibitem[Suri \& Evans(2022)Suri and Evans]{suri2022formalizing}
Anshuman Suri and David Evans.
\newblock Formalizing and estimating distribution inference risks.
\newblock In \emph{Privacy Enhancing Technologies Symposium}, 2022.

\bibitem[Tang et~al.(2022)Tang, Mahloujifar, Song, Shejwalkar, Nasr, Houmansadr, and Mittal]{tang2021mitigating}
Xinyu Tang, Saeed Mahloujifar, Liwei Song, Virat Shejwalkar, Milad Nasr, Amir Houmansadr, and Prateek Mittal.
\newblock Mitigating membership inference attacks by self-distillation through a novel ensemble architecture.
\newblock In \emph{USENIX Security Symposium}, 2022.

\bibitem[Team(2023)]{MosaicML2023Introducing}
MosaicML~NLP Team.
\newblock Introducing mpt-7b: A new standard for open-source, commercially usable llms.
\newblock \url{www.mosaicml.com/blog/mpt-7b}, 2023.
\newblock Accessed: 2024-01-11.

\bibitem[Tirumala et~al.(2022)Tirumala, Markosyan, Zettlemoyer, and Aghajanyan]{tirumala2022memorization}
Kushal Tirumala, Aram Markosyan, Luke Zettlemoyer, and Armen Aghajanyan.
\newblock Memorization without overfitting: Analyzing the training dynamics of large language models.
\newblock In \emph{Advances in Neural Information Processing Systems}, 2022.

\bibitem[{Together AI}(2023)]{together2023redpajama}
{Together AI}.
\newblock Redpajama: an open dataset for training large language models, 2023.
\newblock URL \url{https://github.com/togethercomputer/RedPajama-Data}.

\bibitem[Touvron et~al.(2023{\natexlab{a}})Touvron, Lavril, Izacard, Martinet, Lachaux, Lacroix, Rozière, Goyal, Hambro, Azhar, Rodriguez, Joulin, Grave, and Lample]{touvron2023llama1}
Hugo Touvron, Thibaut Lavril, Gautier Izacard, Xavier Martinet, Marie-Anne Lachaux, Timothée Lacroix, Baptiste Rozière, Naman Goyal, Eric Hambro, Faisal Azhar, Aurelien Rodriguez, Armand Joulin, Edouard Grave, and Guillaume Lample.
\newblock Llama: Open and efficient foundation language models, 2023{\natexlab{a}}.

\bibitem[Touvron et~al.(2023{\natexlab{b}})Touvron, Martin, Stone, Albert, Almahairi, Babaei, Bashlykov, Batra, Bhargava, Bhosale, Bikel, Blecher, Ferrer, Chen, Cucurull, Esiobu, Fernandes, Fu, Fu, Fuller, Gao, Goswami, Goyal, Hartshorn, Hosseini, Hou, Inan, Kardas, Kerkez, Khabsa, Kloumann, Korenev, Koura, Lachaux, Lavril, Lee, Liskovich, Lu, Mao, Martinet, Mihaylov, Mishra, Molybog, Nie, Poulton, Reizenstein, Rungta, Saladi, Schelten, Silva, Smith, Subramanian, Tan, Tang, Taylor, Williams, Kuan, Xu, Yan, Zarov, Zhang, Fan, Kambadur, Narang, Rodriguez, Stojnic, Edunov, and Scialom]{touvron2023llama}
Hugo Touvron, Louis Martin, Kevin Stone, Peter Albert, Amjad Almahairi, Yasmine Babaei, Nikolay Bashlykov, Soumya Batra, Prajjwal Bhargava, Shruti Bhosale, Dan Bikel, Lukas Blecher, Cristian~Canton Ferrer, Moya Chen, Guillem Cucurull, David Esiobu, Jude Fernandes, Jeremy Fu, Wenyin Fu, Brian Fuller, Cynthia Gao, Vedanuj Goswami, Naman Goyal, Anthony Hartshorn, Saghar Hosseini, Rui Hou, Hakan Inan, Marcin Kardas, Viktor Kerkez, Madian Khabsa, Isabel Kloumann, Artem Korenev, Punit~Singh Koura, Marie-Anne Lachaux, Thibaut Lavril, Jenya Lee, Diana Liskovich, Yinghai Lu, Yuning Mao, Xavier Martinet, Todor Mihaylov, Pushkar Mishra, Igor Molybog, Yixin Nie, Andrew Poulton, Jeremy Reizenstein, Rashi Rungta, Kalyan Saladi, Alan Schelten, Ruan Silva, Eric~Michael Smith, Ranjan Subramanian, Xiaoqing~Ellen Tan, Binh Tang, Ross Taylor, Adina Williams, Jian~Xiang Kuan, Puxin Xu, Zheng Yan, Iliyan Zarov, Yuchen Zhang, Angela Fan, Melanie Kambadur, Sharan Narang, Aurelien Rodriguez, Robert Stojnic, Sergey Edunov, and Thomas
  Scialom.
\newblock Llama 2: Open foundation and fine-tuned chat models, 2023{\natexlab{b}}.

\bibitem[Tow(2023)]{StableLMAlphaV2Models}
Jonathan Tow.
\newblock Stablelm alpha v2 models, 2023.
\newblock URL \url{https://huggingface.co/stabilityai/stablelm-base-alpha-3b-v2}.

\bibitem[Watson et~al.(2022)Watson, Guo, Cormode, and Sablayrolles]{watson2022on}
Lauren Watson, Chuan Guo, Graham Cormode, and Alexandre Sablayrolles.
\newblock On the importance of difficulty calibration in membership inference attacks.
\newblock In \emph{International Conference on Learning Representations}, 2022.

\bibitem[Yang et~al.(2023)Yang, Chiang, Zheng, Gonzalez, and Stoica]{yang2023rethinking}
Shuo Yang, Wei-Lin Chiang, Lianmin Zheng, Joseph~E. Gonzalez, and Ion Stoica.
\newblock Rethinking benchmark and contamination for language models with rephrased samples, 2023.

\bibitem[Ye et~al.(2022)Ye, Maddi, Murakonda, Bindschaedler, and Shokri]{ye2022enhanced}
Jiayuan Ye, Aadyaa Maddi, Sasi~Kumar Murakonda, Vincent Bindschaedler, and Reza Shokri.
\newblock {Enhanced Membership Inference Attacks against Machine Learning Models}.
\newblock In \emph{The ACM Conference on Computer and Communications Security}, 2022.

\bibitem[Yeom et~al.(2018)Yeom, Giacomelli, Fredrikson, and Jha]{yeom2018privacy}
Samuel Yeom, Irene Giacomelli, Matt Fredrikson, and Somesh Jha.
\newblock {Privacy Risk in Machine Learning: Analyzing the Connection to Overfitting}.
\newblock In \emph{2018 IEEE 31st Computer Security Foundations Symposium (CSF)}, 2018.

\bibitem[Zarifzadeh et~al.(2023)Zarifzadeh, Liu, and Shokri]{zarifzadeh2023low}
Sajjad Zarifzadeh, Philippe Liu, and Reza Shokri.
\newblock Low-cost high-power membership inference by boosting relativity.
\newblock \emph{arXiv preprint arXiv:2312.03262}, 2023.

\bibitem[Zhang et~al.(2022)Zhang, Roller, Goyal, Artetxe, Chen, Chen, Dewan, Diab, Li, Lin, Mihaylov, Ott, Shleifer, Shuster, Simig, Koura, Sridhar, Wang, and Zettlemoyer]{zhang2022opt}
Susan Zhang, Stephen Roller, Naman Goyal, Mikel Artetxe, Moya Chen, Shuohui Chen, Christopher Dewan, Mona Diab, Xian Li, Xi~Victoria Lin, Todor Mihaylov, Myle Ott, Sam Shleifer, Kurt Shuster, Daniel Simig, Punit~Singh Koura, Anjali Sridhar, Tianlu Wang, and Luke Zettlemoyer.
\newblock Opt: Open pre-trained transformer language models, 2022.

\end{thebibliography}
\bibliographystyle{colm2024_conference}

\newpage
\appendix

\section{Implementation details} \label{sec:appendix-implementation-details}
\subsection{\mimir}\label{subsec:appendix-mimir}

We release our codebase as a Python package (available at \repourl), complete with documentation and tests, for implementing and evaluating membership inference attacks for language models.
\ifisarxiv
    The data used in our experiments is also available via HuggingFace (\url{https://huggingface.co/datasets/iamgroot42/mimir}).
\fi
Some features of the package include:
\begin{itemize}
    \item A base attack class that provides bare-bones code and helper functions that can be easily used in implementations of both existing and new attacks.
    \item Data processing utilities to filter and cache data for membership evaluation, from both provided and available sources.
    \item Support for a vast array of models that can be used as target or reference models.
\end{itemize}
The entire codebase works with modular configuration files, allowing multiple experiments to be run simultaneously with no edits to the code itself.
We used Python 3.9.7 for our experiments, with PyTorch 2.0.1. Our experiments were executed on a mix of machines, with GPUs ranging from RTX6k to A100.

\subsection{Additional Target Model Details}\label{subsec:appendix-target-model-details}
\shortsection{\pythiadedup} Both the \pythia\ and \pythiadedup\ model suites provide intermediate checkpoints for each model. For experiments targeting the \pythiadedup\ model, as the \pythiadedup\ model is trained for greater than 1 epoch, we select the checkpoint that most closely matches the one epoch mark over the deduplicated Pile. We decide this is checkpoint 'step99000'. For experiments targeting the non-deduped \pythia\ models, we use the final checkpoint, which sees just under one ($\approx 0.9$) epoch of the original Pile.

\shortsection{\silo} The models from the \silo\ suite~\citep{silo} consist of 1.3B-parameter transformer LMs based on the OpenLM implementation of the
LLaMA architecture~\citep{touvron2023llama1}. These are trained for multiple epochs on the Open License Corpus, which consists of permissively-licensed text data classified as either \textbf{public domain} (PD) texts, \textbf{permissively licensed software} (SW), or under an \textbf{attribution license} (BY). We target the \textsc{SILO-PDSW} model (alongside its intermediate checkpoints) trained on only texts classified as PD or SW for domains contributing less than 5\% of the data upsampled by a factor of 3x (which includes HackerNews and DM Mathematics).

\shortsection{\datablations} The \datablations\ suite~\citep{muennighoff2023scaling} is a large collection of models trained to study scaling laws in data-constrained regimes. They vary in the extent of data
repetition and compute budget, ranging up to 900 billion training tokens and 9
billion parameters. For the epoch experiment, we choose the 2.8B-parameter subset of models, with each seeing a total of 55B tokens from the C4 dataset across their training runs. These models vary in the number of epochs their training subset is seen, ranging from one to 14 epochs. They also offer a model trained for 44 epochs, which we decided to leave out of evaluation. 

\shortsection{\neo} is a collection of 125M-, 1.3B-, and 2.7B-parameter models of similar architecture to the \textsc{GPT-3} model family. These models are trained on the Pile for about 300B tokens, similar to the \pythia\ suite. This model suite is a precursor to the \textsc{GPT-NeoX}~\citep{gpt-neox-library} model architecture, which \pythiadedup\ and \pythia\ are built on. Noticeable differences include the tokenizer used per model suite, with the \textsc{GPT-NeoX} allocating additional tokens to whitespace characters, as well as intended training settings, with \neo\ geared towards TPU training and \textsc{GPT-NeoX} GPU training.

\shortsection{\olmo} The \olmo\ model suite~\citep{Groeneveld2023OLMo} is a suite of open language models trained on the \dolma\citep{dolma} dataset. \olmo\ models currently available include 1B- and 7B-parameter variants trained on 3T and 2.5T tokens, respectively. While our preliminary results just target the final checkpoints, the \olmo\ suite is similar to the \pythia\ suite in that intermediate checkpoints and exact training order are fully open.

\subsection{Benchmark Details}\label{subsec:appendix-benchmark-details}
We sample 1,000 members and non-members from each target domain from the Pile train and test sets, respectively. We do the same for the aggregate Pile experiment, except we sample 10,000 members and non-members each from the complete Pile train and test sets. We sample documents greater than 100 words and truncate them up to 200 words from the beginning to create our benchmark examples. Previous work~\citep{shi2023detecting} observes that sample length correlates with performance, so we bound the sample length to reduce its impact while picking a reasonable threshold so that our samples are likely to contain ample signal. While further increasing the length of samples could yield greater MIA performance, such an experiment is orthogonal to the ones we conduct; inherent differences in LLM training and MI evaluation would still impact evaluation on longer texts.

We follow the same pipeline when generating the benchmark for targeting the \datablations\ models, picking members and non-members from the C4 train and validation sets, respectively.

For our additional decontamination on Pile benchmarks, we follow~\cite{bff2023}, which uses a bloom filter to check for \ngram\ inclusion. We keep the default filtering settings of $n=13$ and a threshold of $\leq80$\% overlap. Further details about setting up the bloom filter can be found in Appendix~\ref{subsec:appendix-ngram-overlap-implementation}

\subsubsection{Training Data Size Benchmarks}\label{subsubsec:appendix-tds-benchmark-details}

For each model, we pick checkpoints every 5000 steps ending at step 95000, with each step corresponding to 1024 samples of length 2048 tokens. We also include checkpoints at step 1000 and step 99000, the closest checkpoint to the step where one full epoch of the deduplicated Pile was seen. For each checkpoint, we use the same non-member set for evaluation consisting of 1000 samples sampled from the entire Pile test set. We then construct a member set for each checkpoint\footnote{Ideally, the member set should be fixed, which could be done by performing multiple training runs and injecting the fixed member set at various steps. However, this is computationally expensive. Furthermore, because of how the data is shuffled, we'd expect the difficulty of the member set to be reasonably consistent across our samples}: for the checkpoint at step $n$, we sample 1000 random samples from documents seen within the range step \{$n - 100$, $n$\}. EleutherAI provides random seeding for deterministic training data order across the \pythiadedup\ training runs, which we use to determine the seen document order. This allows us to determine which documents to sample from for a given step range. For both members and non-members, we sample with the same criterion as the general experiments above.

\subsubsection{Temporal Benchmarks}\label{subsubsec:appendix-temporal-benchmark-details}

For the temporal Wikipedia benchmark non-members, we collect samples from the RealTimeData "\textbf{wikitext\_latest}" dataset~\citep{li2023avoiding}. This yielded Wikipedia articles created between the week of August 12, 2023 till the week of January 8, 2024~\footnote{Note that while the articles are created in the recent time frame, the contents of the Wikipedia page aren't necessarily about recent topics, people, or events}. We then follow Pile processing steps by simply appending the article titles to the front of each respective article with a ``\textbackslash n\textbackslash n''. Members are sampled from the Wikipedia subdomain of the Pile training set.  Members and non-members are then sampled with the same criterion as in the general experiments.

For the temporal ArXiv benchmarks, the member set for each benchmark is fixed and sampled from the ArXiv subdomain of the Pile training set, which consists of papers posted prior to July 2020~\citep{gao2020pile}. For non-members, we use the ArXiv API again following \citet{li2023avoiding} to collect ArXiv preprints from specific months: August 2020, January 2021, June 2021, January 2022, June 2022, January 2023, and June 2023~\footnote{This slightly differs from the Pile ArXiv data collection, which uses the ArXiv bulk access through S3. However, we believe both ArXiv bulk access and API should yield the preprints in the same manner regardless.}. We then apply the same processing steps used in the Pile~\citep{gao2020pile}. This mainly involves converting the latex sources for a given preprint into a single Markdown file, and then filtering out documents such as those with conversion errors. For each month range, we sample non-members from processed files in the given date range. By sampling non-members from successively later time
ranges after the Pile ArXiv cutoff date, we also seek to explore how greater temporal shift impacts MIA performance. We again sample both members and non-members with the same criterion as in the general experiments.

\shortsection{Difficulties in reproducing and analyzing existing works} While we follow a similar method of non-member candidate selection for our temporal experiments as prior works such as \cite{shi2023detecting} and \cite{meeus2023did}, we were unable to reproduce their settings for analysis for two main reasons: 1) their exact non-member candidates and/or pipelines to reproduce the non-member candidate selection are unavailable or 2) target models used such as \textsc{LLaMa}~\citep{touvron2023llama1} do not release their training data to conduct \ngram\ or other distributional analyses.

\begin{figure}[!t]
\begin{center}
\includegraphics[width=.7\columnwidth]{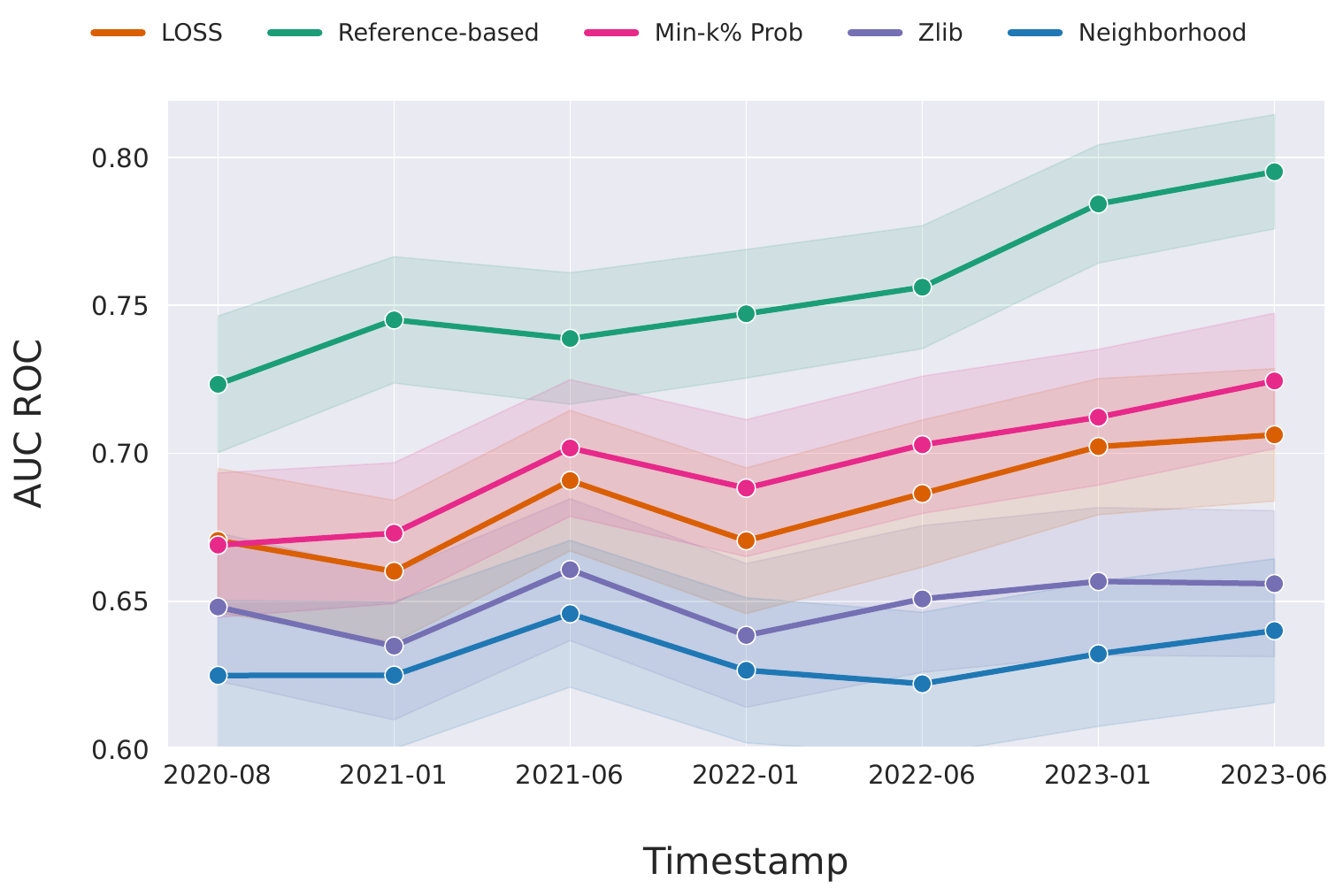}
\vspace{-.75em}
\caption{MIA performance across benchmarks where non-member data is selected from ArXiv preprints created during increasingly later months past the target model's knowledge cutoff. Timestamps are formatted as year-month. The target model is \pythiadedup-12B. In general, \textbf{MIA performance increases as the temporal shift of non-members increases}}
\vspace{-1em}
\label{fig:temporal-arxiv}
\end{center}
\end{figure}

\begin{table}[!h]
\begin{center} \scriptsize
\setlength{\tabcolsep}{0.7pt}
\begin{tabularx}{\textwidth}{l *{12}{>{\centering\arraybackslash}X}@{}}
    \toprule
    \multirow{2}{*}{Thresholding Benchmark} & \multicolumn{4}{c}{1\%} & \multicolumn{4}{c}{5\%}& \multicolumn{4}{c}{10\%}\\
    \cmidrule(lr){2-5}  \cmidrule(lr){6-9}  \cmidrule(lr){10-13} 
    & LOSS & Ref & \mink & zlib & LOSS & Ref & min-$k$ & zlib & LOSS & Ref & \mink & zlib
    \\
    \midrule
        2020-08 & 3.2 & 4.2 & 4.5 & 3.7 & 12.6 & 13.4 & 13.5 & 13.8 & 24.1 & 23.3 & 24.6 & 20.2 \\
        2021-01 & 3.7 & 3.9 & 3.5 & 3.5 & 11.4  & 15.8 & 13.5 & 10.4 & 21.7 & 27.0 & 24.6 & 17.5 \\
        2021-06 & 3.2 & 4.2 & 5.7 & 5.4 & 14.4 & 16.0 & 15.7 & 13.6& 25.5 & 25.5 & 29.5 & 23.0  \\
        2022-01 & 4.5 & 4.2 & 5.3 & 4.1 & 14.4 & 16.3 & 14.6 & 12.7 & 24.5 & 27.0 & 28.7 & 22.0 \\
        2022-06 & 2.8& 3.9 & 3.1 & 2.5 & 10.3 & 18.1 & 13.1& 10.7 & 23.4 & 27.8 & 25.4& 20.6 \\
        2023-01 & 2.9 & 8.5 & 3.5 & 3.1 & 11.9 & 23.5 & 13.5 & 10.9 & 25.0 & 36.1 & 26.3& 21.9 \\
        2023-06 &5.8  & 9.4 & 5.5 & 5.8 & 15.6 & 22.7 & 19.1 & 14.1 & 26.3 & 37.3 & 27.8 & 22.2 \\
    \midrule
        Temporal Wiki & 9.8 & 7.5 & 10.3 & 7.9 & 23.8 & 22.8 & 24.3 & 17.6 & 30.0 & 34.1 & 35.0 & 22.8 \\
    \bottomrule
\end{tabularx}
\vspace{-.5em}
\caption{FPR (\%) on non-members from the Pile (original; not temporally shifted) on various attacks when using a score threshold that achieves a 1, 5, or 10\% FPR on the temporally-shifted ArXiv (for varying levels of temporal shift) and Wikipedia benchmarks. The target model is \pythiadedup-12B. \textbf{FPRs on the original non-members are much higher then the thresholded FPR on the temporally shifted benchmarks}, indicating that such thresholds may be moreso classifying temporal shift rather than member and non-members.}\label{tab:temporal_crossdomain}
\vspace{-.5em}
\end{center}
\end{table}

\subsection{Attack Details} \label{subsec:appendix-attack-details}

MIAs consider a target model $\mathcal{M}$, which outputs a probability distribution of the next token given a prefix, denoted as $P(x_t|x_1...x_{t-1}; \mathcal{M})$.
Their goal is to model $f(\mathbf{x}; \mathcal{M})$, which outputs a score for target sample $\mathbf{x} = x_1...x_n$ with $n$ tokens. This score is then thresholded to determine the target sample's membership in the training data of $\mathcal{M}$.

\bfsection{LOSS}~\citep{yeom2018privacy,carlini2019secret} considers the model's computed loss over the target sample:
$$
f(\mathbf{x}; \mathcal{M}) = \mathcal{L}(\mathbf{x};\mathcal{M}).
$$

\bfsection{Reference-based}~\citep{sablayrolles2019white,watson2022on} attacks assume access to a reference model $\mathcal{M}_\text{ref}$, another LM trained on a disjoint set of training data drawn from a similar distribution. In practice, an assumption of disjoint training data is impractical. Empirically, using an LM that is different from $\mathcal{M}$ has been a reasonable choice and was used in prior work~\citep{kandpal2022deduplicating,watson2022on}.
The attack considers the membership score of the target sample by $\mathcal{M}$ relative to the membership by $\mathcal{M}_\text{ref}$ to calibrate the target model's score given a difficulty estimate through the reference model's score, with goals to improve precision and reduce the false negative rate. For our experiments, we use LOSS as the uncalibrated membership score such that, for the reference-based attacks,
$$
f(\mathbf{x}; \mathcal{M}) = \mathcal{L}(\mathbf{x}; \mathcal{M})  - \mathcal{L}(\mathbf{x}; \mathcal{M}_\text{ref}).
$$

This method exactly follows the method from \cite{watson2022on} and is also largely similar to the offline Likelihood Ratio attack (LiRA; \cite{carlini2022membership}), although LiRA uses many reference models (often trained shadow models).

\bfsection{Zlib Entropy} \citep{carlini2021extracting} functions similarly to reference-based MIA, using the zlib compression size of a sample $\mathbf{x}$ as a local difficulty threshold per sample: $$
f(\mathbf{x}; \mathcal{M}) = \frac{\mathcal{L}(\mathbf{x}; \mathcal{M})}{\text{zlib}(\mathbf{x})},
$$
where zlib($\mathbf{x}$) is the length in bytes of the zlib compressed sample.

\bfsection{Neighborhood Attack}~\citep{mattern2023membership} assumes access to a masking model, and operates by generating ``neighbor'' texts $\tilde{\mathbf{x}}$ to a given text sequence $\mathbf{x}$ by using the masking model to replace a percentage of randomly selected token spans while still maximizing the neighbor's likelihood. If the sample's loss is considerably lower than the neighbor's losses, the difference is attributed to the target model overfitting the sample, and the sample is considered a training member. Formally, we have 
$$
f(\mathbf{x}; \mathcal{M}) = \mathcal{L}(\mathbf{x}; \mathcal{M}) - \frac{1}{n}\sum_{i=1}^n \mathcal{L}(\mathbf{\tilde{x}}_i; \mathcal{M}).
$$
We use BERT~\citep{devlin2019bert} as our masking model of choice, with a masking percentage of 5\%. 

\bfsection{Min-$k$\% Prob} \citep{shi2023detecting} is based on the intuition that non-member
examples tend to have more tokens assigned lower likelihoods than member examples do. Given sample $\mathbf{x} = x_1,...,x_n$ and hyperparameter $k$, let min-$k(\mathbf{x})$ be the set formed by the $k$\% of tokens in $\mathbf{x}$ with minimum likelihood. We then have
$$
f(\mathbf{x}; \mathcal{M}) = \frac{1}{\lvert \text{min-$k(\mathbf{x})$}\rvert} \sum_{x_i \in \text{min-$k(\mathbf{x})$}}-\log(p(x_i \mid x_1,...,x_{i-1})).
$$
We experiment with multiple different $k \in \{10,20,30,40,50\}$ as suggested in \citet{shi2023detecting}, but settle on $k = 20$ for our experiments.

We compute the performance of each attack based on 1,000 bootstrap samples of the benchmark and report the average AUC ROC and TPR@low\%FPR over the bootstraps. 

\shortsection{MIAs involving meta-classifiers} While many recent works study MIAs that perform membership classification through meta-classifiers trained on features extracted from a ground-truth subset of member/non-member data, we primarily focus on blackbox attacks as access to such a subset can be difficult to guarantee in practice, especially as the inclusion of samples in pre-training corpora becomes more ambiguous as these corpora continue to expand.

\subsection{Reference Model Choices}\label{subsec:appendix-ref-model}
We choose a diverse set of reference models to experiment with. For the aggregate method over all reference models, we take the average of the scores per reference model for a target sample\footnote{Note that the scores over different reference models may not be directly comparable due to the reference models having different tokenizers. This may contribute to the poor performance of this naive ensembling method.}. We report results for our complete ablation on reference model choice in \Cref{tab:appendix-ref_model_choice}.

\textsc{GPT-2}~\citep{radford2019language} is suite of pre-trained transformer trained on a large dataset of around 40GB of web text, likely overlapping with the Pile. We use the \textsc{GPT-2}-small variant with 124M parameters.

\textsc{DistilGPT2}~\citep{sanh2019distilbert} is a smaller 82M-parameter model trained with the supervision of \textsc{GPT-2}-small using knowledge distillation.

\textsc{OPT}~\citep{zhang2022opt} is a suite of open-sourced pre-trained transformers that are trained on a curated pre-training corpus including several datasets from the Pile, such as Wikipedia, DM Mathematics, and HackerNews. We use the 1.3B-parameter variant.

As mentioned in \Cref{subsec:appendix-target-model-details}, \textsc{GPT-NEO}~\citep{gpt-neo} is another suite of pre-trained transformers designed using EleutherAI's replication of the \textsc{GPT-3} architecture. These models are trained on the full Pile for a similar amount of tokens as \pythia\ ($\sim300$B), though the data seen may not necessarily be in the same order as the \pythia\ models. We use the 1.3B-parameter variant.

\textsc{SILO-PDSWBY}~\citep{silo} is a 1.4B-parameter transformer pre-trained on all types of permissively licensed data in the Open License Corpus. The training data consists of certain Pile domains such as HackerNews and DM Mathematics.

\textsc{LLAMA}~\citep{touvron2023llama1} is a collection of large, open-sourced pre-trained LMs ranging in size from 7B to 65B parameters. The pre-training corpus is on the scale of trillions of tokens, much larger than the Pile, and likely has significant overlap with the Pile. We use the 7B-parameter variant.

\textsc{StableLM-Alpha-v2}~\citep{StableLMAlphaV2Models} is a set of open-source pre-trained LMs also trained on a large pre-training corpus with trillions of tokens. Training is conducted in two stages, with the first stage seeing 1 trillion tokens of a mixture of data from sources such as RedPajama~\citep{together2023redpajama} and the Pile, with an emphasis on refined web text. The second stage is trained on 100 billion tokens with a higher context length, increasingly sampling naturally long texts and adding the StarCoder~\citep{li2023starcoder} dataset. We use the 3B-parameter variant.

We also experiment with the non-deduped \pythiadedup-1.4B model as a reference model to see how using a smaller version of the target model (same architecture and training data order) impacts reference-based attack performance~\citep{carlini2021extracting}. 
\begin{table*}[t]
\begin{center} \scriptsize
\setlength{\tabcolsep}{0.7pt}
\begin{tabularx}{\textwidth}{l *{16}{>{\centering\arraybackslash}X}@{}}
    \toprule
    \multirow{2}{*}{}  & \multicolumn{8}{c}{Wikipedia} & \multicolumn{8}{c}{Pile CC} \\
    \cmidrule(lr){2-9}  \cmidrule(lr){10-17}
    \textbf{\# Params} & \tiny{\textsc{Gpt2}} & \tiny{\textsc{Distil}} & \tiny{\textsc{Opt}} & \tiny{\textsc{Neo}} & \tiny{\textsc{Silo}} & \tiny{\textsc{Llama}} & \tiny{\textsc{Stable}} & \tiny{\textsc{Pythia}} & \tiny{\textsc{Gpt2}} & \tiny{\textsc{Distil}} & \tiny{\textsc{Opt}} & \tiny{\textsc{Neo}} & \tiny{\textsc{Silo}} & \tiny{\textsc{Llama}} & \tiny{\textsc{Stable}} & \tiny{\textsc{Pythia}}
    \\
    \midrule
        160M & .498& .502& .494& .490& .492& .511& \textbf{.515}& .480     & \textbf{.520}& .504& .488& .473& .504& .487 & .497& .480\\
        1.4B & .503& .505& .507& .500& .502& .521& \textbf{.544}& .476     & .523& .507& .513& .500& .516& .504& \textbf{.525}&.496 \\
        2.8B & .511 & .510& .519& .532& .531& .539& \textbf{.565}& .526     & .526& .509 & .521 & .499 & .520 & .510& \textbf{.537}& .504\\
        6.9B & .510& .507& .517& .518& .516 & .536& \textbf{.571}& .501     & .538& .520& .542& .525& .531& .530 & \textbf{.564}& .540\\
        12B & .514 & .510& .522& .528& .529 & .546& \textbf{.579}& .517     & .548 & .525 & .555& .538& .541&  .545& \textbf{.582}& .555\\
        \vspace{-.6em} \\
    \toprule
    \multirow{2}{*}{}  & \multicolumn{8}{c}{PubMed Central} & \multicolumn{8}{c}{ArXiv} \\
    \cmidrule(lr){2-9}  \cmidrule(lr){10-17}
    \textbf{\# Params} & \tiny{\textsc{Gpt2}} & \tiny{\textsc{Distil}} & \tiny{\textsc{Opt}} & \tiny{\textsc{Neo}} & \tiny{\textsc{Silo}} & \tiny{\textsc{Llama}} & \tiny{\textsc{Stable}} & \tiny{\textsc{Pythia}} & \tiny{\textsc{Gpt2}} & \tiny{\textsc{Distil}} & \tiny{\textsc{Opt}} & \tiny{\textsc{Neo}} & \tiny{\textsc{Silo}} & \tiny{\textsc{Llama}} & \tiny{\textsc{Stable}} & \tiny{\textsc{Pythia}}
    \\
    \midrule
        160M & .495& .491& .515& .511& .513& .515 & \textbf{.516}& .497     & \textbf{.523}& .518& .516& .480& .496& .492 & .486&.472 \\
        1.4B & .493 & .491& .514& .517& .514& .515& \textbf{.530}& .503     & \textbf{.529}& .524& .523& .501& .512& .506 & .510& .484 \\
        2.8B & .494 & .492 & .513& .518& .515& .518& \textbf{.536}& .500     & \textbf{.534}& .528& .528& .524& .522& .516& .531& .528\\
        6.9B & .499& .496& .519& .527& .520 & .530& \textbf{.552}& .526     & .540& .532& .534& .539& .531& .528 & .538& \textbf{.554}\\
        12B & .504 & .498& .523& .531& .524& .538& \textbf{.559}&.533     & .546& .538 & .541& .555& .540& .538& .555& \textbf{.581}\\
        \vspace{-.6em} \\
    \toprule
    \multirow{2}{*}{}  & \multicolumn{8}{c}{DM Math} & \multicolumn{8}{c}{HackerNews} \\
    \cmidrule(lr){2-9}  \cmidrule(lr){10-17}
    \textbf{\# Params} & \tiny{\textsc{Gpt2}} & \tiny{\textsc{Distil}} & \tiny{\textsc{Opt}} & \tiny{\textsc{Neo}} & \tiny{\textsc{Silo}} & \tiny{\textsc{Llama}} & \tiny{\textsc{Stable}} & \tiny{\textsc{Pythia}} & \tiny{\textsc{Gpt2}} & \tiny{\textsc{Distil}} & \tiny{\textsc{Opt}} & \tiny{\textsc{Neo}} & \tiny{\textsc{Silo}} & \tiny{\textsc{Llama}} & \tiny{\textsc{Stable}} & \tiny{\textsc{Pythia}}
    \\
    \midrule
        160M & .489& .488& .520& .509& .487& .502& \textbf{.523}&.514     & \textbf{.496}& \textbf{.496}& \textbf{.496}& .480& .398& .486 & .490& .466\\
        1.4B & .487& .485& .509& .496& .485& .503 & \textbf{.512}& .496     & .508& .509& .511& .496& .401& .504 & \textbf{.514}& .483\\
        2.8B & .485& .486& \textbf{.511}& .503& .483& .500& .504& .509     & .521& .522 & .529& .534& .421& .521& \textbf{.549}& .527 \\
        6.9B & .485& .485& \textbf{.510}& .499& .484 & .502& .508&.497     & .525& .526& .534& .536& .436& .531 & \textbf{.546}& .542\\
        12B & .487& .486& \textbf{.514}& .504& .485& .502& .512& .503    & .534& .533 & .545& .559& .453& .545& \textbf{.565}& .561\\
        \vspace{-.6em} \\
    \bottomrule
\end{tabularx}
\end{center}
\vspace{-.5em}
\caption{The effect of the choice of a reference model to \pythiadedup\ models across various domains. The reference model yielding the highest performance, per target domain and target model, is bolded. ROC-AUC values are reported.}
\label{tab:appendix-ref_model_choice}
\vspace{-.5em}
\end{table*}

\subsubsection{Stablelm-Base-Alpha-3B-v2 Performance}
\label{app:stablelm_perf}

We speculate that the slightly higher performance with \textsc{Stablelm-Base-Alpha-3B-v2} as the reference model, even though its pre-training corpus has high overlap with the Pile, is because 1) larger target models\footnote{Also target models that are domain specific like \datablations\ or are trained on a less diverse corpus like \silo} such as the \pythiadedup-12B model may considerably overfit certain member samples and 2) the \textsc{Stablelm-Base-Alpha-3B-v2} is trained on a much larger corpus, which helps it generalize well and achieve similar losses as the target model on the non-member data. As a result, member samples are more likely to have a greater magnitude of difference between the target and reference model losses compared to the difference between losses on non-members.

\subsection{Results with GPT-Neo models} \label{app:neo_results}

We repeat our experiments with the \neo\ family of models. \Cref{tab:appendix-gen_exp_results_gptneo} demonstrates similar trends targeting the \neo\ models as seen when targeting the \pythia\ model family (\Cref{tab:gen_exp_results}, \Cref{tab:appendix-nondedupe_pythia_results}), such as MIA performance generally increasing as the target model size increases. In general, performance against the \neo\ models is similar, if not lower than, performance against the \pythiadedup\ and \pythia\ models when comparing similarly sized variants. In some domains such as HackerNews, the best performing MIA differs between target models (Min-$k$\% Prob for \neo, reference-based for \pythiadedup), though marginally.

\begin{table*}[h]
\begin{center} \scriptsize
\setlength{\tabcolsep}{0.7pt}
\begin{tabularx}{\textwidth}{l *{16}{>{\centering\arraybackslash}X}@{}}
    \toprule
    \multirow{2}{*}{}  & \multicolumn{4}{c}{Wikipedia} & \multicolumn{4}{c}{Github} & \multicolumn{4}{c}{Pile CC} & \multicolumn{4}{c}{Pubmed Central} \\
    \cmidrule(lr){2-5}  \cmidrule(lr){6-9} \cmidrule(lr){10-13} \cmidrule(lr){14-17}
    \textbf{\# Params} & LOSS & Ref & min-$k$ & zlib
    & LOSS & Ref & min-$k$ & zlib
    & LOSS & Ref & min-$k$ & zlib
    & LOSS & Ref & min-$k$ & zlib
    \\
    \midrule
        125M & .504 & \textbf{.511} & .492 & \textbf{.511}     &.641&.582&.642&\textbf{.660}   &.495&.492&\textbf{.500}&.497    &.499&\textbf{.506}&.502&.499\\
        1.3B & .510 & \textbf{.531} & .506 & .517     &.681&.570&.681&\textbf{.696}     &.500&\textbf{.517}&.503&.501    &.496&\textbf{.499}&\textbf{.499}&.497\\
        2.7B & .513 & \textbf{.545} & .513 & .519    &.699&.570&.700&\textbf{.712}    &.504&\textbf{.531}&.507&.506    &.498&\textbf{.507}&.501&.499\\
        \vspace{-.6em} \\
    \toprule
    \multirow{2}{*}{}  & \multicolumn{4}{c}{ArXiv} & \multicolumn{4}{c}{DM Math} & \multicolumn{4}{c}{HackerNews} & \multicolumn{4}{c}{The Pile}\\
    \cmidrule(lr){2-5}  \cmidrule(lr){6-9} \cmidrule(lr){10-13} \cmidrule(lr){14-17}
    \textbf{\# Params} & LOSS & Ref & min-$k$ & zlib
    & LOSS & Ref & min-$k$ & zlib
    & LOSS & Ref & min-$k$ & zlib
    & LOSS & Ref & min-$k$ & zlib
    \\
    \midrule
        125M &\textbf{.507}&.494&.503&.501    &.492&\textbf{.522}&.493&.484    &.489&.480&\textbf{.505}&.496    &.502&\textbf{.507}&.505&.505\\
        1.3B &.511&.506&\textbf{.512}&.507    &.486&\textbf{.511}&.491&.481    &.499&.500&\textbf{.514}&.501    &.505&\textbf{.514}&.509&.507\\
        2.7B &.515&\textbf{.520}&.517&.510    &.486&\textbf{.509}&.492&.481   &.502&.512&\textbf{.516}&.503    &.507&\textbf{.519}&.511&.509\\
    \bottomrule
\end{tabularx}
\end{center}
\vspace{-.5em}
\caption{AUC ROC of MIAs against \neo\ across different datasets from the Pile.
The highest performance across the different MIAs is bolded per domain. Similar to \pythiadedup, \textbf{MIA methods perform near random ($<.55$) in most domains}.%
}\label{tab:appendix-gen_exp_results_gptneo}
\vspace{-.5em}
\end{table*}

\subsection{Results with OLMo models} \label{app:olmo_results}

We perform preliminary experiments targeting the \olmo\ family of models. Benchmark construction is similar to what is detailed in \Cref{subsec:appendix-benchmark-details}. However, we sample from domains that make up \dolma~\citep{dolma}, namely Wikipedia, C4, Reddit, Common Crawl, and peS2o~\citep{peS2o}. These are similar to domains used in Pile. We note that peS2o consists of both abstracts (s2ag) and full papers (s2orc), and evaluate them as separate domains. To get non-member, we use held-out \dolma\ data from \textsc{Paloma}~\citep{paloma}. 

\Cref{tab:appendix-gen_exp_results_olmo} also demonstrates generally near-random performance trends, with both the 1B- and 7B-parameter model variants exhibiting similar performances. We speculate that, due to the incredibly large amounts of training data (3T tokens for 1B-parameter model, 2.5T tokens for the 7B-parameter model), performance across different model sizes begins to converge to near-random performance even with such distinct model sizes. Interestingly, the Reference-based attack using \textsc{Stablelm-Base-Alpha-3B-v2} performs much worse than when used to calibrate the \pythia\ models, reinforcing the difficulty in finding suitable reference models for different LLMs. We also observe many settings where MIA performance is considerably less than .5, suggesting that the MIAs are more likely to predict members as non-members, and vice versa. More investigation is needed to understand such behaviors on specific domains such as s2ag from peS2o in the \dolma\ data.

\begin{table*}[h]
\begin{center} \scriptsize
\setlength{\tabcolsep}{0.7pt}
\begin{tabularx}{\textwidth}{l *{12}{>{\centering\arraybackslash}X}@{}}
    \toprule
    \multirow{2}{*}{}  & \multicolumn{4}{c}{Wikipedia} & \multicolumn{4}{c}{C4} & \multicolumn{4}{c}{Reddit} \\
    \cmidrule(lr){2-5}  \cmidrule(lr){6-9} \cmidrule(lr){10-13}
    \textbf{\# Params} & LOSS & Ref & min-$k$ & zlib
    & LOSS & Ref & min-$k$ & zlib
    & LOSS & Ref & min-$k$ & zlib
    \\
    \midrule
        1B & .484 & \textbf{.510} & .495 & \textbf{.510}    & .515 & .479 & \textbf{.520} & .513    & .464 & \textbf{.495} & .478 & .470\\
        7B & .481 & .488 & .493 & \textbf{.500}      & .516 & .499 & \textbf{.520} & .514   & .463 & \textbf{.501} & .480 & .469\\
        \vspace{-.6em} \\
    \toprule
    \multirow{2}{*}{}  & \multicolumn{4}{c}{Common Crawl} & \multicolumn{4}{c}{s2ag} & \multicolumn{4}{c}{s2orc} \\
    \cmidrule(lr){2-5}  \cmidrule(lr){6-9} \cmidrule(lr){10-13}
    \textbf{\# Params} & LOSS & Ref & min-$k$ & zlib
    & LOSS & Ref & min-$k$ & zlib
    & LOSS & Ref & min-$k$ & zlib
    \\
    \midrule
        1B & .509 & .412 & \textbf{.517} & .511     & .449 & .376 & \textbf{.461} & .392    & .484 & .480 & \textbf{.500} & .463 \\
        7B & .498 & .410 & \textbf{.505} & .500     & .465 & \textbf{.483} & .475 & .406   & .491 & \textbf{.507} & .503 & .470\\
    \bottomrule
\end{tabularx}
\end{center}
\vspace{-.5em}
\caption{AUC ROC of MIAs against \olmo\ across different datasets from the Dolma dataset.
The highest performance across the different MIAs is bolded per domain.%
}\label{tab:appendix-gen_exp_results_olmo}
\vspace{-.5em}
\end{table*}

\section{\texorpdfstring{\ngram}{n-gram}\ Overlap Details and Takeaways}\label{sec:appendix-other-ngram-takeaways}

\subsection{Measuring \texorpdfstring{\ngram}{n-gram}\ Overlap}\label{subsec:appendix-ngram-overlap-implementation}
We create a bloom filter following \citet{bff2023}. Due to the scale of the Pile training data and limited memory, we shard the bloom filter. In our construction, we split the training data in half, resulting in two bloom filter shards. Since each shard only sees half of the training data, to check for \ngram\ inclusion across the entire Pile, we check for containment in both of the sharded bloom filters, counting an \ngram\ included only if it is included in at least one of the bloom filters.

For each shard, we configure the bloom filter according to the data size such that the false positive rate of the bloom filter is less than 1\% (0.6\%). Then, for each document, we tokenize at the word level. We then add \ngram\~s to the filter by using a striding window over $n$ words at a time with a stride of 1. We use the same method of gathering \ngram\~s when checking the non-members for \ngram\ overlap.

\subsection{Reference-based Attack Performance}\label{subsec:appendix-ref-vs-other-performance} \Cref{tab:ngram_threshold_results} shows that, interestingly, reference-based MIAs have a noticeably smaller increase in performance compared to non-referenced-based MIAs for domains such as GitHub or PubMed Central under \ngram\ overlap thresholding. We speculate that, since numerous low \ngram\ overlap non-members are outliers to the relevant domain, these non-members will also be outliers to the similar/overlapping data seen by the reference model. As a result, even though these non-members may yield higher losses from the target model, we see similar high losses for the reference model as well, which makes the difference between target and reference model loss for non-members and members relatively less distinguishable compared to signals from the other attacks.

At the same time, domains like Pile CC do not see this dampened performance, likely because the 20\% threshold in the case of Pile CC is not sufficient to select outliers, as samples from this domain have naturally low \ngram\ overlap. Another case where the reference-based attack seems to avoid this observation is in the temporally shifted non-member setting for both Wikipedia and ArXiv despite the temporally shifted non-members being more out-of-distribution relative to the Pile Wikipedia and ArXiv distributions, respectively. We speculate this is due to the reference model of choice, \textsc{Stablelm-Base-Alpha-3B-v2}, which has not only been trained on a corpus with high overlap with the Pile, but also trained on datasets that capture more recent data such as RedPajama-Data-1T~\citep{together2023redpajama} which contains Wikipedia and ArXiv samples from a much more recent cutoff date (\ie RedPajama uses the 2023-03-20 Wikipedia dump), allowing it to generalize better over the temporally shifted non-members and avoiding a shift towards higher losses that weaker or older reference models might experience.

Attacks like LOSS or \minkprob\ do not utilize any external signal or difficulty calibration, and thus rely exclusively on signals from the target model for member classification. Calibration-based methods like zlib and reference-based attacks, on the other hand, account for the inherent ``difficulty" of a seen sample. Thus, in situations where the non-member data is significantly out of domain, even for a reference model or calibration method, it is likely that the signals from the target model and difficulty calibration would cancel out, leading to a weakened MIA signal. On the other hand, difficulty calibration can further boost MIA signal in settings where the member data is inherently more likely to be memorized, such as in \Cref{subsubsec:ablation-llm-characteristics} where reference-based attacks yielded considerably higher MIA performance in low training data size and high effective epoch count settings, with performance being further amplified in the extremes of both settings. Thus, MIA baselines for new MIAs should include both kinds of methods: calibration-based and calibration-free. Having baseline coverage for both styles of MIA can help uncover inherent characteristics of the evaluation setting such as unintentional member/non-member distributional shift or overfitted target models that influence MIA performance and also paints a holistic picture with regards to what MIAs are most suitable for specific attack settings.

\subsection{GitHub as an Outlier}\label{subsec:appendix-github-outlier}
\begin{wrapfigure}{r}{0.5\textwidth}
\includegraphics[width=.5\columnwidth]{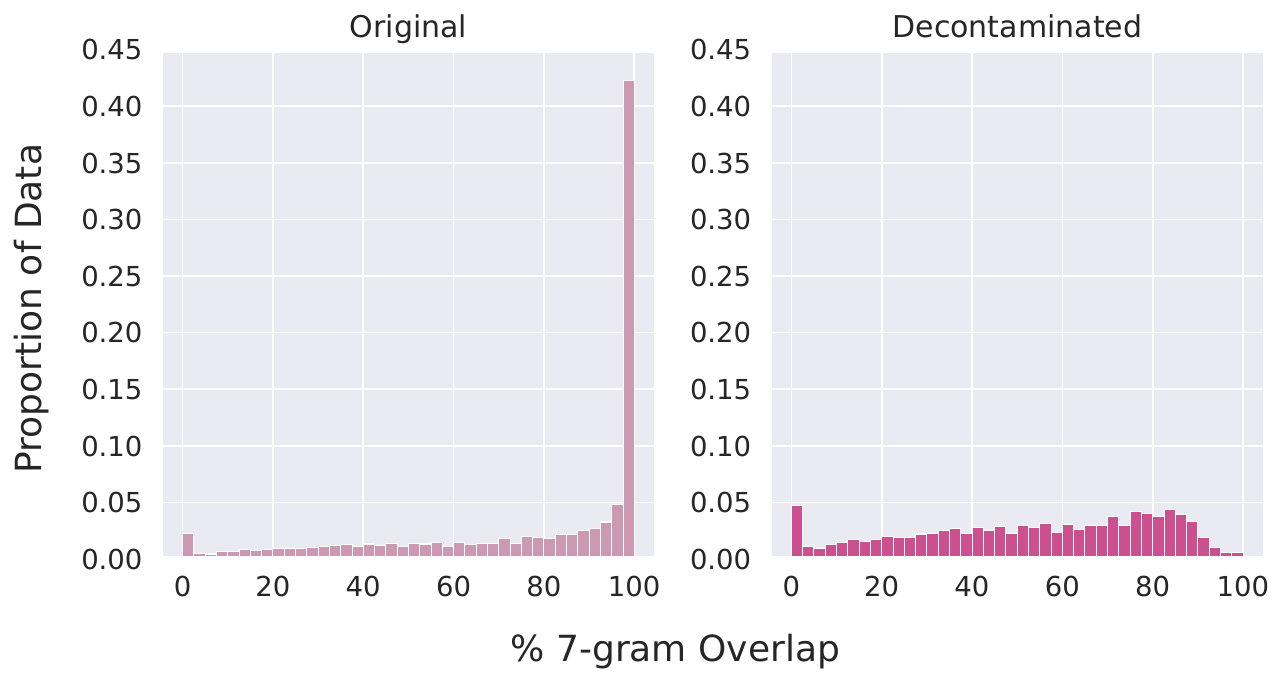}
\vspace{-2em}
\caption{7-gram overlap of GitHub non-member data before and after 13-gram decontamination at threshold $\leq80\%$.}
\label{fig:change-ngram-overlap-distribution-github}
\end{wrapfigure}

As seen in Table~\ref{tab:gen_exp_results}, MIA performance in the Github domain even without thresholding is notably higher than that in other domains, with the best method (zlib) achieving an AUC ROC of $\sim.70$. We speculate this is not because the GitHub domain, or code in general, is an easier domain to attack, but because the presumably reasonable decontamination threshold of $\leq$80\% 13-gram overlap threshold only captures a small percentile of non-members as GitHub is naturally very high overlap.

\begin{figure}[h]
\begin{center}
\includegraphics[width=1.05\textwidth]{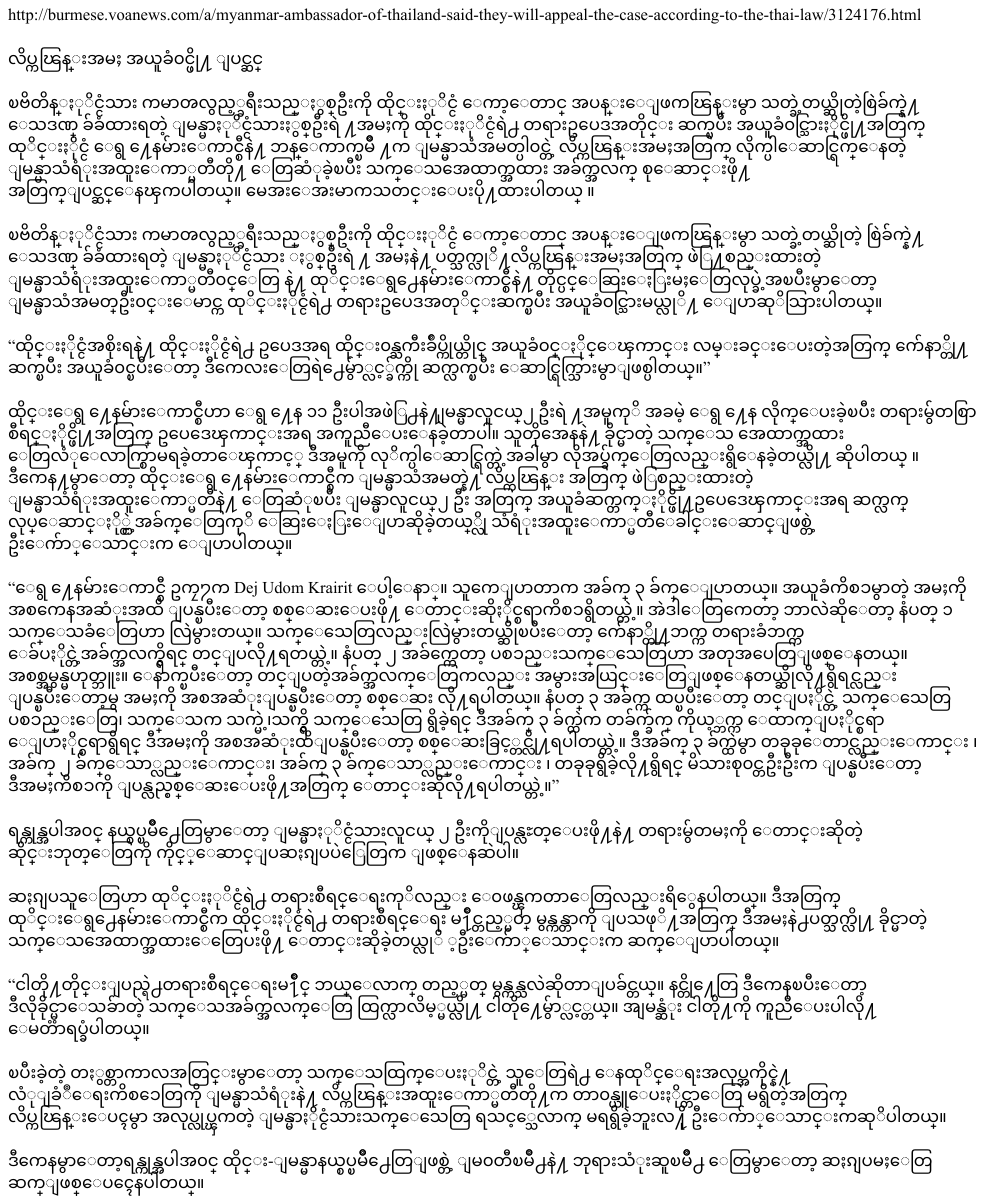}
\vspace{-11em}
\caption{A sample GitHub non-member outlier captured by the $\leq80\%$ 13-gram overlap threshold. This sample is from a language resource repository under Google, but is a clear outlier to the code-dominant GitHub domain}
\label{fig:appendix-github-outlier}
\end{center}
\end{figure}

We speculate a large factor contributing to the high overlap is the repetitive nature of code, such as copyright notices, function definitions, and syntax like HTML tags. Figure~\ref{fig:change-ngram-overlap-distribution-github} demonstrates how our decontamination threshold impacts the 7-gram distribution of non-members. Non-members under our decontamination threshold are more likely outliers to the GitHub domain (see Figure~\ref{fig:appendix-github-outlier} for an example of such an outlier). The additional \ngram\ overlap threshold experiments (\Cref{subsec:ablation-ambiguity}) only exacerbate the impact of thresholding, which leads to notably higher MIA performance.

Such observations indicate why lexical non-member boundaries may lead to ambiguous interpretations of MIA performance in high-overlap domains. Here, using semantic differences between samples to draw non-member boundaries may be key to better understanding membership leakage in such domains.

\subsection{Temporal Shift as Change in \texorpdfstring{\ngram}{n-gram}\ Overlap}\label{subsec:appendix-temporal-arxiv}

\begin{figure*}[h!]
\begin{center}
\includegraphics[width=.5\paperwidth]{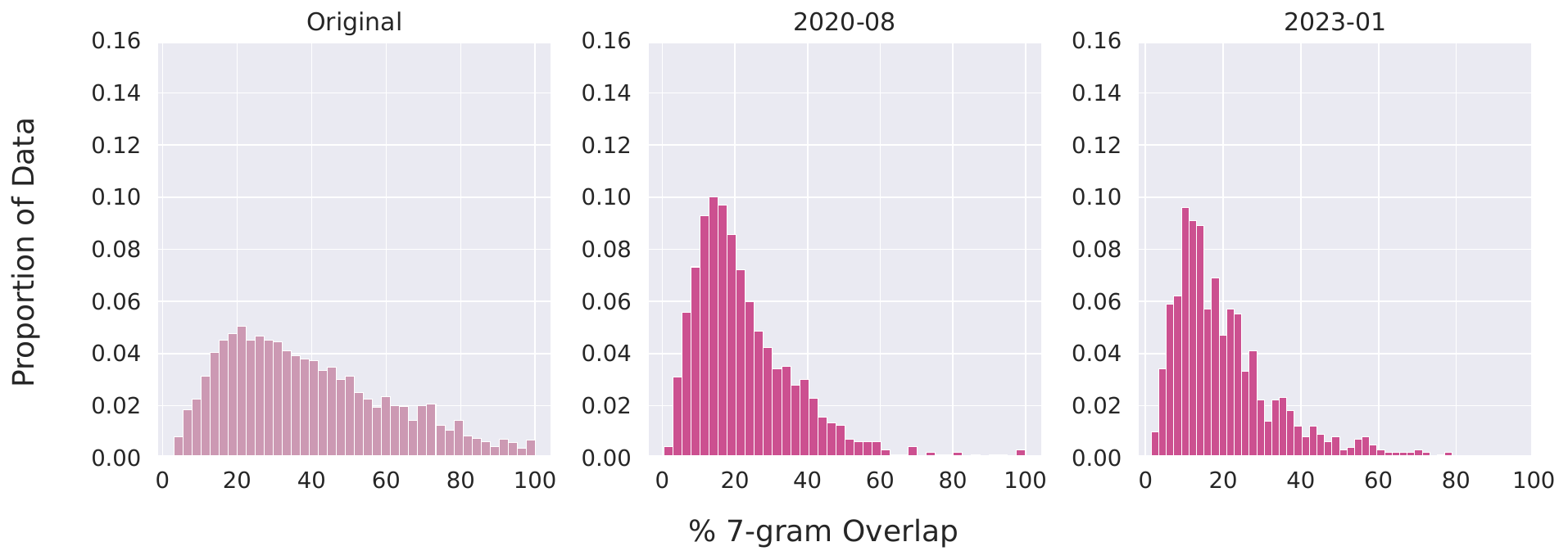}
\caption{Distribution of \ngram\ overlap for non-member ArXiv preprints sampled from the months 2020-08 and 2023-06, respectively. We also plot the \ngram\ overlap distribution of the original Pile ArXiv non-members. Between the original non-members and both temporally shifted non-member sets, \textbf{the temporally shifted non-member \ngram\ overlap distributions are considerably more concentrated at lower \% \ngram\ overlap}. The original non-members have an average 7-gram overlap of 39.3\%, while non-members from months 2020-08 and 2023-06 have 7-gram overlap of 22.7\% and 20.5\%, respectively.}
\label{fig:appendix-temporal-arxiv-ngram-distribution}
\end{center}
\end{figure*}

\Cref{fig:appendix-temporal-arxiv-ngram-distribution} reinforces our observations in \Cref{fig:temporal_wiki}, as similar to the temporal Wikipedia setting, temporally shifted non-members from after the target model's knowledge cutoff date are concentrated at considerably lower \% \ngram\ overlap than non-members from the natural ArXiv non-member set. This contributes to the greater MIA performance in general over the temporally shifted ArXiv benchmarks. However, we note that \ngram\ overlap distribution shift does not provide a strong interpretation for the increase in MIA performance as non-members are increasingly temporally shifted. For example, the average 7-gram overlap of non-members from the month 2020-08 is 22.7\% while the average for the month of 2023-06 is 20.5\%. While there is a small decrease in average 7-gram for later non-members, the change is quite small and doesn't clearly justify the considerable difference in MIA performance when evaluating on benchmarks using non-members from the different months (i.e., .723 $\rightarrow$ .795 AUC ROC from the 2020-08 benchmark to the 2023-06 benchmark). We speculate other factors that contribute to this increase include changes in the distribution of topics (i.e., increasing popularity of research into LLMs) 
and the presence of specific identifying tokens (i.e., dates, references, new terminology). We believe such factors only further reinforce the need to carefully analyze MIA benchmark construction when evaluating MIAs to understand what signals are truly being captured.

\section{Characteristics of LLM Training}\label{sex:appendix-llm-training}

\subsection{Recency of Member Samples}\label{sec:appendix-data-recency}

We explore how the recency of member samples seen in training impacts MIA performance. We follow the same setup as the training data experiment (\ref{subsubsec:appendix-tds-benchmark-details}), but instead of evaluating the checkpoint at step $n$ with the member data sampled from within steps \{$n -100, n$\} and the fixed non-member set, we fix the target model, only targeting the checkpoint at step 99000 for all the benchmarks. 
\begin{figure*}[h]
\begin{center}
\includegraphics[width=0.95\textwidth]{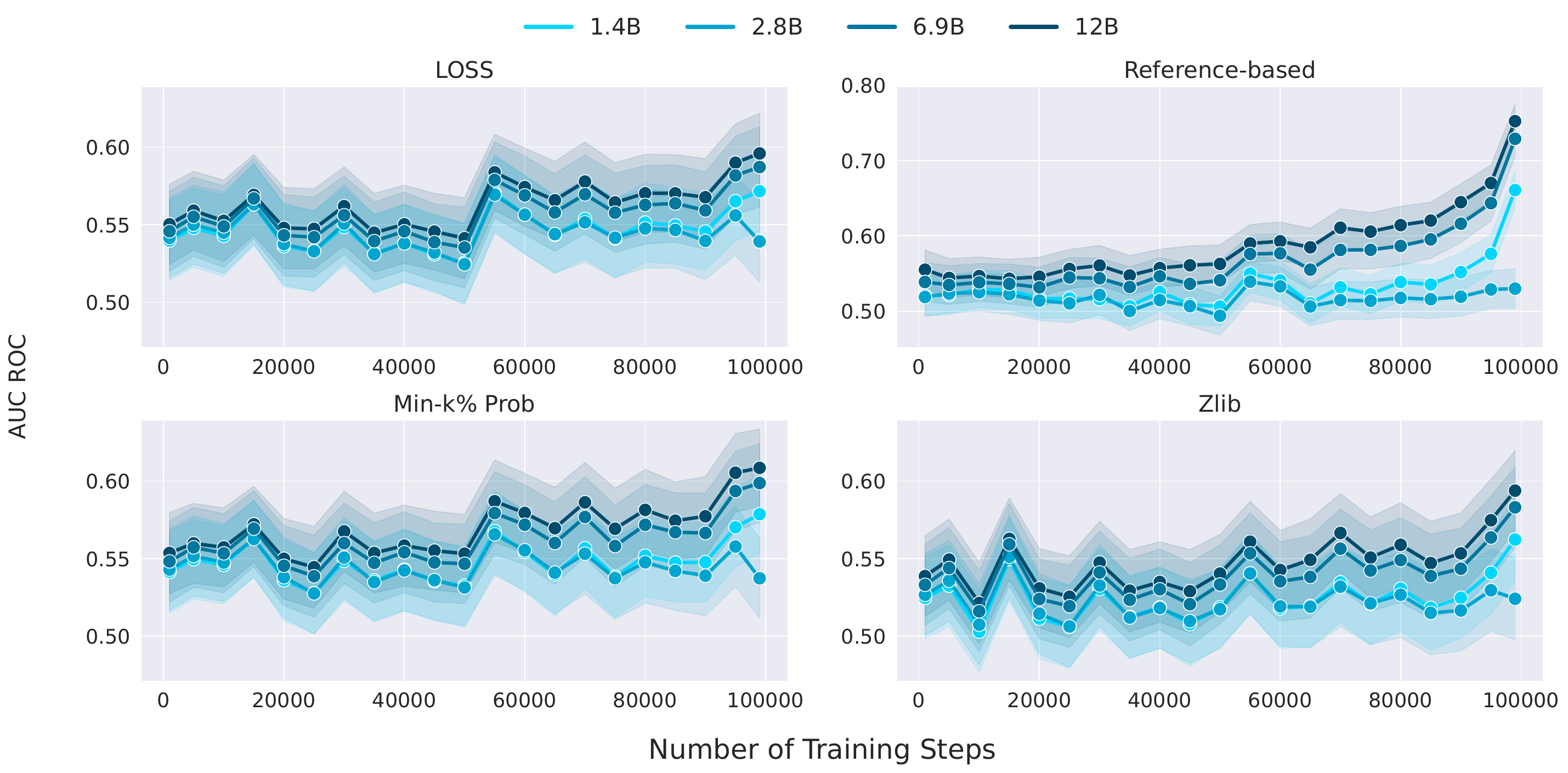}
\vspace{-1em}
\caption{MIA performance for different member data sets sampled at different training steps across 1 epoch of the deduplicated Pile pretraining corpus, visualized across different attacks. Target model is the \pythiadedup-12B checkpoint at step-99000. AUC-ROC reported. \textbf{Performance on benchmarks with more recently seen members is higher, but gradually decreases to a plateau for less recently seen members}.
}
\label{fig:data-recency}
\end{center}
\end{figure*}

Figure~\ref{fig:data-recency} demonstrates that, in general, member data seen more recently by the given checkpoint contributes to slightly higher MIA performance. We believe this supports existing work in LM forgetting~\citep{jagielski2023measuring}, where observed patterns in recently seen training data are better preserved in the model parameters, while earlier seen data are less memorized. 

We also note that the MIA performance trajectories seem to drop slightly more quickly for smaller models, though the trajectories across all model sizes seem to converge when evaluating on member data from much earlier in the training run. We speculate this is a result of larger models having more parameters, allowing them to capture more seen data before having to drop older knowledge.

We also note that, in the context of MIA against fine-tuning datasets, our results indicate that data seen during fine-tuning or continued pre-training may also be increasingly vulnerable due to how recent they are seen. This aligns with previous work demonstrating high MIA performance on fine-tuning datasets~\citep{mireshghallah2022quantifying, fu2023practical}. This is especially relevant in practice since fine-tuning is a popular option to re-purpose large pre-trained models for varying downstream tasks such as commercial use cases, which often involves tuning with sensitive data.

\subsection{Number of Training Epochs} \label{sec:appendix-training-epoch}

\begin{figure*}[h]
\begin{center}
\includegraphics[width=.95\columnwidth]{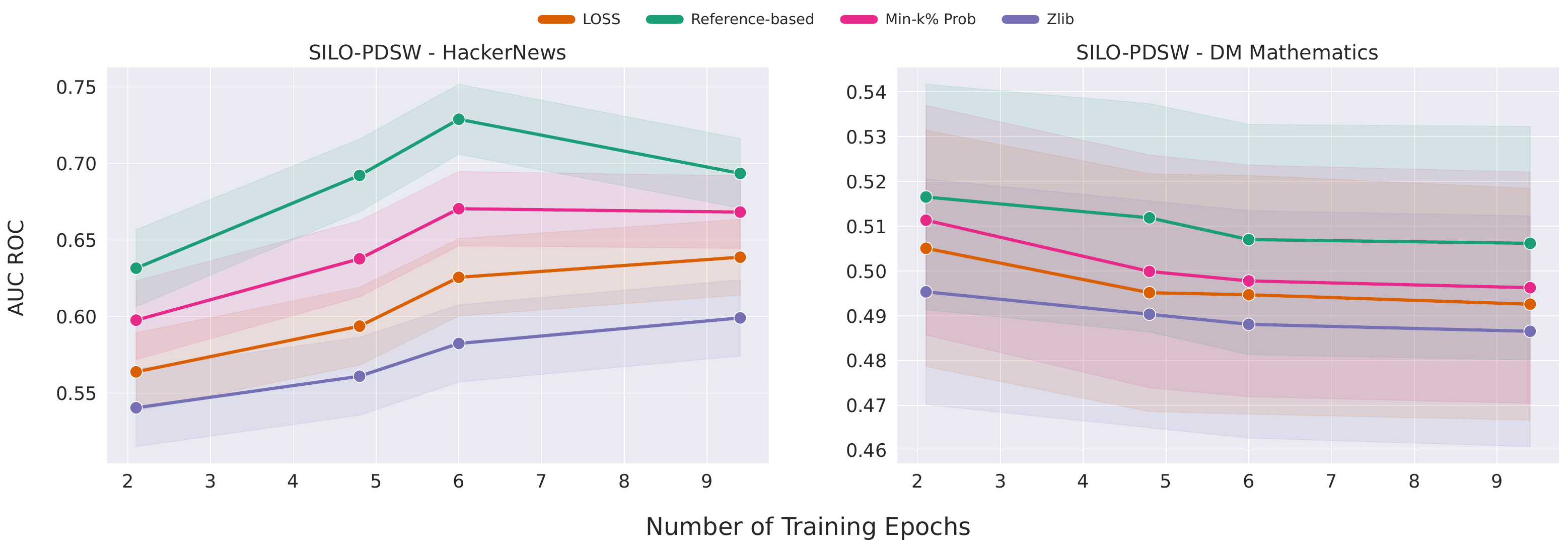}
\vspace{-1em}
\caption{MIA performance on target model \silo-PDSW as the number of effective epochs in which the member domain data has been seen increases. AUC-ROC reported. For HackerNews, \textbf{performance does increase with an increasing number of effective epochs initially, but begins to plateau or even drop with further epochs}. For DM Mathematics, \textbf{performance surprisingly drops with increasing effective epochs}.}
\label{fig:num-epochs-silo}
\end{center}
\end{figure*} 

While experiments against the \datablations\ model (\Cref{fig:training-data-size}, right) operate in a fixed training data size setting, we also explore a more realistic setting where the amount of training data the target model sees increases alongside the effective epoch count by targeting the \silo-PDSW model and intermediate checkpoints.
In \Cref{fig:num-epochs-silo}, for HackerNews, we observe MIA performance initially increases with more effective epochs, similar to the \datablations\ setting, but then begins to plateau or drop as effective epoch count continues to increase. DM mathematics, on the other hand, surprisingly decreases as the number of effective epochs increases. We speculate over factors that may contribute to these observations:
\begin{itemize}
    \item HackerNews, even when up-sampled for this variant of the \silo-PDSW model, still only makes up 5.9\% of the training data~\citep{silo}. In the first few epochs, when the total training data seen is low, the model can memorize the HackerNews samples. However, as the number of epochs increases, the target model may tend to overfit data more so from domains with greater representation. As the \silo\ model is on the smaller side with 1.4B parameters, we suspect the target model begins to memorize less of the HackerNews samples, leading to a plateau or drop in MIA performance.
    \item DM Mathematics also makes up only 3.5\% of the training data. In addition, with DM Mathematics being a dataset of mathematical problems, we suspect that the abundance of tokens from a concentrated token space (\ie digits, variables) that are largely symbolic rather than semantic makes memorization of specific samples unlikely. Overall, it simply fails to perform well on such data even after multiple epochs (as observed when looking at model loss values for this data).
\end{itemize}
For both cases, further investigation is needed into the target domains and attack setting setup to better understand these counter-intuitive phenomena.

\newpage
\section{Revisiting Membership through Semantic Similarity}\label{sec:appendix-semantic-mi}

We follow the same pipeline as~\Cref{app:edit_distance_members} but focus on generating member paraphrases that are semantically similar while preserving as much specific information from the original member sample as possible. To do so, we follow~\cite{yang2023rethinking} and prompt GPT4~\citep{openai2024gpt4} to paraphrase member samples in a different style (5 trials per member). We perform this paraphrasing over the ArXiv, Wikipedia, and HackerNews domains; see~\Cref{tab:appendix_llm_prompts} for domain-specific prompts. In general, we don't focus on the lexical similarity of the paraphrases unlike the earlier semantically similar samples generated via masking and replacing a small percentage of tokens. However, with HackerNews, we do specify a comment structure different from how HackerNews records were formatted for model training, a noticeable lexical difference.

\begin{figure}[]
\begin{center}
\includegraphics[width=.8\textwidth]{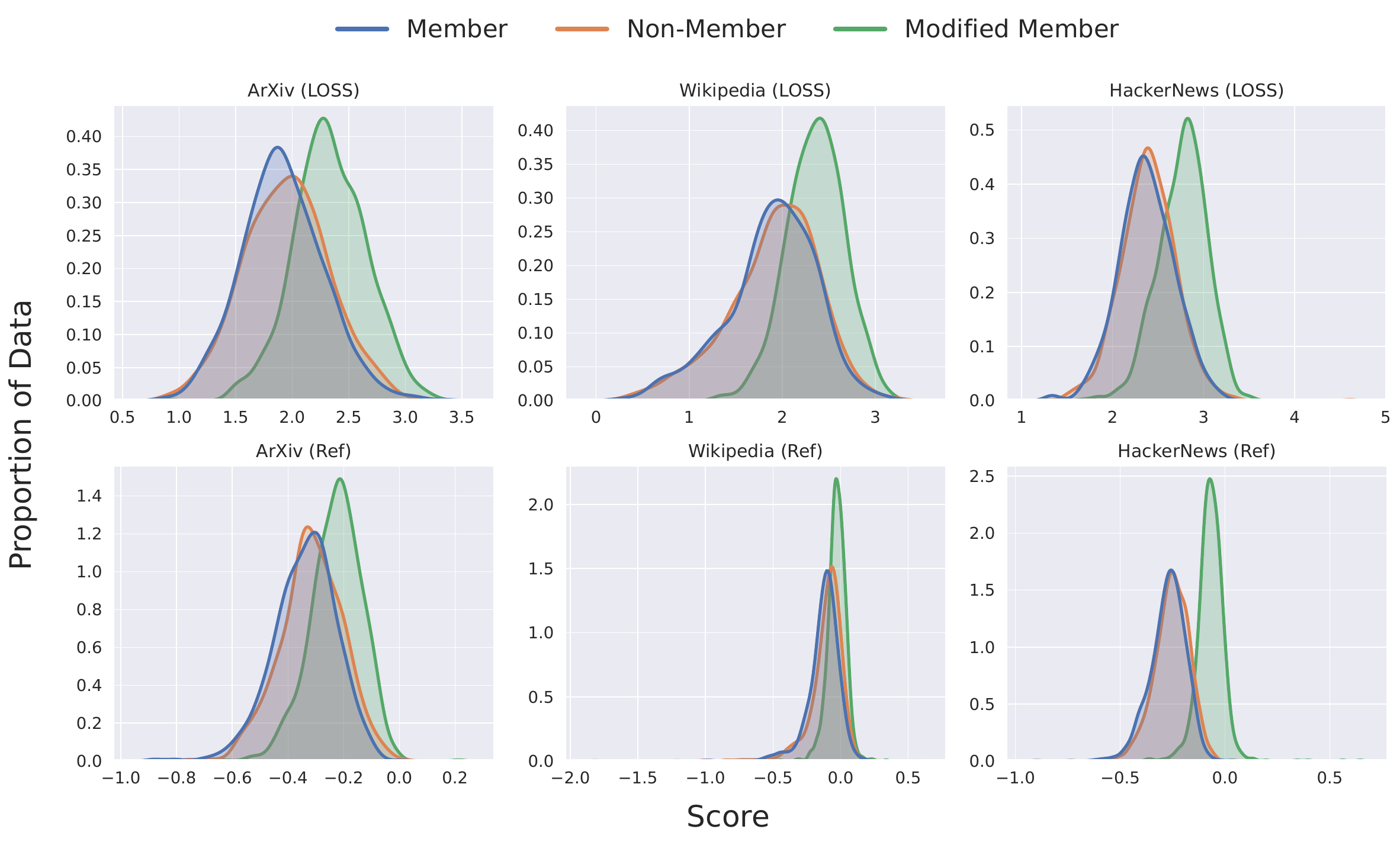}
\caption{Distribution of scores for LOSS and Reference-based attacks for members, non-members, and GPT4-paraphrased (modified) members across ArXiv, Wikipedia, and HackerNews domains. 
}
\label{fig:score_distributions_gpt_paraphrase}
\end{center}
\end{figure}

\begin{table}[h]
    \centering
    \renewcommand{\arraystretch}{1.5}
    \begin{tabularx}{\textwidth}{p{0.2\textwidth}|X}
        \hline
        \textbf{Domain} & \textbf{Prompt} \\
        \hline
        ArXiv & Please help me paraphrase the following text chunk from a research paper in a different style. Importantly, for sentences containing specific details like mathematical definitions or proofs, only make minimal changes and ensure these details are included exactly in the paraphrase.  If the paper includes a title or authors, please keep them in the rephrase. If not, please DO NOT make up a title. Use a similar number of words. \\
        \hline
        Wikipedia & Please help me paraphrase the following text chunk from Wikipedia in a different but concise style. Importantly, for sentences containing specific details, make minimal changes and ensure all details are included correctly in the paraphrase. Use a similar number of words.\\
        \hline
        HackerNews & Please help me paraphrase the following conversation chunk from a thread in HackerNews while maintaining the conversational style. Follow this structure for each comment in the thread: [user] - [comment]. Ensure all user's comments are represented in the paraphrase. Make sure all details in each user's comments are included correctly in the paraphrase, such as links. Be specific and don't generalize. \\
        \hline
    \end{tabularx}
    \caption{Instructions used to prompt GPT4 to obtain paraphrased members.}
    \label{tab:appendix_llm_prompts}
\end{table}
We again visualize the score distributions between the paraphrased members, and original members and non-members (\Cref{fig:score_distributions_gpt_paraphrase}). We observe in general across both the LOSS and Reference-based attacks over the three domains that the paraphrased member score distributions are distinguishable from the original member and non-member score distributions but have noticeable overlap, similar to what was observed with masking-based semantic neighbors. However, when we perform the FPR experiment (\Cref{tab:gpt_paraphrase_fprs}), we see that in high-confidence settings, the paraphrased members are likely to be classified as non-members. Both the LOSS and Reference-based attacks seem noticeably insensitive to such paraphrased neighbors.

\begin{table}[]
\begin{center} 
\scriptsize %
\begin{tabularx}{.75\columnwidth}{l *{6}{>{\centering\arraybackslash}X}@{}}
    \toprule
    \multirow{2}{*}{Domain} & \multicolumn{3}{c}{LOSS} & \multicolumn{3}{c}{Ref} \\
    \cmidrule(lr){2-4}  \cmidrule(lr){5-7}
    & 1\% & 5\% & 10\% & 1\% & 5\% & 10\% \\
    \midrule
    ArXiv     & 0.1 & 0.2 & 0.5 & 0.2 & 0.7 & 1.7\\
    Wikipedia & 0.0 & 0.0 & 0.2 & 0.0 & 0.2 & 0.7\\
    HackerNews & 0.1 & 1.1 & 1.7 & 0.0 & 0.1 & 0.2\\
    \bottomrule
\end{tabularx}
\end{center}
\vspace{-.5em}
\caption{FPR (\%) on modified members (treated as non-members) when using a score threshold that achieves a 1, 5, or 10\% FPR on the original member and non-member data for the ArXiv, Wikipedia, and HackerNews domains. Modified members are generated by prompting GPT4 to paraphrase member samples with significant lexical difference. LOSS and Reference-based attack reported.}\label{tab:gpt_paraphrase_fprs}
\vspace{-.5em}
\end{table}

\newpage
\section{Additional Figures and Tables}\label{sec:additional-experimental-results}

\begin{table*}[h]
\scriptsize
\begin{center}
\setlength{\tabcolsep}{.7pt}
\begin{tabularx}{\textwidth}{l *{20}{>{\centering\arraybackslash}X}@{}}
    \toprule
    \multirow{2}{*}{}  & \multicolumn{5}{c}{Wikipedia} & \multicolumn{5}{c}{Github} & \multicolumn{5}{c}{Pile CC} & \multicolumn{5}{c}{Pubmed Central} \\
    \cmidrule(lr){2-6}  \cmidrule(lr){7-11} \cmidrule(lr){12-16} \cmidrule(lr){17-21}
    \textbf{\# Params} & \tiny{LOSS} & \tiny{Ref} & \tiny{\mink} & \tiny{zlib} & \tiny{Ne}
    & \tiny{LOSS} & \tiny{Ref} & \tiny{\mink} & \tiny{zlib} & \tiny{Ne}
    & \tiny{LOSS} & \tiny{Ref} & \tiny{\mink} & \tiny{zlib} & \tiny{Ne}
    & \tiny{LOSS} & \tiny{Ref} & \tiny{\mink} & \tiny{zlib} & \tiny{Ne}
    \\
    \midrule
        70M & 0.8 & 0.9 & \textbf{1.0} & \textbf{1.0} & 0.5 &
        \textbf{13.1} & 7.8 & 12.6 & 13.0 & 11.0 &
        0.7 & \textbf{1.0} & 0.4 & 0.4 & 0.1 &
        0.7 & \textbf{1.0} & 0.9 & 0.5 & 0.6
        \\
        160M & 1.1 &0.8&1.2&\textbf{1.4}&1.3    &13.5&4.6&12.3&\textbf{14.7}&5.9    &0.4&\textbf{0.8}&0.5&0.4&0.4    &0.7&0.9&\textbf{1.0}&0.3&0.1\\
        1.4B & 0.6&\textbf{0.9}&0.5&0.7&0.4    &12.8&0.7&12.9&\textbf{16.4}&3.9    &0.6&0.6&0.5&0.7&\textbf{0.8}    &0.4&\textbf{0.7}&0.6&0.5&0.1\\
        2.8B &0.6&0.8&0.5&0.7&\textbf{0.9}    &20.8&4.5&20.8&\textbf{23.4}&11.1    &0.6&0.5&0.7&0.8&\textbf{0.9}    &0.4&1.0&\textbf{1.4}&0.6&0.9\\
        6.9B &\textbf{0.6}&\textbf{0.6}&0.4&\textbf{0.6}&0.5    &12.9&0.6&13.1&\textbf{16.8}&6.1    &1.0&\textbf{1.4}&1.2&1.3&1.0    &0.8&\textbf{1.6}&0.8&0.3&0.8\\
        12B & \textbf{0.7} & 0.6 & 0.6 & \textbf{0.7} & 1.0    & 13.9 & 0.8 & 14.2 & \textbf{17.4} & 4.9    & 1.0 & \textbf{1.7} & 1.1 & 1.5 & 1.0    & 1.0& \textbf{1.5} & 1.3 & 0.7 & 0.9\\
        \vspace{-.6em} \\
    \toprule
    \multirow{2}{*}{}  & \multicolumn{5}{c}{ArXiv} & \multicolumn{5}{c}{DM Math} & \multicolumn{5}{c}{HackerNews} & \multicolumn{5}{c}{The Pile}\\
    \cmidrule(lr){2-6}  \cmidrule(lr){7-11} \cmidrule(lr){12-16} \cmidrule(lr){17-21}
    \textbf{\# Params} & \tiny{LOSS} & \tiny{Ref} & \tiny{\mink} & \tiny{zlib} & \tiny{Ne}
    & \tiny{LOSS} & \tiny{Ref} & \tiny{\mink} & \tiny{zlib} & \tiny{Ne}
    & \tiny{LOSS} & \tiny{Ref} & \tiny{\mink} & \tiny{zlib} & \tiny{Ne}
    & \tiny{LOSS} & \tiny{Ref} & \tiny{\mink} & \tiny{zlib} & \tiny{Ne}
    \\
    \midrule
        70M & 0.8 & \textbf{1.0 }& 0.7 & 0.5 & 0.9 & 
        0.6 & \textbf{1.3} & 0.5 & 0.7 & 0.8 &
        1.1 & 0.7 & \textbf{1.3} & \textbf{1.3} & 0.5 &
        \textbf{2.2} & 1.4 & 1.8 & 2.1 & 2.0\\
        160M &\textbf{0.8}&0.4&0.2&0.7&0.3    &0.5&\textbf{1.4}&0.6&1.2&0.7    &1.0&0.8&\textbf{1.2}&0.6&0.7    &\textbf{2.4}&1.3&2.0&2.2& 2.2\\
        1.4B &0.3&\textbf{1.0}&0.2&0.4&0.7    &0.8&0.8&0.6&1.0&\textbf{1.7}    &0.7&0.9&\textbf{1.2}&0.9&0.8    &\textbf{2.4}&1.4&\textbf{2.4}&2.3&2.3\\
        2.8B &0.5&\textbf{2.1}&0.5&0.5& 0.5   &0.8&0.4&1.0&\textbf{1.3}&0.8    &0.6&1.4&0.8&1.1&\textbf{1.7}    &\textbf{2.8}&2.2&\textbf{2.8}&\textbf{2.8}&2.4\\
        6.9B &0.6&\textbf{1.8}&0.6&0.6&0.6    &0.9&0.2&0.6&\textbf{1.0}&0.7    &.9&\textbf{1.9}&1.0&0.9&1.3    &\textbf{2.6}&1.8&2.5&2.5& 2.2\\
        12B & 0.6& \textbf{2.5}& 0.6&0.5 & 0.9    & 1.0&0.5&0.5&\textbf{0.9}& 0.8    & 0.7&\textbf{2.3}&0.8&0.8& 1.4    &\textbf{2.7}&1.8&2.6&2.6& 2.2\\
    \bottomrule
\end{tabularx}
\end{center}
\vspace{-.5em}
\caption{\%TPR@1\%FPR of MIAs against \pythiadedup\ across different datasets from the Pile. The highest performance across the different MIAs is bolded per domain. In general, \textbf{leakage in high confidence settings is low ($<3$\%)}. As with AUC ROC, GitHub is an exception, still yielding considerably higher leakage with most attacks. Unlike with AUC ROC, trends in performance are much noisier in the high-confidence setting, with trends in model size and best-performing attacks in certain domains no longer holding, reinforcing the difficulty in determining a \textit{best} attack.}\label{tab:appendix-gen_exp_tprlowfpr_results}
\vspace{-.5em}
\end{table*}

\begin{table*}[h]
\begin{center} \scriptsize
\setlength{\tabcolsep}{0.7pt}
\begin{tabularx}{\textwidth}{l *{16}{>{\centering\arraybackslash}X}@{}}
    \toprule
    \multirow{2}{*}{}  & \multicolumn{4}{c}{Wikipedia} & \multicolumn{4}{c}{Github} & \multicolumn{4}{c}{Pile CC} & \multicolumn{4}{c}{Pubmed Central} \\
    \cmidrule(lr){2-5}  \cmidrule(lr){6-9} \cmidrule(lr){10-13} \cmidrule(lr){14-17}
    \textbf{\# Params} & LOSS & Ref & \mink & zlib
    & LOSS & Ref & \mink & zlib
    & LOSS & Ref & \mink & zlib 
    & LOSS & Ref & \mink & zlib
    \\
    \midrule
        160M & .503& \textbf{.512} &.491& \textbf{.512}    & .657& .639 & .652& \textbf{.674}    & .496 & .491 & \textbf{.504} & .497   
 & .499 & \textbf{.513} & .506 & .500 \\
        1.4B &.513& \textbf{.552}& .511& .520    & .698 & .670  & .699 & \textbf{.710}    & .501 & \textbf{.522 }& .510 & .502   
 & .498 & \textbf{.531} & .502 & .500\\
        2.8B & .518& \textbf{.582} & .518& .525    & .712 & .653 & .713 & \textbf{.723}  & .501 & \textbf{.537} & .508 & .504  
  & .500 & \textbf{.537} & .504 & .501 \\
        6.9B & .528& \textbf{.618}& .536& .536    & .730 & .644 & .733 & \textbf{.739}    & .507& \textbf{.550}& .515& .509  
  & .506 & \textbf{.558} & .511 & .506 \\
        12B & .535& \textbf{.639} &.544& .544    & .740 & .630 & .743 & \textbf{.748 }   & .511 & \textbf{.567} & .517 & .512   & .513 & \textbf{.582} & .523 & .512 \\
        \vspace{-.6em} \\
    \toprule
    \multirow{2}{*}{}  & \multicolumn{4}{c}{ArXiv} & \multicolumn{4}{c}{DM Math} & \multicolumn{4}{c}{HackerNews} & \multicolumn{4}{c}{The Pile}\\
    \cmidrule(lr){2-5}  \cmidrule(lr){6-9} \cmidrule(lr){10-13} \cmidrule(lr){14-17}
    \textbf{\# Params} & LOSS & Ref & \mink & zlib
    & LOSS & Ref & \mink & zlib
    & LOSS & Ref & \mink & zlib
    & LOSS & Ref & \mink & zlib
    \\
    \midrule
        160M & \textbf{.510} & .494 & .507 & .502    & .489 & \textbf{.510} & .495 & .481   & .494 & .491 & \textbf{.509 }& .498    &.503 &\textbf{.511} &.505 &.506 \\
        1.4B & .515 & .516 & \textbf{.517} & .509   & .486 & \textbf{.512} & .497 & .482    & .505 & \textbf{.522} & .512 & .504    &.505 &\textbf{.522} &.510 &.508 \\
        2.8B & .519 & \textbf{.531} & .525 & .514    & .484 & \textbf{.505} & . 480 & .480     & .513 & \textbf{.551} & .524 & .509    &.508 &\textbf{.533} &.513 &.511 \\
        6.9B & .529 & \textbf{.558} & .535 & .523    & .485 & \textbf{.511} &.496 & .481    & .521 & \textbf{.579} & .536 & .513    &.514 &\textbf{.554} &.522 &.516 \\
        12B  & .534& \textbf{.575} & .546 & .527    & .485 & \textbf{.510}& .497 & .481    & .528 & \textbf{.606} & .546 & .517     &.519 &\textbf{.569} &.528 &.520 \\
    \bottomrule
\end{tabularx}
\end{center}
\vskip -0.5em
\caption{AUC ROC of MIAs against non-deduped \pythia\ across different datasets from the Pile. The reference-based attack uses \textsc{stablelm-base-alpha-3b-v2} as the reference model. The highest performance across the different MIAs is bolded per domain.
\textbf{Performance follows similar trends seen when targeting the \pythiadedup~models, but performance is, in general, marginally higher}. Due to computational limitations, we leave out evaluations for the Neighborhood attack, but expect similar trends.}\label{tab:appendix-nondedupe_pythia_results}
\vskip -0.5em
\end{table*}

\begin{figure*}[t]
\begin{center}
\includegraphics[width=.95\textwidth]{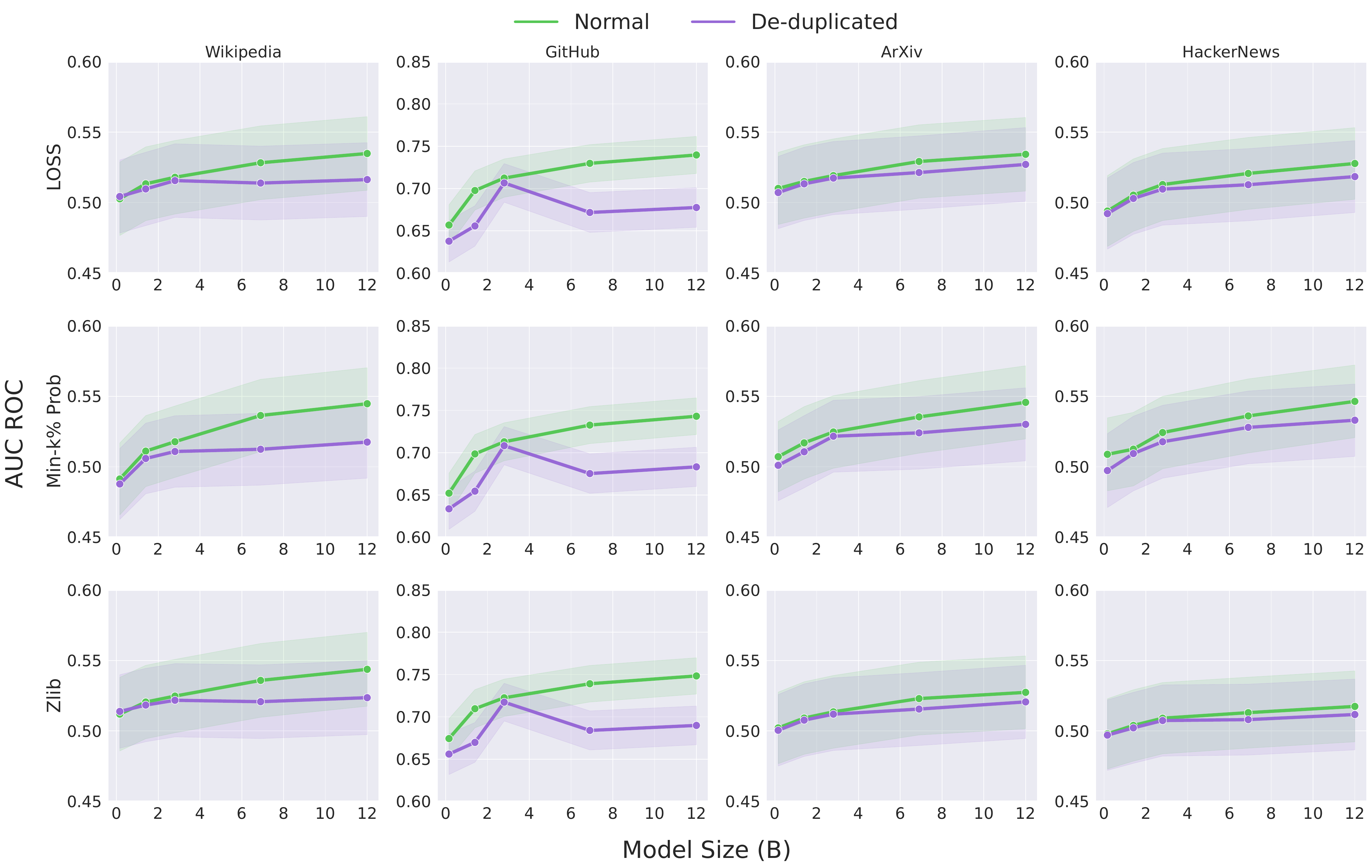}
\vspace{-1em}
\caption{
    MIA performance as model size increases over select domains for various other attacks. We additionally plot the AUC ROC trajectory against the non-deduped Pythia suite for comparison. Similar to the reference-based attack, \textbf{increasing model size slightly boosts MIA performance while deduplication decreases performance}.
}\label{fig:gen_exp_model_size+dedup_others}
\end{center}
\end{figure*}

\begin{figure*}[t]
\begin{center}
\includegraphics[width=\columnwidth]{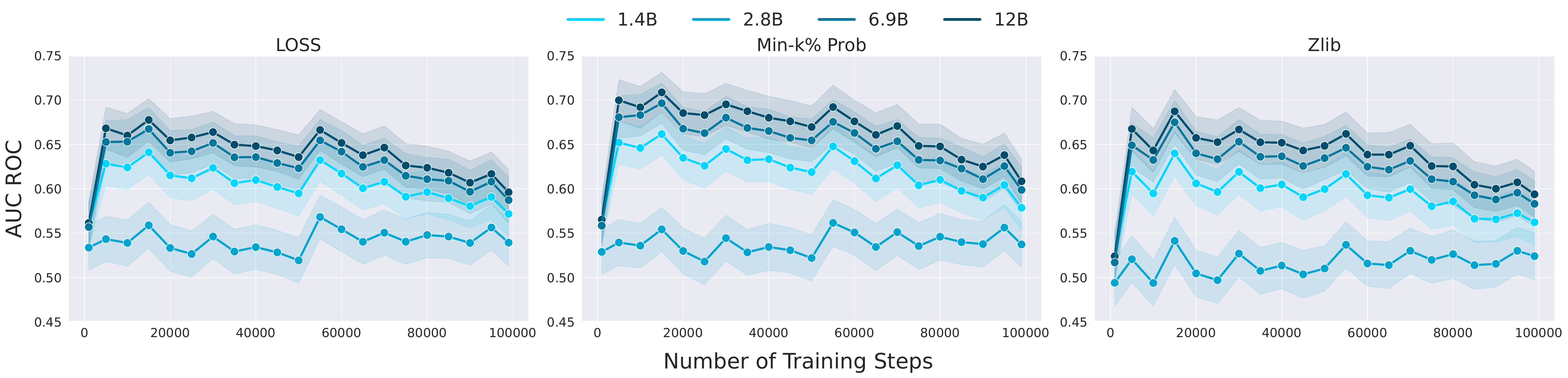}
\caption{MIA performance as the amount of training data seen increases across 1 epoch of the deduplicated Pile pretraining corpus, visualized over a range of model sizes for various attacks. We use the training step as a unit for the amount of training data seen, with 1 step corresponding to seeing 2097152 tokens. AUC-ROC reported. Similar to the reference-based attack, for all attacks, \textbf{performance drastically increases before gradually decreasing as the amount of training data seen increases}.
}
\label{fig:training-data-size-others}
\end{center}
\end{figure*}

\begin{figure*}[t]
\begin{center}
\includegraphics[width=.85\textwidth]{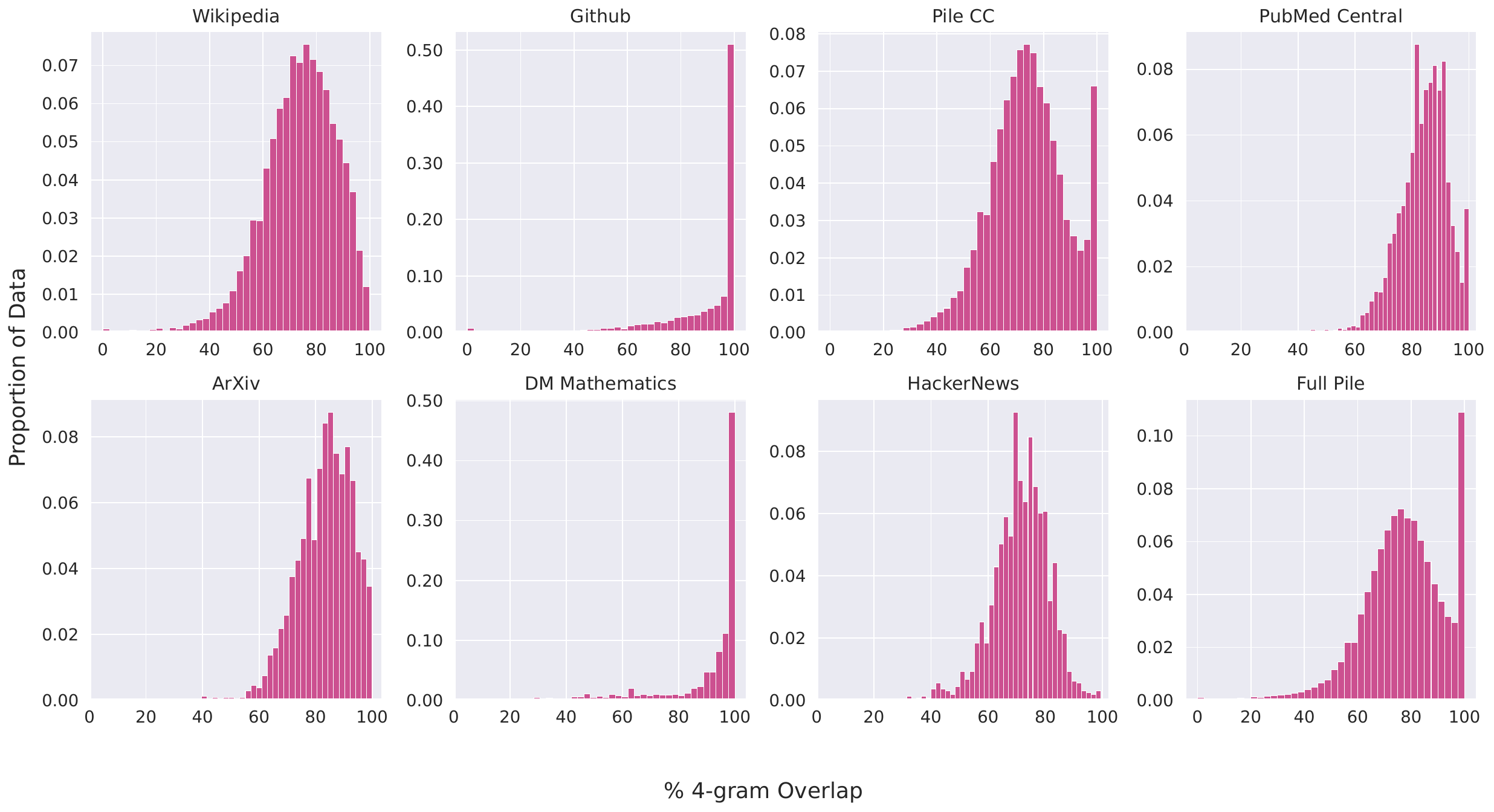}
\includegraphics[width=.85\textwidth]{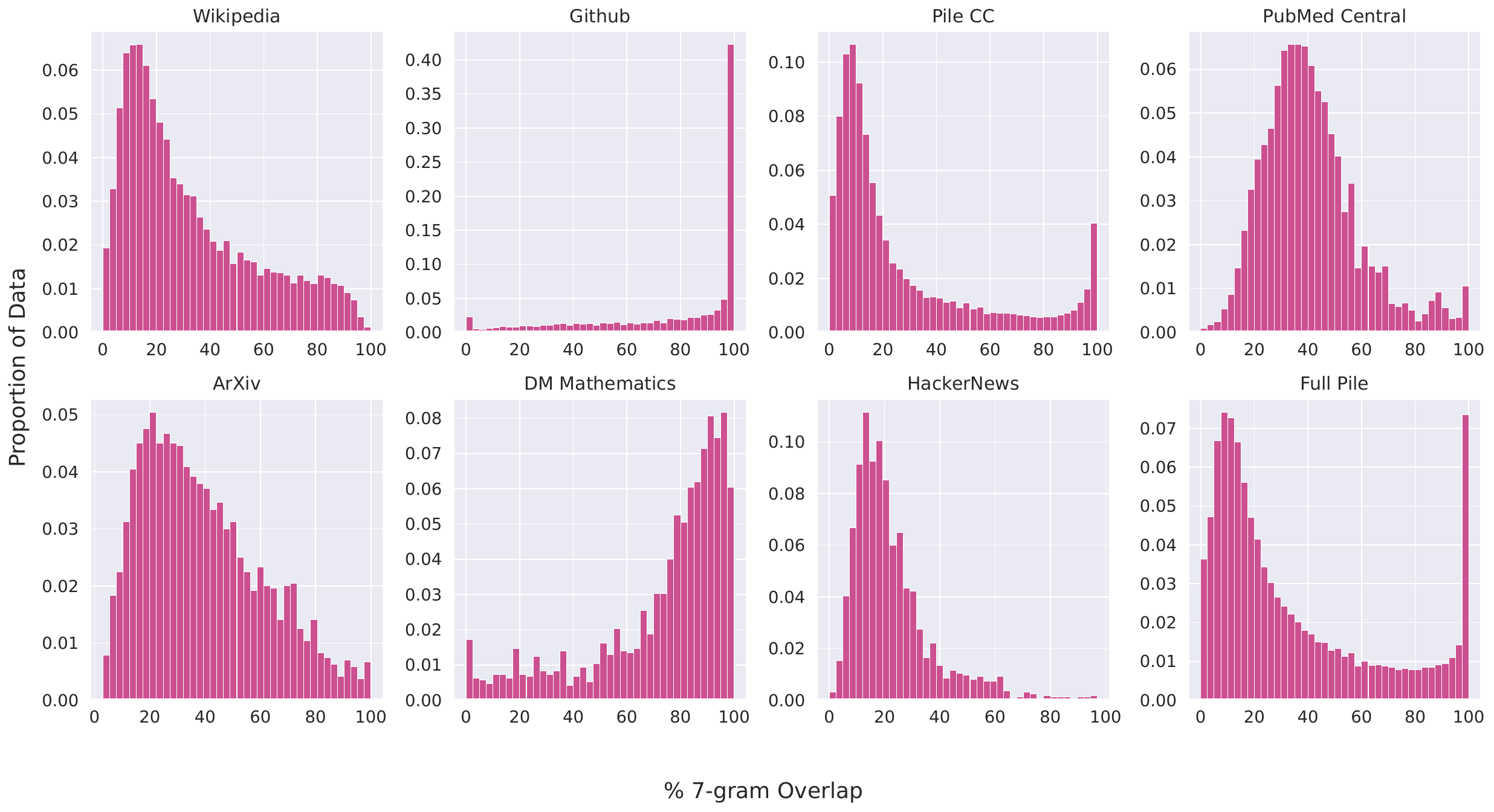}
\includegraphics[width=.85\textwidth]{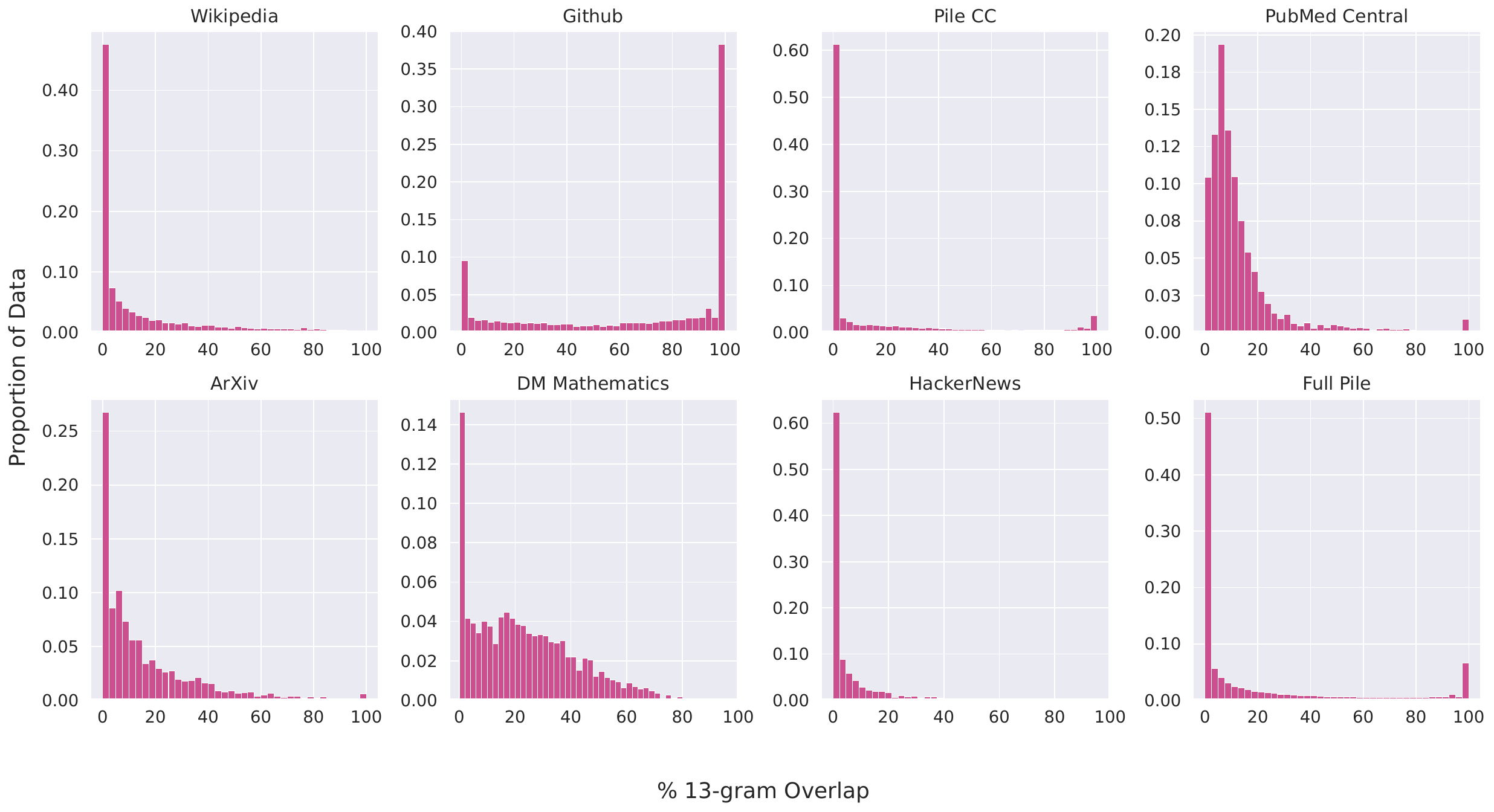}
\vspace{-1em}
\caption{Distribution of \ngram\ overlap over all evaluation domains for $n=4,7,13$.}
\label{fig:appendix_other_ngram_overlap}
\end{center}
\end{figure*}

\end{document}